\PassOptionsToPackage{export}{adjustbox}
\documentclass{arxivtmpl}

\usepackage{fontawesome5}

\def\1{\bm{1}}

\DeclareMathAlphabet{\mathsfit}{\encodingdefault}{\sfdefault}{m}{sl}
\SetMathAlphabet{\mathsfit}{bold}{\encodingdefault}{\sfdefault}{bx}{n}

\tcbuselibrary{skins,breakable}
\newtcolorbox{findingbox}{enhanced,breakable,frame hidden,colback=gred!6,borderline west={3pt}{0pt}{gred},sharp corners,boxsep=0pt,left=12pt,right=10pt,top=8pt,bottom=8pt,before skip=12pt,after skip=12pt}
\newtcolorbox{notebox}{enhanced,breakable,frame hidden,colback=gblue!6,borderline west={3pt}{0pt}{gblue},sharp corners,boxsep=0pt,left=12pt,right=10pt,top=8pt,bottom=8pt,before skip=12pt,after skip=12pt}
\newtcolorbox{tipbox}{enhanced,breakable,frame hidden,colback=ggreen!6,borderline west={3pt}{0pt}{ggreen},sharp corners,boxsep=0pt,left=12pt,right=10pt,top=8pt,bottom=8pt,before skip=12pt,after skip=12pt}
\newtcolorbox{warnbox}{enhanced,breakable,frame hidden,colback=gyellow!6,borderline west={3pt}{0pt}{gyellow},sharp corners,boxsep=0pt,left=12pt,right=10pt,top=8pt,bottom=8pt,before skip=12pt,after skip=12pt}

\tcbuselibrary{listings}
\newcommand{\boxfont}{\fontsize{9}{11}\selectfont}
\newtcblisting{promptbox}[1][]{%
  enhanced, listing only, breakable, frame hidden, width=\linewidth,
  colback=gblue!6, borderline west={3pt}{0pt}{gblue},
  sharp corners, boxsep=0pt,
  left=12pt, right=10pt, top=8pt, bottom=8pt, before skip=12pt, after skip=12pt,
  before upper={\def\promptlabel{#1}\ifx\promptlabel\empty\else{\sffamily\bfseries\color{gblue}#1}\par\smallskip\fi},   
  listing options={basicstyle=\ttfamily\boxfont, escapeinside={(*}{*)}},
}

\newtcolorbox{transcriptbox}{%
  enhanced, breakable, frame hidden,
  colback=ggreen!6, borderline west={3pt}{0pt}{ggreen},
  sharp corners, boxsep=0pt, fontupper=\ttfamily\boxfont, halign upper=flush left,   
  left=12pt, right=10pt, top=6pt, bottom=8pt, before skip=12pt, after skip=12pt,
}

\newcommand{\teaserfigure}[2]{%
  \begingroup\centering\captionsetup{type=figure}
  #1\par\vspace{0.4em}\caption{#2}%
  \par\endgroup\vspace{1em}
}

\colorlet{codebg}{gyellow!6}
\setminted{fontsize=\boxfont, breaklines, autogobble, tabsize=4, style=vs}
\newminted[codebox]{python}{bgcolor=codebg, frame=leftline, framerule=2.5pt, rulecolor=gyellow, framesep=2.5mm}

\hypersetup{
    colorlinks=true,
    linkcolor=linkcol,
    citecolor=linkcol,
    urlcolor=linkcol,
    pdfborderstyle={},
    pdfborder={0 0 0}
}
\renewcommand\Affilfont{\sffamily\mdseries}

\makeatletter
\newcommand\cardaffil[2]{\@namedef{card@inst@#1}{#2}}
\newcount\card@n \newcount\card@cols \newdimen\card@wd
\newcommand\carda[3]{%
  \parbox[t]{\card@wd}{\raggedright\sffamily
    {\fontsize{10}{12}\selectfont\strut#3\ifx\relax#2\relax\else\smash{\textsuperscript{\sffamily\color{ggrey}#2}}\fi\par}%
    {\fontsize{8}{10}\selectfont\color{ggrey}\strut\ifx\relax#1\relax\else\@nameuse{card@inst@#1}\fi\par}}%
  \global\advance\card@n 1
  \ifnum\card@n=\card@cols \global\card@n=0 \par\vspace{5pt}\noindent\fi}
\newcommand\authorcards[2]{\par{\card@cols=#1 \card@wd=\dimexpr\linewidth/#1\relax \global\card@n=0
  \setlength\parskip{0pt}\setlength\lineskip{0pt}\setlength\lineskiplimit{0pt}\setlength\parindent{0pt}\noindent#2\par}}
\makeatother

\colorlet{perfhue}{gblue}
\colorlet{perf0}{white}
\colorlet{perf10}{perfhue!12}
\colorlet{perf20}{perfhue!22}
\colorlet{perf30}{perfhue!33}
\colorlet{perf40}{perfhue!44}
\colorlet{perf50}{perfhue!55}
\colorlet{perf60}{perfhue!66}
\colorlet{perf70}{perfhue!77}
\colorlet{perf80}{perfhue!88}

\newcommand{\RN}[1]{%
	\textup{\lowercase\expandafter{\it \romannumeral#1}}%
}

\SetKwInput{KwInput}{Input}
\SetKwInput{KwOutput}{Output}
\DontPrintSemicolon

\SetKwComment{Comment}{\color{ggreen}\# }{}

\SetKwProg{Function}{def}{:}{}

\SetKwProg{For}{for}{:}{}
\SetKwProg{If}{if}{:}{}

\usepackage[T1]{fontenc}
\usepackage{array,longtable}
\usepackage{xspace}
\usepackage{nicefrac}
\usepackage{boldline}
\usepackage{xurl}
\usepackage{pgfplots}
\pgfplotsset{compat=newest}
\usepgfplotslibrary{units}

\newcommand{\dataset}{\textsc{OS-Omni}\xspace}

\newcommand{\platformicon}[1]{\raisebox{0.075em}{\includegraphics[height=1.05em, valign=c]{#1}}}
\newcommand{\windowsicon}{\platformicon{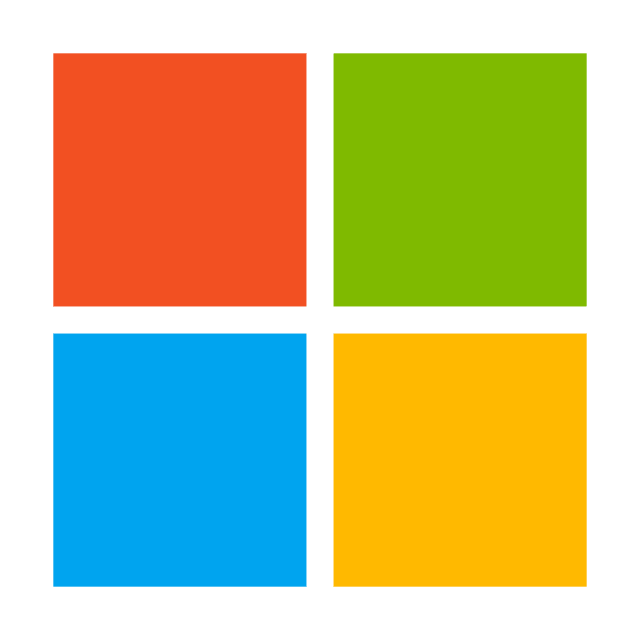}}
\newcommand{\macicon}{\platformicon{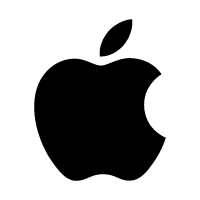}}
\newcommand{\linuxicon}{\platformicon{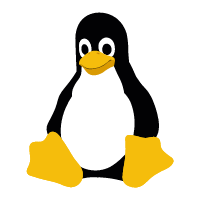}}
\newcommand{\androidicon}{\platformicon{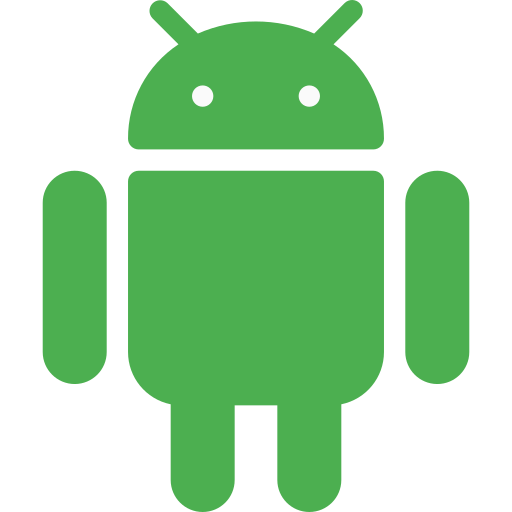}}
\newcommand{\iosicon}{\macicon}
\newcommand{\chromeicon}{\platformicon{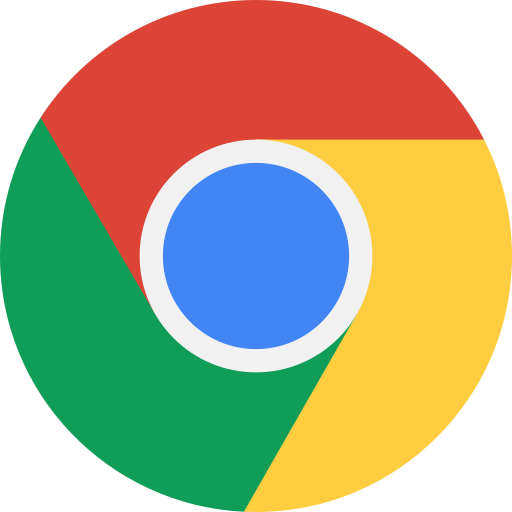}}

\newcolumntype{Y}{>{\centering\arraybackslash}X}

\newenvironment{ack}{\section*{Acknowledgments}}{}

\let\plainabstract\abstract
\renewcommand{\abstract}{\plainabstract{\sffamily\fontseries{eb}\normalsize\selectfont OS-Omni Team}\par\vspace{0.15\baselineskip}\small}

\title{OS-Omni: Benchmarking Generalist Computer-Using Agents on Long-Horizon Workflows Across Platforms}

\cardaffil{1}{University of Michigan}
\cardaffil{2}{Touchstone AI}
\cardaffil{3}{University of Pennsylvania}
\cardaffil{4}{Boston University}
\cardaffil{5}{Together AI}
\cardaffil{6}{University of Waterloo}
\cardaffil{7}{New York University}
\cardaffil{8}{UW--Madison}
\cardaffil{9}{The Ohio State University}
\cardaffil{10}{Morgan Stanley}
\cardaffil{11}{UT Austin}
\cardaffil{12}{Bytedance Seed}
\cardaffil{13}{Stanford University}
\cardaffil{14}{OpenAI}
\cardaffil{15}{MIT}
\cardaffil{16}{SGLang Omni}
\cardaffil{17}{Carnegie Mellon University}
\cardaffil{18}{Princeton University}
\cardaffil{19}{University of Hong Kong}
\cardaffil{20}{University of Minnesota}
\cardaffil{21}{NUS}
\cardaffil{22}{University of Illinois Chicago}
\cardaffil{23}{Independent}
\cardaffil{24}{Georgia Institute of Technology}
\cardaffil{1+2}{UMich, Touchstone AI}
\cardaffil{2+3}{UPenn, Touchstone AI}
\newcommand\equalauthors{%
  \carda{1+2}{*}{Hui~Shen}%
  \carda{2+3}{*}{Yunta~Hsieh}%
  \carda{2}{*}{Jianing~Ma}%
  \carda{14}{*}{Zhuoran~Yin}%
  \carda{2}{*}{Zhuhan~Bao}%
  \carda{2}{*}{Ziyuan~Liu}%
  \carda{2}{*}{Qi~Han}%
  \carda{2}{*}{Xiuqi~Xu}%
}
\newcommand\otherauthors{%
  \carda{15}{}{Xianbang~Wang}%
  \carda{18}{}{Wenhao~Chai}%
  \carda{1}{}{Sihan~Xu}%
  \carda{16}{}{Chenyang~Zhao}%
  \carda{2}{}{Yanheng~Shang}%
  \carda{6}{}{Dongfu~Jiang}%
  \carda{15}{}{Rui~Sun}%
  \carda{2}{}{Xingzhi~Li}%
  \carda{4}{}{Chenqi~Yin}%
  \carda{1}{}{Zesen~Zhao}%
  \carda{1}{}{Boyuan~Zheng}%
  \carda{5}{}{Jisen~Li}%
  \carda{5}{}{Xiaoxia~Wu}%
  \carda{2}{}{Ruoyan~Zhang}%
  \carda{3}{}{Ruiyao~Liu}%
  \carda{19}{}{Jing~Xiong}%
  \carda{19}{}{He~Xiao}%
  \carda{2}{}{Nan~Zheng}%
  \carda{2}{}{Hangxiang~Chen}%
  \carda{7}{}{Shengyang~Tao}%
  \carda{8}{}{Deyang~Luorong}%
  \carda{15}{}{Yifeng~Liu}%
  \carda{9}{}{Jingxuan~Zhang}%
  \carda{19}{}{Chaofan~Tao}%
  \carda{2}{}{Yuren~Wang}%
  \carda{8}{}{Jiaqi~Mo}%
  \carda{2}{}{Rui~Hu}%
  \carda{2}{}{Xing~Wei}%
  \carda{2}{}{Xiaofeng~Liu}%
  \carda{2}{}{Xin~Xu}%
  \carda{2}{}{Han~Wu}%
  \carda{2}{}{Mengyao~Wen}%
  \carda{2}{}{Guobin~Zhang}%
  \carda{13}{}{Shatong~Zhu}%
  \carda{1}{}{Hei~Ting~(Una)~Chan}%
  \carda{9}{}{Sicheng~Chen}%
  \carda{9}{}{Kai~Zhang}%
  \carda{19}{}{Mengkang~Hu}%
  \carda{9}{}{Zhongwei~Wan}%
  \carda{9}{}{Xin~Wang}%
  \carda{10}{}{Chuanyang~Zheng}%
  \carda{11}{}{Ziheng~Zhang}%
  \carda{19}{}{Hengyuan~Zhang}%
  \carda{19}{}{Zunhai~Su}%
  \carda{}{}{Kanzhi~Cheng}%
  \carda{24}{}{Dehan~Kong}%
  \carda{21}{}{Yian~Wang}%
  \carda{22}{}{Hanrong~Zhang}%
  \carda{8}{}{Jiawei~Lu}%
  \carda{1}{}{Yifan~Zhang}%
  \carda{16}{}{Junnan~Li}
  \carda{17}{}{Changyi~Yang}%
  \carda{16}{}{Jintao~Qu}%
  \carda{16}{}{Wenyao~Gao}%
  \carda{16}{}{Jiaxuan~Luo}%
  \carda{16}{}{Jiaxin~Deng}%
  \carda{16}{}{Yuhao~Chen}%
  \carda{16}{}{Xuesong~Ye}%
  \carda{16}{}{Jingwen~Gu}%
  \carda{16}{}{Yichi~Zhang}%
  \carda{16}{}{Qun~Lou}%
  \carda{12}{}{Shen~Yan}%
  \carda{5}{}{Ben~Athiwaratkun}%
  \carda{19}{\textcolor{black}{\faEnvelope[regular]}}{Qiushi~Sun}%
  \carda{9}{}{Mi~Zhang}%
  \carda{19}{}{Ping~Luo}%
  \carda{6}{}{Wenhu~Chen}%
  \carda{19}{}{Ngai~Wong}%
}

\hypersetup{pdftitle={OS-Omni: Benchmarking Generalist Computer-Using Agents on Long-Horizon Workflows Across Platforms},pdfauthor={Hui Shen et al.}}

\begin{document}

\makeatletter\gdef\@author{}\makeatother
\maketitle

\thispagestyle{firstpagestyle}

\begin{abstract}
{\sffamily\noindent
\begin{minipage}[t]{0.5\linewidth}{\small\bfseries\faGlobe\enspace Website}\par\vspace{1pt}{\small\href{https://os-copilot.github.io/OS-Omni-Home/}{os-copilot.github.io}}\end{minipage}%
\begin{minipage}[t]{0.5\linewidth}{\small\bfseries\faGithub\enspace Code}\par\vspace{1pt}{\small\href{https://github.com/OS-Copilot/OS-Omni}{github.com/OS-Copilot/OS-Omni}}\end{minipage}\par}
\vspace{0.9\baselineskip}
{\sffamily\fontseries{eb}\normalsize\selectfont Abstract}\par\vspace{0.15\baselineskip}
\small
Computer-using agents are increasingly expected to complete realistic user goals rather than isolated GUI actions. However, existing benchmarks typically isolate web, mobile, or desktop environments, leaving it unclear whether a single generalist agent can execute long-horizon workflows that span multiple operating systems and persistent application states.
We introduce \dataset, a live cross-platform benchmark and execution environment for evaluating computer-using agents across Windows, Ubuntu, macOS, iOS, Android, and Web environments. \dataset is released in versioned waves of 300 tasks each; this paper focuses on the first release, v1, which contains 48 tasks per platform and 12 cross-platform workflows that pair a desktop with an Android phone. \dataset provides reproducible initialization, platform-specific adapters, trajectory logging, and executable final-state evaluators, and includes controlled service-backed applications for workflows involving account, cart, budget, calendar, message, and other persistent state. 
Human studies show that \dataset tasks require 67.54 interaction steps, substantially exceeding OSWorld references of 15.0 steps. On v1, the strongest evaluated agent, Claude Opus 5.5, reaches 68.5\% average success, trailing the 77.3\% human reference by 8.8 points; the gap is uneven across platforms, with frontier models exceeding humans on Linux but trailing them by more than 25 points on iOS and macOS. These results indicate that current agents remain far from platform-robust, long-horizon computer use, especially when success requires sustained state tracking and durable final-state changes. 
\end{abstract}

\teaserfigure{\includegraphics[width=\linewidth]{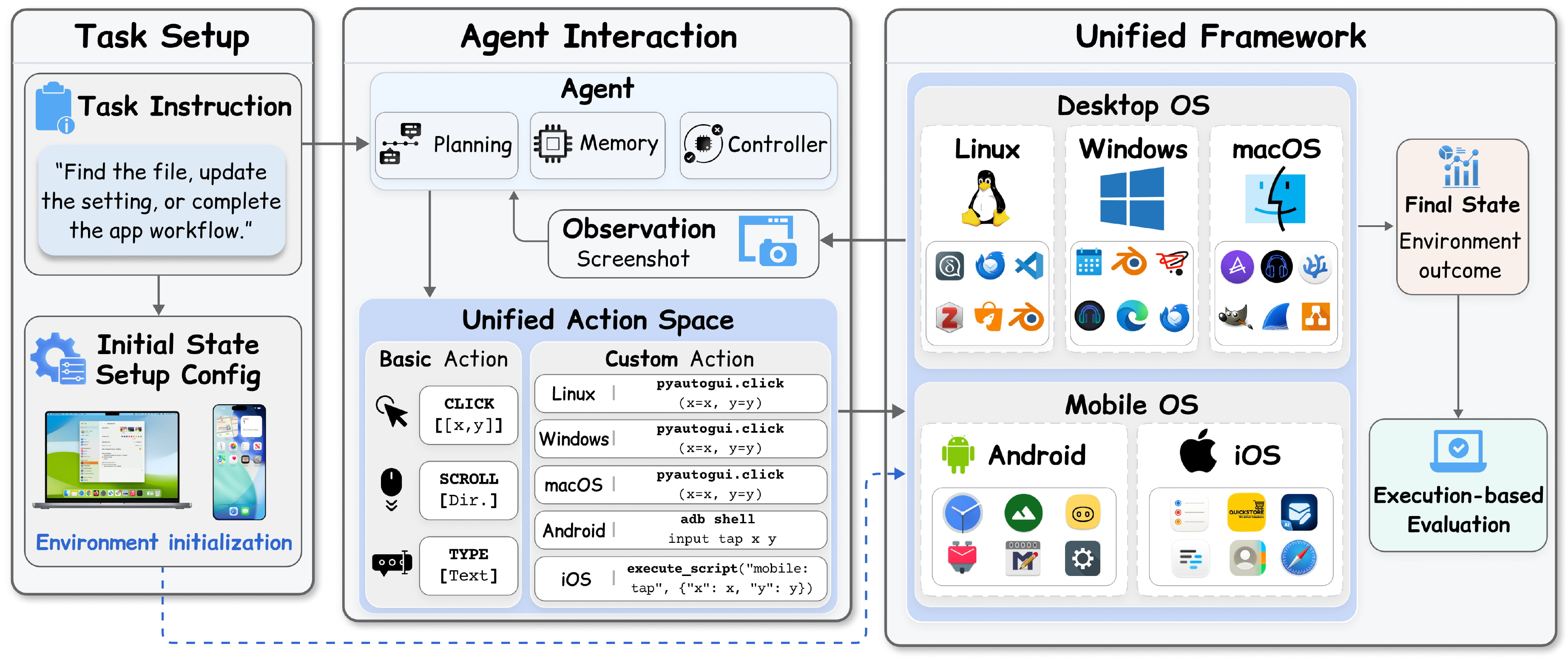}}{\textbf{Overview of OS-Omni.} \label{fig:teaser}}

\clearpage
\section*{Author list}
{\setlength{\parindent}{0pt}\raggedright\sffamily
\patchcmd{\carda}{\vspace{5pt}}{\vspace{2pt}}{}%
  {\PackageError{osomni}{Could not set author-card spacing}{}}
\vspace{0.5em}\authorcards{4}{\equalauthors}\vspace{0.5em}
\authorcards{4}{\otherauthors}\vspace{0.5em}
{\fontsize{8.5}{11.5}\selectfont\color{ggrey}* Equal contribution.\quad \textcolor{black}{\faEnvelope[regular]} Corresponding author.\\
Zhuoran Yin contributed to this work independently and in his personal capacity, not as part of his role at OpenAI\\
Contact: Hui Shen (\texttt{huishen@umich.edu}), Qiushi Sun (\texttt{qiushisun@connect.hku.hk}).\par}}
\clearpage

\section{Introduction}
A useful computer-using agent~\citep{wu2025osatlas,gou2025navigating,wang2025opencua,liu2025scalecua} should be able to carry an ordinary user goal through the software environments where that goal actually lives. 
Consider a user preparing a small event: the agent may need to compare supplies in a controlled shopping service, add selected items to a cart, update a budget spreadsheet on a desktop, create follow-up calendar entries on a phone, export a short document, and send a message with the right details attached. None of these steps is remarkable in isolation, yet the full request crosses platform conventions, spans many intermediate decisions, and can only be judged by checking the resulting cart, records, calendar entries, exported file, and message state. 
Completing such a request requires far more than isolated widget grounding: the agent must maintain cross-application context and leave every underlying software state consistent~\citep{sun2026osreward}.

This work asks: \emph{can generalist computer-using agents reliably complete realistic, long-horizon, stateful workflows across heterogeneous software platforms?} Beyond sheer horizon length, agents face compounded challenges: tracking partially updated records across window switches, adapting to OS-specific UI conventions (e.g., desktop dialogs vs. mobile picker wheels), and committing changes that survive strict post-execution inspection. A benchmark for this capability therefore needs to evaluate both breadth across platforms and depth within workflows.


Existing benchmarks have made substantial progress by studying important slices of this problem. Browser-focused benchmarks evaluate web navigation, information seeking, and visual web interaction through browser-centric action and state spaces \citep{zhou2023webarena,koh2024visualwebarena,deng2023mind2web,he2024webvoyager,sun2025genesis,gou2026mindweb}. Enterprise-web benchmarks emphasize structured work over service-backed interfaces \citep{drouin2024workarena}, while mobile benchmarks study dynamic interaction with phone applications \citep{rawles2024androidworld,sun2025sentinel,cheng2026openmobile,yan2026openmobile}. Desktop benchmarks move evaluation into executable operating-system environments, where agents must operate real applications rather than predict isolated actions \citep{xie2024osworld,bonatti2024windows}. These efforts establish that execution with verification~\citep{sun2026vericua} matters, but they still expose agents to one dominant software surface at a time, often with different platform scopes, action interfaces, task styles, and scoring protocols. 
As a result, strong performance on one family of environments can be hard to interpret as evidence of general computer use, rather than adaptation to a particular platform.

OSWorld was a major step toward general computer-using evaluation because it established execution-based evaluation in real desktop environments \citep{xie2024osworld}. However, desktop-only evaluation leaves open whether agents can handle realistic workflows that span heterogeneous software surfaces, longer task horizons, and persistent application state. \dataset extends this paradigm along two axes: \textbf{cross-platform evaluation} and \textbf{long-horizon workflows}, coupled with an execution-based evaluation protocol based on \textbf{final-state correctness}. Together, these features shift the benchmark from testing whether an agent can operate a particular interface toward testing whether it can complete a user goal across platforms. Human users require an average of \textbf{67.54} interaction steps and \textbf{200.97} seconds to complete \dataset tasks, compared with OSWorld reference tasks requiring 15.0 steps and 111.94 seconds, indicating that \dataset evaluates substantially longer and more involved workflows rather than only broader platform coverage.

These axes are coupled because failures often emerge at their intersections. Changing the platform changes the interaction bottleneck: a calendar update that is straightforward in a desktop dialog may require picker wheels, modal sheets, and soft-keyboard input on a phone. Extending the horizon changes the burden of memory and coordination: adding one contact is easier than reconciling a directory with stale duplicates, and editing one field is easier than creating, deleting, reordering, and verifying a target set of records. Final-state evaluation also changes what counts as success: a document may look visually complete while its exported file or internal formatting remains wrong, and a shopping workflow may appear complete while the cart or budget record is inconsistent. Consequently, \dataset scores tasks strictly by inspecting the persisted system state rather than intermediate visual progress.

\begin{figure}[htbp]
    \centering
    \includegraphics[width=\linewidth]{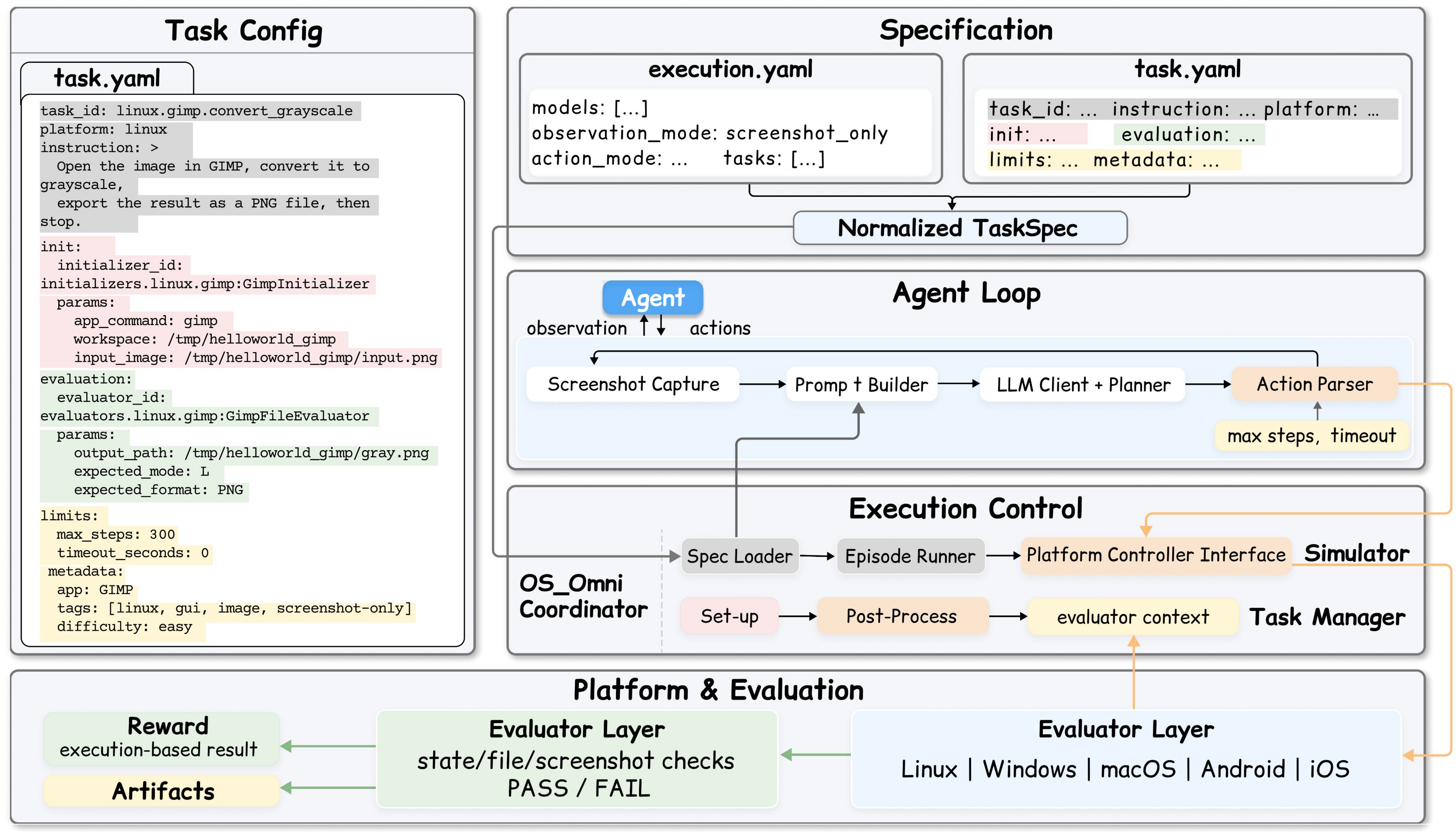}
    \caption{
    Overview of the \dataset environment infrastructure. Execution specifications select tasks, models, observation modes, and action modes; platform runners instantiate the corresponding runtime, run the agent loop, and invoke task-specific evaluators.
    }
    \vspace{-15pt}
    \label{fig:arch_figure}
\end{figure}

To study this setting, we introduce \dataset, a live cross-platform benchmark and execution environment for evaluating computer-using agents across Windows, Ubuntu, macOS, iOS, Android, and web environments. Rather than a fixed task pool, \dataset is released in versioned waves of 300 tasks (v1, v2, \ldots), and each new version includes more mock applications and harder tasks (Section~\ref{sec:live}). This paper focuses on the first release, v1. Each task specifies a natural-language instruction, reproducible initialization, platform-specific adapters, trajectory logging, and an executable final-state evaluator. On v1, we evaluate nine models: GPT-6-Astra, Claude Opus 5.5, Gemini 4 Argon, GLM-5.3-753B, Kimi K3, Qwen 3.8 Max, Qwen 3.8 27B, Qwen 3.8 9B, and DeepSeek V4.1 Flash. The strongest model, Claude Opus 5.5, reaches 68.5\% average success, 8.8 points below the human reference of 77.3\%. Performance is highly uneven across platforms: all three frontier models (GPT-6-Astra, Claude Opus 5.5, and Gemini 4) exceed the human success rate on Linux, yet the best iOS and macOS results (41.7\% and 54.2\%) remain 33.5 and 25.6 points below humans, and the smaller Qwen 3.8 27B and 9B models solve almost no tasks. These results suggest that current agents have not yet learned platform-robust, long-horizon computer use: they can solve many visually explicit tasks, but often fail when success requires sustained state tracking, platform-specific control, and durable final-state changes.

In summary, this paper makes the following contributions:
\begin{itemize}[leftmargin=1.5em,itemsep=0pt,topsep=2pt,parsep=0pt]
    \item We introduce \dataset, a live benchmark for cross-platform computer-using agents released in versioned 300-task waves that progressively add mock applications and harder tasks; its first release, v1, covers six platforms and cross-platform workflows.
    \item We design long-horizon workflows centered on state tracking, cross-application coordination, and final-state correctness.
    \item We build a unified evaluation framework with reproducible initialization, native action backends, trajectory logging, and executable final-state evaluators.
    \item We evaluate nine models on v1, showing that even the best model remains 8.8 points behind humans on average, with gaps of more than 25 points on iOS and macOS.
\end{itemize}

\clearpage
\setcounter{tocdepth}{2}
\begingroup
\hypersetup{linkcolor=black}
\tableofcontents
\endgroup
\clearpage

\section{\dataset Environment}
This section specifies the execution environment behind \dataset, including task specification, environment initialization, agent interaction, and final-state evaluation.

\subsection{Task Definition}
Each task in \dataset specifies a stable identifier, a target platform, a curated instruction, optional initialization, evaluation logic, and descriptive attributes. We adopt a final-state evaluation protocol, where task success depends solely on the terminal environment state:
$R(\tau) = g(s_T)$,
where $\tau=(o_0,a_0,\ldots,o_T)$ denotes the interaction trajectory, $s_T$ is the final environment state, and $g(\cdot)$ is the task-specific evaluator.

At each step, the agent receives the task instruction and the current observation $o_t$, then produces an action $a_t$, which is validated and executed by the platform controller. All cross-platform tasks follow a single-screen setup: an Ubuntu or Windows desktop displays a companion Android device through \texttt{scrcpy}. The agent interacts with both devices through the desktop interface using mouse and keyboard actions, including transferring artifacts between devices when required. This setup exposes a unified interaction surface without requiring separate platform endpoints or host-mediated transfers. An episode terminates when the agent emits \texttt{done} or \texttt{fail}, reaches the step or time limit, or otherwise stops. The final score is determined by the resulting environment state rather than similarity to a reference trajectory.

\subsection{Environment Infrastructure}
\vspace{-5pt}

\dataset is a scalable and executable environment for task design, evaluation, and interactive agent learning. It is organized as a shared orchestration layer plus platform-specific adapters covering most mainstream operating systems (\textit{e.g.}, Windows, macOS, Ubuntu, Android, iOS).

\subsubsection{Environment Initialization}
\label{init_setup}
Real-world scenarios often begin with an open document, a seeded mailbox, or a transfer object that must move between machines. \dataset aims to simulate these intermediate states to replicate real-world computer-using scenarios. Initialization procedures can be generic or application-specific. Generic desktop setup includes clearing stale processes, preparing assets, writing required data, launching applications, and arranging windows. Application-specific procedures seed app databases, configure local mail accounts, prepare mobile app storage, install or copy support assets, and connect benchmark apps to local services. The same task format can therefore describe both simple single-app starts and long multi-app workflow states. 

State reset is implemented differently on each platform. Linux can revert a clean VM snapshot, Windows can restore golden VM storage, macOS can clone from a baseline Tart VM, iOS can clone a simulator, and Android can reset app and device state through ADB-backed preparation. These reset mechanisms are part of the environment contract because GUI agents routinely modify files, accounts, settings, and application caches. \dataset first restores a platform baseline, then materializes task-specific data, and finally arranges the applications that the agent will observe.

\subsubsection{Task Evaluation}
\label{eval}
Every benchmark task declares an evaluator and its parameters. After the agent loop ends, the platform runner invokes this evaluator against the final environment state and records a structured result. This result includes success status and may include diagnostic information such as failure messages, extracted values, saved outputs, and step counts. As shown in Figure~\ref{fig:arch_figure}, \dataset uses execution-based evaluation, which includes evaluators that parse office documents, inspect application databases, check external storage, and validate domain-specific application state. Explicit evaluators also make benchmark reliability auditable. \dataset records trajectories and evaluation records, and also includes oracle variants of tasks for debugging.

\begin{figure}[t]
    \centering
    \makebox[\textwidth][c]{%
        \includegraphics[width=1\textwidth]{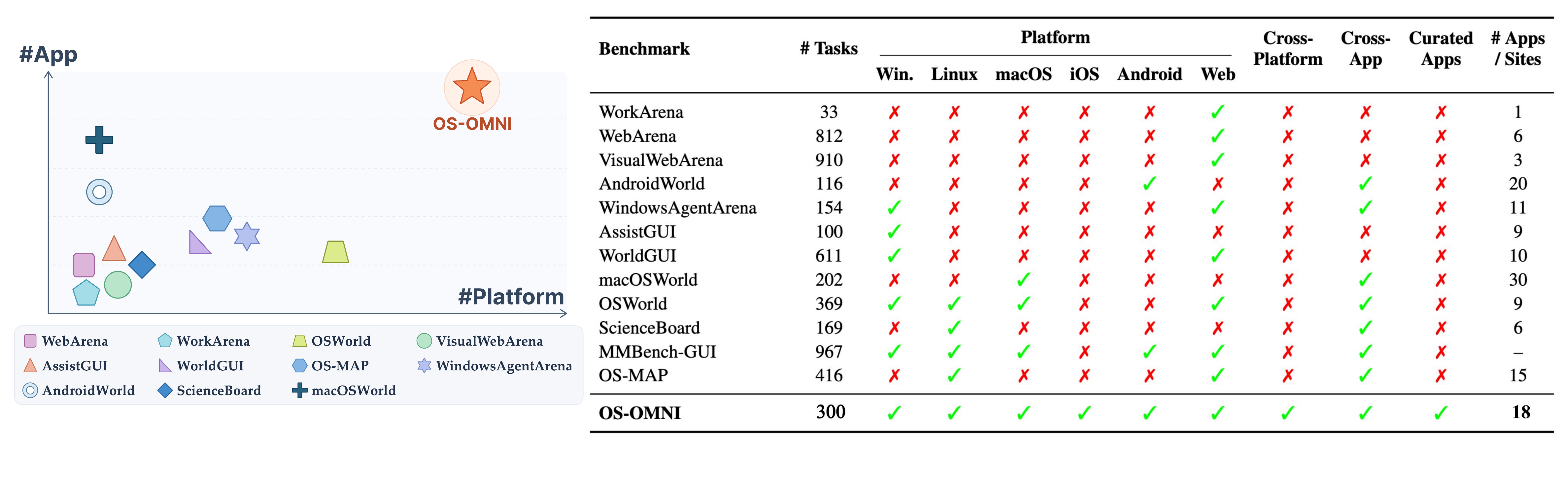}
    }
    \vspace{-3em}
    \caption{
\#Platform indicates the number of supported platforms among Windows, Linux, macOS, iOS, Android, and Web, and \#App indicates the number of covered applications or websites. 
Compared with existing benchmarks, \dataset provides broader platform coverage, richer application diversity, cross-platform workflows, and curated applications.
}
    \vspace{-15pt}
    \label{fig:comparison_benchmarks}
\end{figure}

\subsection{Observation Space}
\label{observation_space}

\dataset supports three observation modes through the execution specification: screenshot-only, accessibility tree-only, and combined observation. We adopt screenshot-only observation as the main experimental setting because screenshots are available across all supported platforms and provide the most uniform interface for generalist computer-using agents. This choice is also aligned with the recent trend of vision-centric CUA agents, where agents increasingly operate from screenshots to improve generality across heterogeneous applications and operating systems~\citep{cheng2024seeclick,wang2024mobile,lu2024omniparser,lin2025showui}. 
Accessibility trees, while useful when available, are platform- and application-dependent: their metadata can be incomplete or unavailable. Therefore, all experiments reported in this paper use screenshot-only observation unless otherwise stated. At the same time, \dataset also supports accessibility tree-only and combined observation modes, enabling future studies to evaluate agents that rely on structured UI metadata or combine visual and accessibility-based information.

\subsection{Action Space}
\label{action_space}


In \dataset, we introduce a two-level action space that combines a basic, platform-agnostic action interface with customized platform execution. The basic layer covers common human-computer operations, including pointer actions, keyboard shortcuts, text entry, scrolling, dragging, touch-style gestures, waiting, and terminal decisions \texttt{DONE} and \texttt{FAIL}. This layer follows the widely used PyAutoGUI-style semantics for mouse and keyboard control, making the agent-facing syntax compact and familiar. The customized layer maps each high-level action intent to the most reliable native backend available on each endpoint. For example, a \texttt{click} action may be executed as a PyAutoGUI desktop event on Linux, a Windows or macOS runtime call, an ADB tap on Android, or an Appium/XCUITest action on iOS. This separation between action syntax and execution backend allows \dataset to evaluate agents under a shared interface without forcing heterogeneous operating systems to expose the same automation API. As a result, the action space is broad enough to cover realistic GUI workflows across desktop and mobile platforms, while remaining structured enough for cross-platform comparison.


\section{\dataset Benchmark}
We introduce the \dataset benchmark, a live benchmark of real-world computer use tasks executed across Linux, Windows, macOS, Android, iOS, Web, and cross-platform workflows (Section~\ref{sec:live}).~\footnote{Due to copyright issues, Windows tasks require activation by the user, while macOS and iOS tasks require Apple hardware.} Each task includes a natural-language goal, a reproducible initialization procedure, and a final-state evaluator.

\subsection{Operating System and Software Environments}

\dataset supports all mainstream operating systems, including \textbf{\textit{desktop operating systems (Windows\windowsicon, macOS\macicon, Ubuntu\linuxicon)}}, \textbf{\textit{mobile operating systems (iOS\iosicon, Android\androidicon)}}, \textbf{{web (Chrome\chromeicon)}}, and \textbf{\textit{workflows that cross platforms}} (Figure~\ref{fig:platform-distribution}). All platforms share the same specification format, while each platform retains the runtime mechanisms required by its own software stack. As shown in Figure~\ref{fig:task-composition}, \dataset covers the most common categories of applications used in everyday digital life. 


\subsection{Custom Software for Controlled Application Coverage}
\label{sec:custom_software}

Real-world CUA workflows frequently involve commercial services or specialized applications that are difficult to reproduce in benchmark environments. We address these coverage gaps by developing a suite of 18 custom applications, collectively referred to as the \texttt{Open} series. Notably, these applications support \textbf{180 of the 300 tasks (60.0\%)} in \dataset v1, making controlled software development a central component of our benchmark design. We organize these applications into two categories; Appendix~\ref{app:open_series} describes their design and deployment and lists the tasks that use each of them.

\paragraph{Controlled Cross-Platform Applications.}
The first category targets common consumer and content-oriented workflows whose real-world counterparts often depend on user accounts, payments, or unstable third-party services. We develop nine applications: \texttt{OpenNote} (social note sharing, including posting, liking, bookmarking, and commenting), \texttt{OpenShop} (e-commerce, including shopping carts, checkout, and reviews), \texttt{OpenTwitter} (social networking), \texttt{OpenBank} (online banking, transfers, and transaction disputes), \texttt{OpenDriver} (ride-hailing with passenger and driver interfaces), \texttt{OpenTrip} (flight and hotel booking), \texttt{OpenFood} (food delivery, coupons, and customer support), \texttt{OpenNavigation} (route planning, waypoints, and incident reporting), and \texttt{OpenPhotoEdit} (image editing and export). These applications provide realistic GUIs across multiple platforms while keeping task data and backend state under benchmark control, enabling reproducible evaluation without relying on live commercial services or real user accounts.

\paragraph{Web-Based Applications for Software Coverage Gaps.}
The second category targets specialized software domains that are unevenly supported across operating systems. We develop nine browser-based applications hosted in the Linux environment: \texttt{OpenDoc} (document editing), \texttt{OpenSheet} (spreadsheets), \texttt{OpenSlide} (presentations), \texttt{OpenMail} (email and contact management), \texttt{OpenFinance} (personal finance, budgeting, and portfolio management), \texttt{OpenCode} (code editing, debugging, and version control), \texttt{OpenBlender} (3D scene editing), \texttt{OpenWireshark} (network packet analysis), and \texttt{OpenZotero} (reference management). By reproducing these workflows in controlled browser environments, we extend task coverage beyond the software available in the underlying operating systems while maintaining a consistent evaluation interface.

Together, these applications enable \dataset to evaluate a broader range of realistic CUA workflows under reproducible conditions, reducing the dependence of benchmark coverage on external services and platform-specific software availability.

\subsection{Live Benchmark and Versioned Releases}
\label{sec:live}
\dataset is a live benchmark rather than a fixed task pool. As agents improve, a static task set gradually saturates and its tasks may leak into training data, so a single release cannot measure frontier agents for long. \dataset therefore releases tasks in versioned waves of 300 tasks each, starting with v1; later waves are published as v2, v3, and so on. Each release is a fixed task set, and results are reported together with the release version so that comparisons are always made on the same tasks.

Each new version is designed to be more demanding than the previous one in two ways. First, it includes more mock applications, i.e., the controlled applications described in Section~\ref{sec:custom_software}, so that more service-backed workflows, such as banking, shopping, travel booking, and social media, can be evaluated without live commercial services or real user accounts. The first release, v1, already draws 180 of its 300 tasks from these Open-series applications. Second, each version makes the tasks harder, for example through longer horizons, more applications or devices per workflow, and more state that must be tracked and left correct at the end. Task specification, initialization, and final-state evaluation follow the same protocol across versions, so the benchmark can keep pace with agent progress without changing how tasks are run or scored. This paper introduces and evaluates the first release, v1.

\subsection{Tasks}
\vspace{-5pt}
The first release, \dataset v1, consists of 300 curated CUA tasks: 48 tasks on each of six platforms, \textbf{\textit{macOS}}, \textbf{\textit{Windows}}, \textbf{\textit{Ubuntu}}, \textbf{\textit{iOS}}, \textbf{\textit{Android}}, and \textbf{\textit{Web}}, plus 12 \textbf{\textit{cross-platform}} workflows. Web tasks run in Chrome or Chromium on the Ubuntu VM. 
Each cross-platform workflow runs on a single screen: the Ubuntu or Windows desktop shows a companion Android phone in a scrcpy window, and the agent works on both devices, moves a file between them, and returns results to the phone (six Ubuntu + Android and six Windows + Android workflows). Each task is represented by a specification that names the platform, instruction, initialization, and evaluator. Overall, the dataset construction was completed by 30 computer science graduate students over two months, with approximately 2,400 person-hours devoted to annotation and an additional 400 person-hours devoted to double-checking.

\subsection{Task Annotation Pipeline}
\label{sec:annotation_pipeline}

\begin{figure*}[t]
\centering
\includegraphics[width=\textwidth]{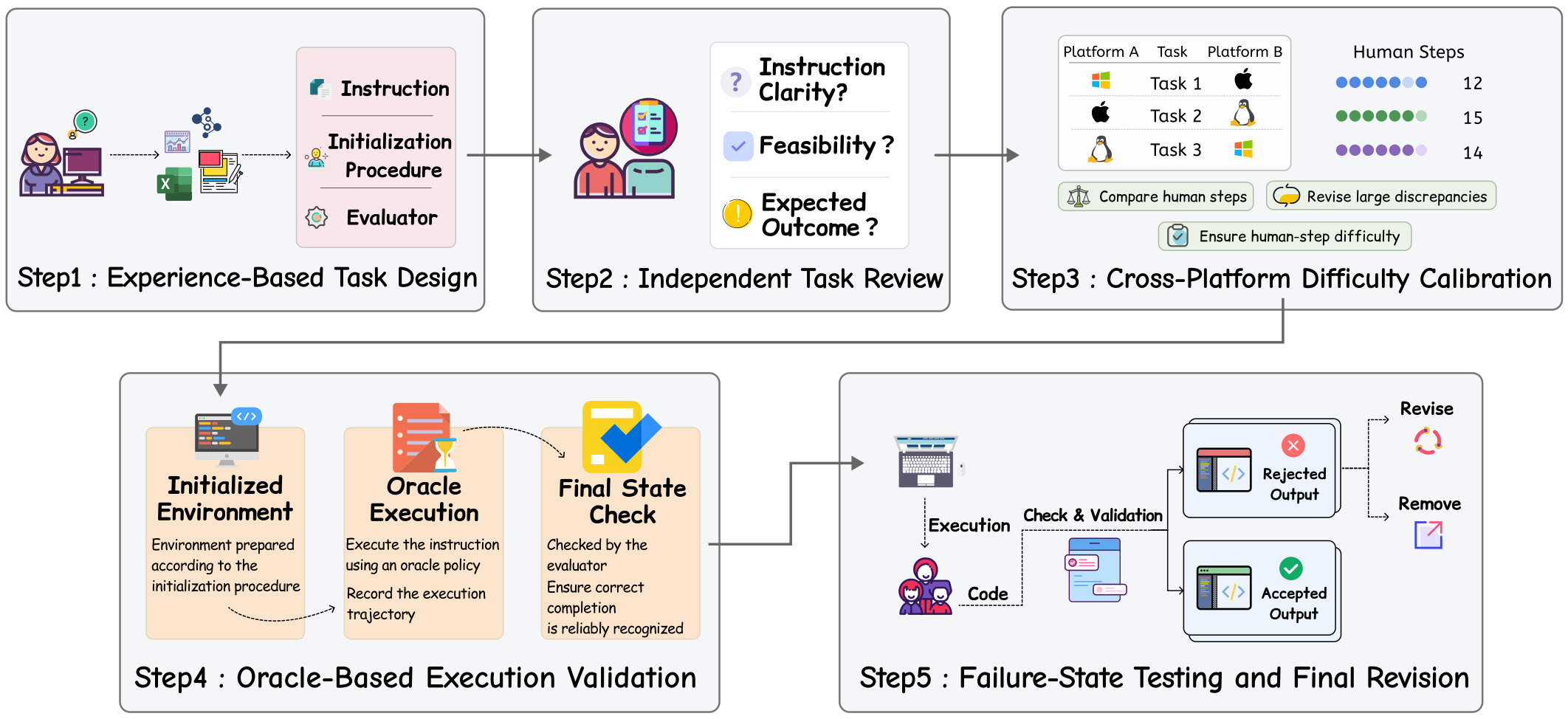}
\caption{The five-stage task annotation pipeline for \dataset.}
\vspace{-15pt}
\label{fig:annotation}
\end{figure*}

As shown in Figure~\ref{fig:annotation}, we use a five-stage annotation pipeline:

(1) \textbf{Experience-Based Task Design.} Each task is constructed by an annotator with hands-on experience using the target software. The annotator designs a natural-language instruction together with the initialization procedure and evaluator. This stage ensures that the task is grounded in realistic software usage rather than being limited to basic operations.
(2) \textbf{Independent Task Review.} A second annotator examines each task for feasibility and ambiguity. This review checks whether the instruction is understandable, whether the task can be completed under the initialized environment, and whether the expected outcome is sufficiently well-defined for reliable evaluation.
(3) \textbf{Cross-Platform Difficulty Calibration.} Annotators then calibrate task difficulty across platforms by comparing the expected number of human operation steps required to complete each task. Tasks are revised when necessary to reduce large difficulty discrepancies, helping ensure that tasks from different platforms have comparable difficulty.
(4) \textbf{Oracle-Based Execution Validation.} We use oracle trajectories to validate both the initializer and the evaluator. Specifically, the oracle execution is run from the initialized environment to verify that the task is solvable under the provided state. The resulting final state is then checked by the evaluator to ensure that a correct completion is reliably recognized.
(5) \textbf{Failure-State Testing and Final Revision.} To further test evaluator robustness, we examine representative failed cases and verify that the evaluator rejects incorrect solutions. Tasks whose initializer, instruction, difficulty level, or evaluator fails any stage of the checking process are revised through additional inspection or removed from the benchmark.

\subsection{Data Statistics}

\begin{figure*}[t]
    \centering
    \includegraphics[width=\textwidth]{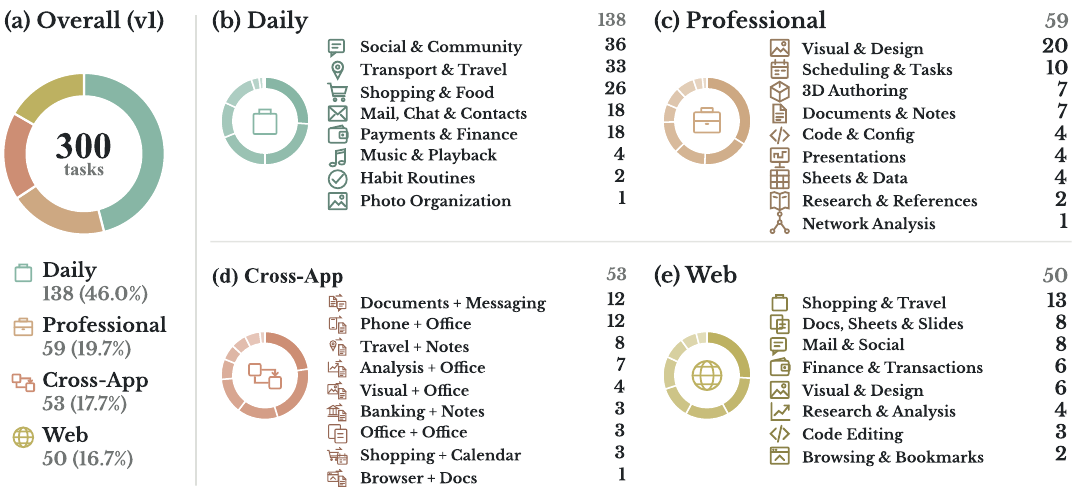}
    \caption{Task composition of \dataset v1.
    (a) Overall distribution across task categories.
    (b--e) Secondary-category distributions within Daily,
    Professional, Cross-App, and Web tasks, respectively.
    Numbers denote task counts.}
    \label{fig:task-composition}
\end{figure*}

\begin{figure}[t]
    \centering
    \includegraphics[width=\columnwidth]{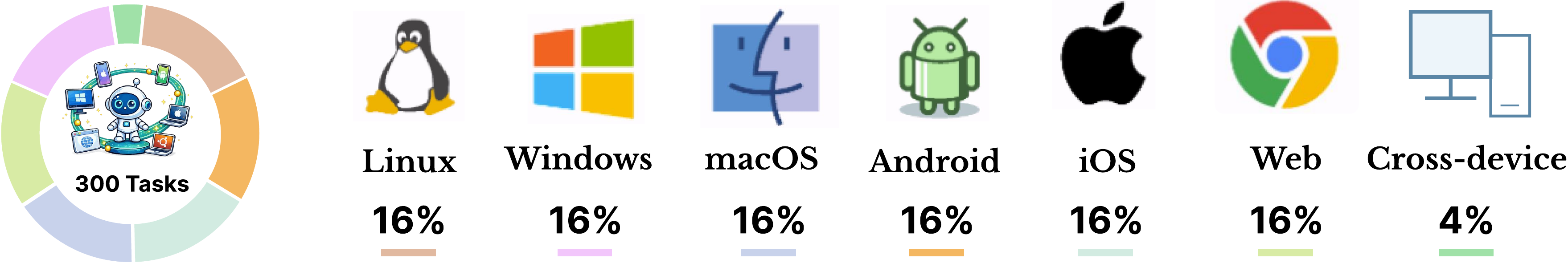}
    \caption{Distribution of the 300 tasks across evaluation
    tracks. Windows, macOS, Ubuntu, iOS, Android, and Web each
    contain 48 tasks (16\%), while the cross-device track
    contains 12 workflows (4\%).}
    \label{fig:platform-distribution}
\end{figure}

The main statistics of \dataset are summarized in
Figures~\ref{fig:task-composition}
and~\ref{fig:platform-distribution}.
As shown in Figure~\ref{fig:platform-distribution},
\dataset v1 contains 300 tasks: 48 tasks on each of the
six platforms (16\% each) and 12 cross-device workflows (4\%).
Figure~\ref{fig:task-composition} presents the task taxonomy,
which comprises four primary categories: daily-life tasks
(138; 46.0\%), professional productivity tasks (59; 19.7\%),
cross-application workflows (53; 17.7\%), and web tasks
(50; 16.7\%), further organized into 34 secondary categories.
Each task is assigned to exactly one primary category,
with cross-application workflows taking precedence.
The Web category includes the 48 Web-track tasks and two
Safari browsing tasks from the iOS track.

\subsection{Comparison with existing benchmarks}
Figure~\ref{fig:comparison_benchmarks} compares \dataset with representative existing benchmarks in terms of platform coverage, task scale, application diversity, cross-platform workflows, cross-application workflows, and curated applications. WorkArena~\citep{drouin2024workarena}, WebArena~\citep{zhou2023webarena}, and VisualWebArena~\citep{koh2024visualwebarena} evaluate agents in web environments. AndroidWorld~\citep{rawles2024androidworld} focuses on Android. OSWorld~\citep{xie2024osworld}, WindowsAgentArena~\citep{bonatti2024windows}, AssistGUI~\citep{gao2023assistgui}, WorldGUI~\citep{zhao2025worldgui}, macOSWorld~\citep{yang2025macosworld}, ScienceBoard~\citep{sun2025scienceboard}, MMBench-GUI~\citep{wang2025mmbench}, and OS-MAP~\citep{chen2025map} extend evaluation to desktop or multi-platform GUI settings, but they do not provide the same level of coverage across desktop, mobile, web, and cross-platform workflows.

In contrast, \dataset unifies Windows, Linux, macOS, iOS, Android, and Web environments under a single benchmark setting. It further introduces benchmark-controlled applications and services to support workflows involving persistent user state, backend records, shopping carts, orders, and payment-like options. Compared with benchmarks that rely on public websites, external accounts, or preloaded local profiles, this design enables reproducible initialization and reliable final-state evaluation without depending on live commercial platforms or real user accounts. As a result, \dataset evaluates agents on realistic service-oriented workflows that span multiple applications and platforms, rather than isolated GUI operations alone.
\section{Experiments}

\subsection{VLM and Computer-Using Agent Baselines}
We conduct human evaluations with annotators who possess basic software usage skills but have not been exposed to the tasks before. For each human trajectory, we record the number of interaction steps and the wall-clock completion time, and aggregate the measurements by platform.
Then, we evaluate nine models on \dataset v1 (Section~\ref{sec:live}): GPT-6-Astra, Claude Opus 5.5, Gemini 4 Argon, GLM-5.3-753B, Kimi K3, Qwen 3.8 Max, Qwen 3.8 27B, Qwen 3.8 9B, and DeepSeek V4.1 Flash. By default, agents receive the current screenshot and the three most recent interaction steps and screenshots. Agents complete each task within the limits declared in its specification: a maximum of 300 steps and a per-step time limit of 180 seconds for single-platform tasks, and 300--650 steps with a per-step time limit of 120 seconds for cross-platform workflows. We present a summary of the results in Table~\ref{tab:main_results}.


\begin{table*}[t]
\centering
\scriptsize
\setlength{\tabcolsep}{3pt}
\renewcommand{\arraystretch}{1.18}

\caption{\textbf{Main results on \dataset v1.}
Each cell reports task success rate (\%) on one platform
or on the cross-platform workflows.
\textbf{Macro-Avg.} is the mean over all seven columns
(the six platforms and Cross-Platform).
\textbf{Micro-Avg.} weights every task equally:
it is the share of all 300 v1 tasks solved
(288 single-platform tasks and 12 cross-platform workflows);
for Human Performance it is the task-weighted mean
of the column rates.
Among models, the best result in each column is in
\textbf{bold} and the second best is \underline{underlined}.}

\label{tab:main_results}

\resizebox{\textwidth}{!}{%
\begin{tabular}{>{\raggedright\arraybackslash}p{3.1cm}*{9}{c}}

\toprule

\multirow{2}{*}{\textbf{Model / Agent}}
& \multicolumn{9}{c}{\textbf{Success Rate ($\uparrow$)}} \\

\cmidrule(lr){2-10}

& \windowsicon\,\textbf{Windows}
& \linuxicon\,\textbf{Linux}
& \macicon\,\textbf{macOS}
& \iosicon\,\textbf{iOS}
& \androidicon\,\textbf{Android}
& \chromeicon\,\textbf{Web}
& \textbf{Cross-Platform}
& \textbf{Macro-Avg.}
& \textbf{Micro-Avg.} \\

\midrule


\textbf{Human Performance}
& 78.4
& 74.6
& 79.8
& 75.2
& 72.6
& 76.9
& 83.3
& \cellcolor[HTML]{DFF3E4}77.3
& 76.5 \\

\midrule


\raisebox{-0.18\height}{%
  \includegraphics[height=1.05em]{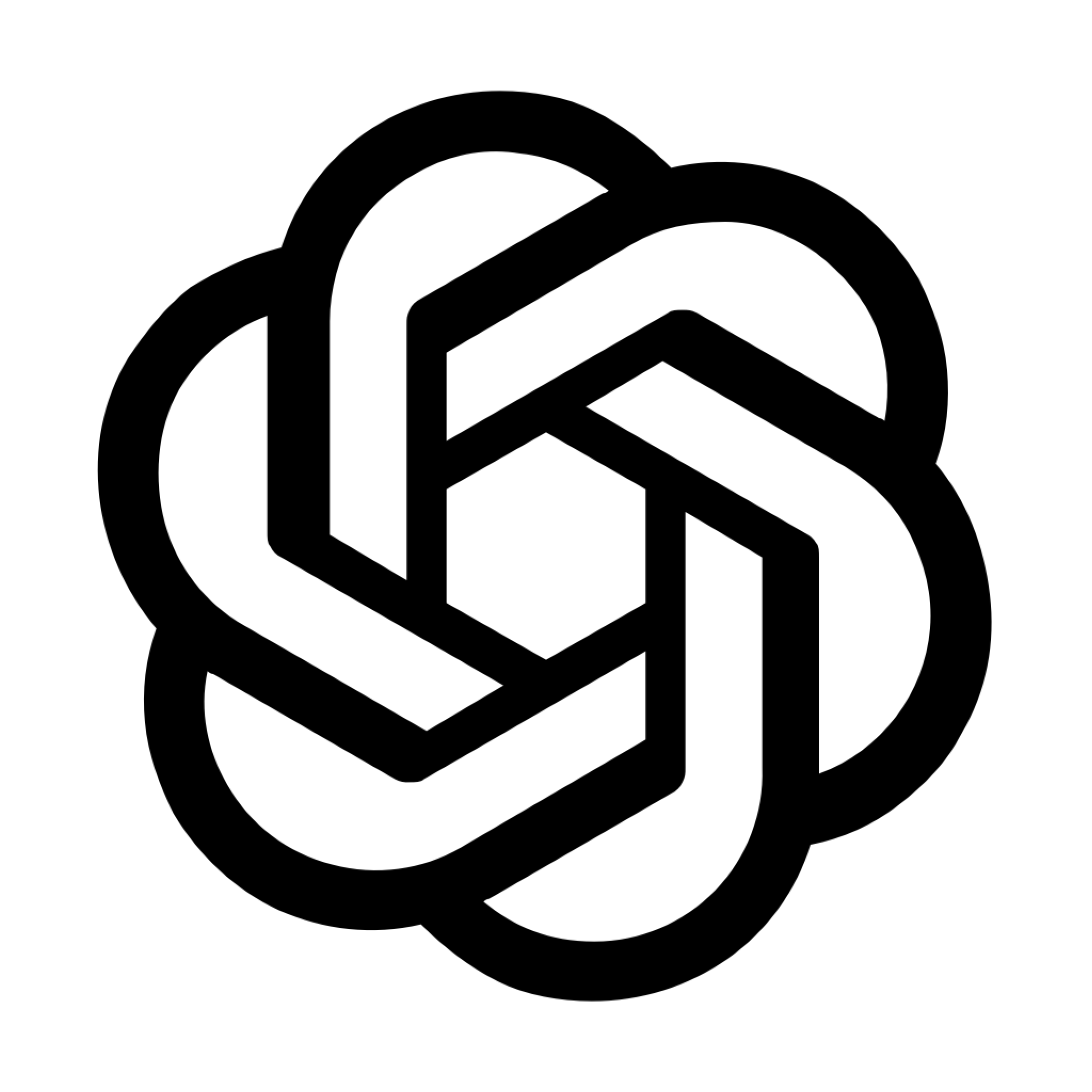}}\,GPT-6-Astra
& \textbf{66.7}
& \underline{89.6}
& \underline{50.0}
& \underline{33.3}
& \underline{77.1}
& \textbf{87.5}
& \underline{58.3}
& \cellcolor[HTML]{F0F7E6}\underline{66.1}
& \textbf{67.0} \\

\raisebox{-0.18\height}{%
  \includegraphics[height=1.05em]{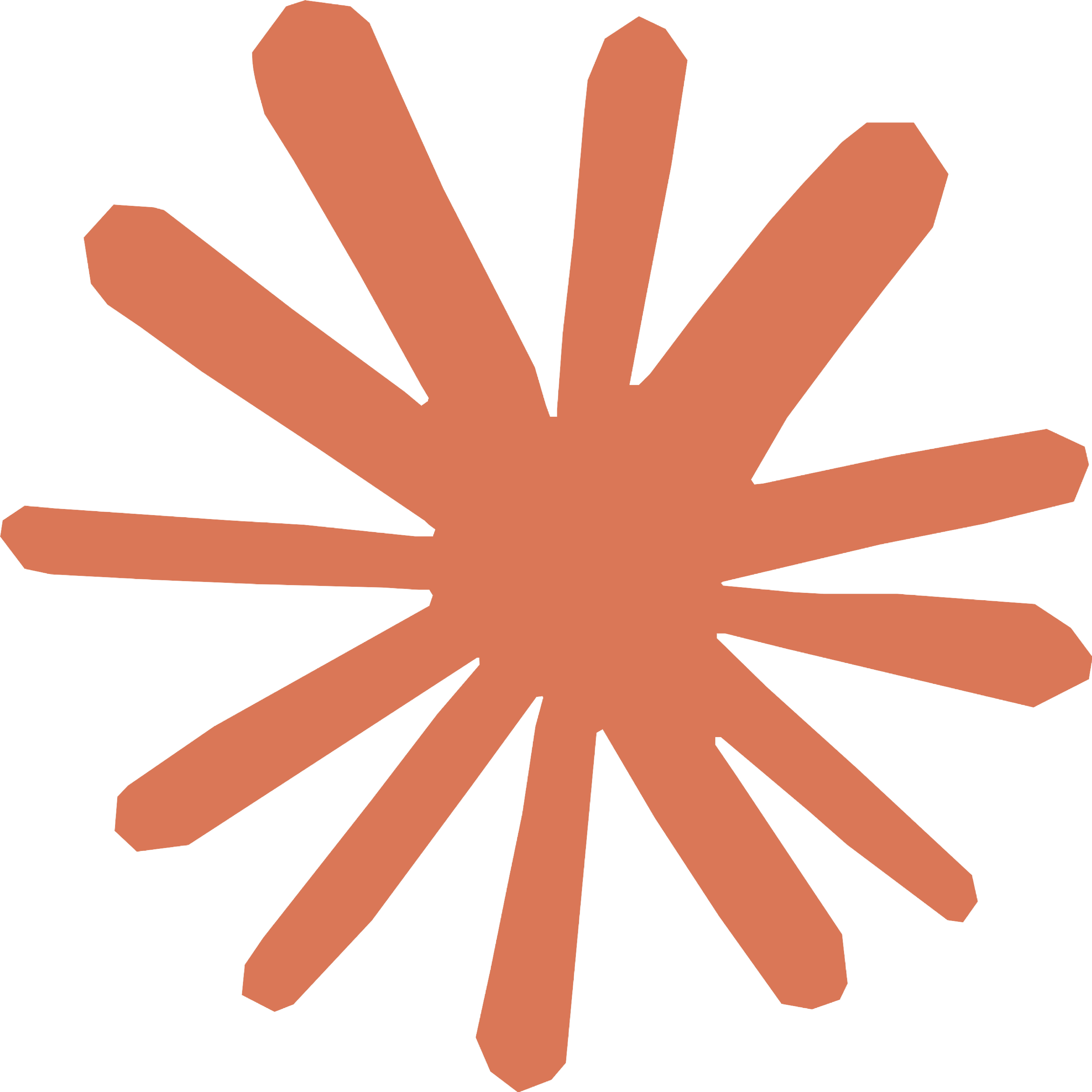}}\,Claude Opus 5.5
& \underline{62.5}
& 85.4
& \textbf{54.2}
& \textbf{41.7}
& \textbf{81.3}
& 70.8
& \textbf{83.3}
& \cellcolor[HTML]{E7F6EA}\textbf{68.5}
& \underline{66.7} \\

\raisebox{-0.18\height}{%
  \includegraphics[height=1.05em]{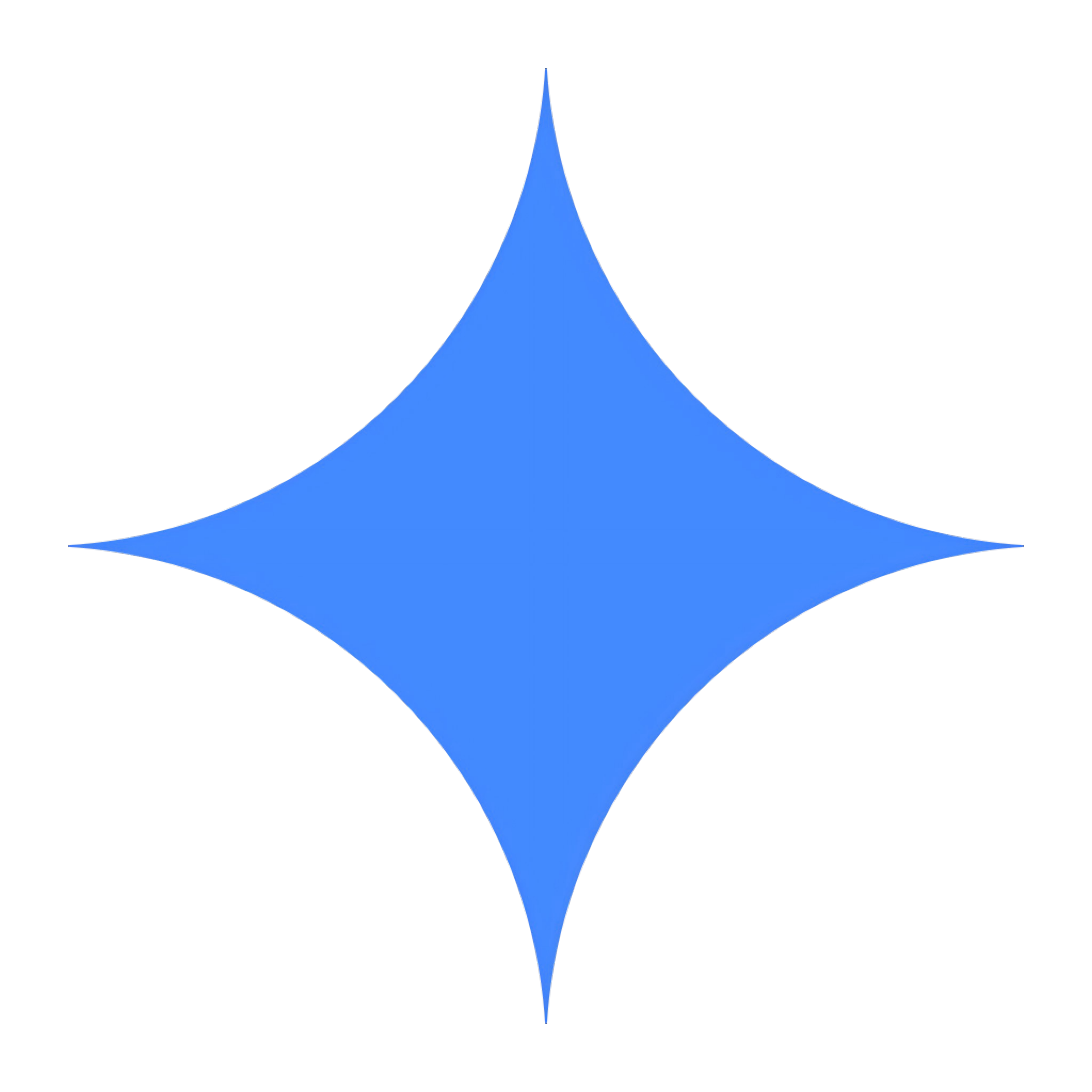}}\,Gemini 4 Argon
& 58.3
& \textbf{95.8}
& 39.6
& 18.8
& 66.7
& \underline{81.3}
& 50.0
& \cellcolor[HTML]{FFF6D8}58.6
& 59.7 \\


\raisebox{-0.18\height}{%
  \includegraphics[height=1.05em]{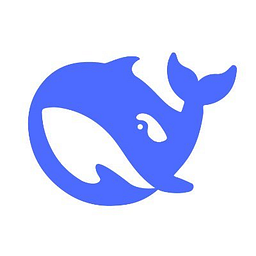}}\,DeepSeek V4.1 Flash
& 22.9
& 62.5
& 16.7
& 12.5
& 41.7
& 22.9
& 33.3
& \cellcolor[HTML]{FDEAD2}30.4
& 30.0 \\

\raisebox{-0.18\height}{%
  \includegraphics[height=1.05em]{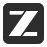}}\,GLM-5.3-753B
& 6.3
& 68.8
& 8.3
& 10.4
& 8.3
& 45.8
& 16.7
& \cellcolor[HTML]{FBE4D3}23.5
& 24.3 \\

\raisebox{-0.18\height}{%
  \includegraphics[height=1.05em]{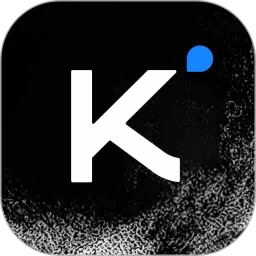}}\,Kimi K3 2.8T
& 20.8
& 50.0
& 4.2
& 12.5
& 16.7
& 25.0
& 25.0
& \cellcolor[HTML]{FBE3D3}22.0
& 21.7 \\

\raisebox{-0.18\height}{%
  \includegraphics[height=1.05em]{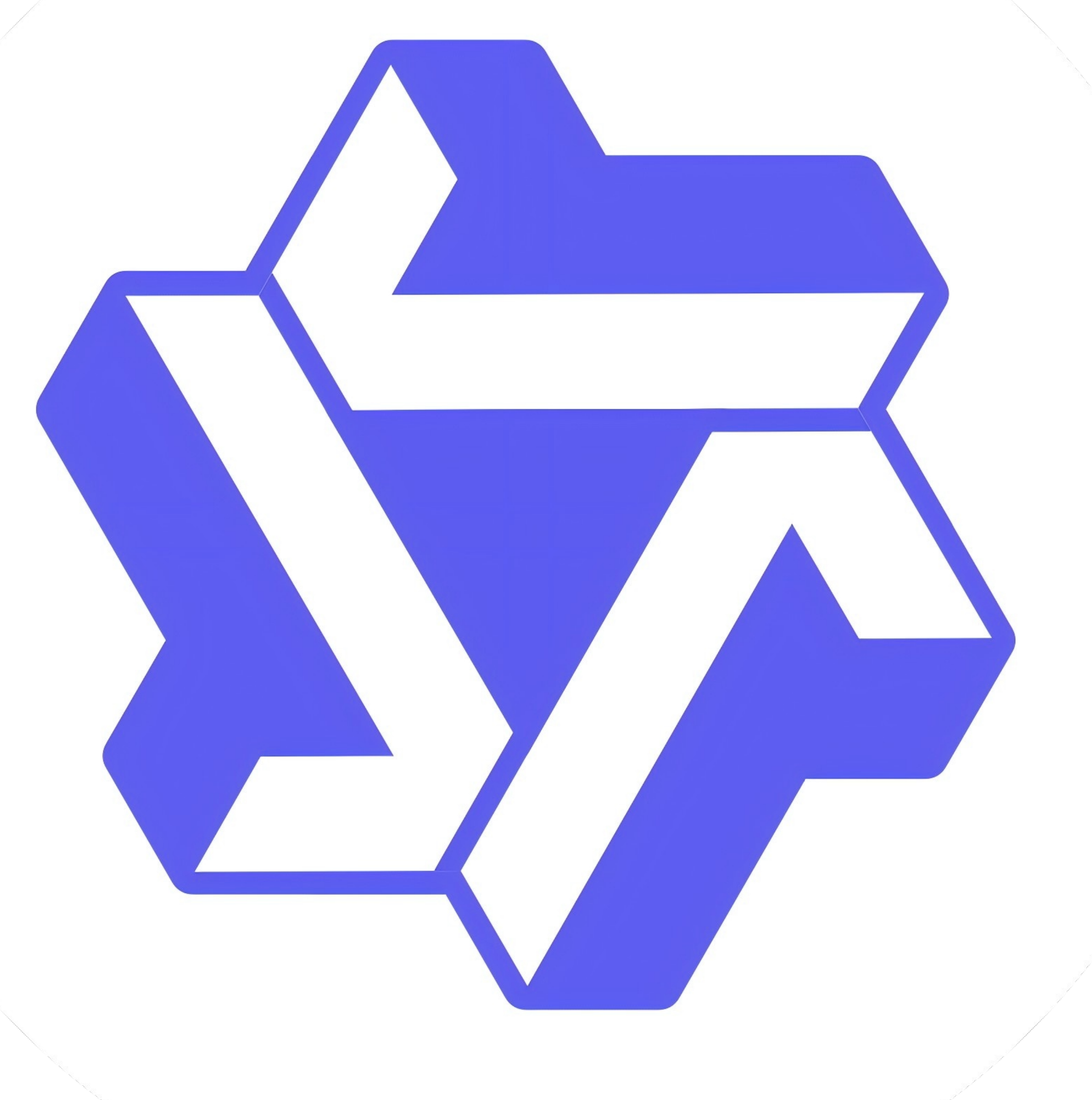}}\,Qwen 3.8 Max
& 6.3
& 6.3
& 4.2
& 12.5
& 31.3
& 6.3
& 25.0
& \cellcolor[HTML]{F9DCD5}13.1
& 11.7 \\

\raisebox{-0.18\height}{%
  \includegraphics[height=1.05em]{figures/icons/qwen.pdf}}\,Qwen 3.8 27B
& 0.0
& 0.0
& 2.1
& 6.3
& 0.0
& 0.0
& 0.0
& \cellcolor[HTML]{F5C9C5}1.2
& 1.3 \\

\raisebox{-0.18\height}{%
  \includegraphics[height=1.05em]{figures/icons/qwen.pdf}}\,Qwen 3.8 9B
& 0.0
& 0.0
& 0.0
& 0.0
& 0.0
& 0.0
& 0.0
& \cellcolor[HTML]{F2B8B5}0.0
& 0.0 \\

\bottomrule

\end{tabular}
}

\end{table*}

\subsection{Main Results}

\noindent
\textbf{Even the strongest agents remain below humans, and most models are far behind.}
Claude Opus 5.5 obtains the best Macro-Avg. success rate in Table~\ref{tab:main_results}, 68.5\%, 8.8 points below the 77.3\% human average, followed by GPT-6-Astra (66.1\%) and Gemini 4 Argon (58.6\%). Under Micro-Avg., which weights every task equally, GPT-6-Astra and Claude Opus 5.5 are nearly tied (67.0\% and 66.7\%, i.e., 201 and 200 solved tasks): the Macro-Avg. gives the 12 cross-platform workflows, where Claude Opus 5.5 solves three more tasks, the same weight as each 48-task platform. The remaining models fall far behind: DeepSeek V4.1 Flash reaches 30.4\%, Qwen 3.8 Max 13.1\%, and Qwen 3.8 27B and 9B solve almost no tasks (1.2\% and 0.0\%). These results indicate that current agents can solve a portion of visually explicit and short-horizon tasks, but still lack the robustness needed for long-horizon planning, precise GUI grounding, state tracking, and final-state persistence in real operating systems.

\noindent
\textbf{Humans are consistent across platforms, while agents are not.}
The human reference row is consistent across all platforms (72.6--83.3\%), whereas agent performance varies sharply by operating system. Gemini 4 Argon reaches 95.8\% on Linux but only 18.8\% on iOS, and even Claude Opus 5.5 ranges from 85.4\% on Linux to 41.7\% on iOS. On Linux, all three frontier models exceed the human success rate of 74.6\%, while on iOS and macOS the best model remains 33.5 and 25.6 points below humans. Similar drops appear for the other models, suggesting that the main challenge is not only performing well on a specific platform, but also succeeding consistently on every platform.


\subsection{Analysis}

\noindent
\textbf{Cross-application workflows are harder than single-application commands.}
Cross-application tasks add a second layer of difficulty: the agent must preserve information across window switches, identify which application owns each subgoal, and ensure that all destination states are updated before stopping. The failures are also qualitatively different: agents often complete a visually plausible intermediate action, but fail the final evaluator because a calendar event or follow-up message is missing or inconsistent. This makes cross-application evaluation closer to real-world computer use, where user intent is rarely confined to a single GUI command.

\noindent
\textbf{Cross-platform workflows separate the strongest model from the rest.}
Cross-platform tasks require correct use of platform-specific interaction conventions and reliable transfer of artifacts between devices. In single-platform tasks, the agent can remain within one action backend and one set of visual conventions. The v1 cross-platform workflows instead require the agent to edit or build a file on the Android phone, move it to the desktop itself, build the deliverable in LibreOffice, and send results computed on the desktop back to the phone. On these workflows, Claude Opus 5.5 matches the human success rate of 83.3\%, while GPT-6-Astra and Gemini 4 reach 58.3\% and 50.0\%, below their averages over the six single platforms, and Qwen 3.8 27B and 9B solve none. Rankings also shift: Qwen 3.8 Max and DeepSeek V4.1 Flash score higher on this split (25.0\% and 33.3\%) than on most single platforms. Because the split contains only 12 workflows, each task changes the success rate by 8.3 points, so these differences should be read with caution. Together with the platform sensitivity observed in Table~\ref{tab:main_results}, this suggests that cross-platform CUA is not simply a larger task set, but a distinct generalization problem involving heterogeneous UI conventions and input primitives.

\noindent
\textbf{Error analysis.}
We further inspect trajectories to categorize common failure modes. The most frequent errors fall into four groups. First, \emph{visual grounding errors} occur when the agent clicks a nearby control or misreads dense document and diagram interfaces. Second, \emph{state-tracking errors} appear when the agent loses copied information across applications or forgets a required destination file. Third, \emph{persistence errors} occur when the visible interface looks correct but the durable final state is not saved or committed in the format expected by the evaluator. Fourth, \emph{platform and timing errors} arise from modal dialogs, delayed application loading or missing confirmation steps. These categories explain why screenshot-only agents can often produce plausible trajectories while still failing the task.

\section{Related Work}

\noindent
\textbf{Benchmarks for GUI Agents.} 
GUI-agent benchmarks have progressed from static web navigation to executable environments. Mind2Web, WebArena, VisualWebArena, WebVoyager, and WorkArena evaluate browser or enterprise web workflows \citep{deng2023mind2web,zhou2023webarena,koh2024visualwebarena,he2024webvoyager,drouin2024workarena}; AndroidWorld focuses on mobile-device interaction \citep{rawles2024androidworld}; and OSWorld, WindowsAgentArena, WorldGUI, macOSWorld, OS-MAP, and MMBench-GUI study desktop GUI automation \citep{xie2024osworld,bonatti2024windows,zhao2025worldgui,yang2025macosworld,chen2025map,wang2025mmbench}. These benchmarks establish the value of executable evaluation, but most still emphasize one platform family, one dominant interface assumption, or benchmark-specific runtime protocols.

\noindent
\textbf{Vision-language models for GUI Agents.} 
Recent CUA agents are specialized for screen understanding and long-horizon interaction. AppAgent and MobileAgent study smartphone control, while SeeClick, ShowUI, and OmniParser improve visual grounding and screen parsing for GUI operation \citep{zhang2025appagent,wang2024mobile,cheng2024seeclick,lin2025showui,lu2024omniparser}. Agent~S explores open-ended desktop operation with explicit planning and recovery \citep{agashe2024agent}. More recently, OpenCUA and ScaleCUA develop open computer-using agents and training data for cross-platform control \citep{wang2025opencua,liu2025scalecua}. These systems show rapid progress, but their behavior remains sensitive to platform conventions and action interfaces, motivating unified cross-platform evaluation in \dataset.

\section{Conclusion}


We built \dataset to study whether a single computer-using agent can reliably complete long-horizon workflows across desktop, mobile, and web environments. The first release contains 300 tasks, including 12 cross-device workflows, with success determined by the resulting environment state. Future releases will expand application coverage and task complexity in waves of 300 tasks.

Our results reveal two limitations of current computer-using agents. First, strong performance on one platform does not imply generality: all three frontier models outperform humans on Linux, yet the best results on iOS and macOS remain 33.5 and 25.6 percentage points below human performance, respectively (Section 4.2). Second, progress through an interface does not necessarily translate into task completion. In OpenBank, an agent submits the same transfer twice; in OpenTrip, an agent creates five confirmed bookings but duplicates one and omits another (Appendix E).

These findings suggest that generalist computer use requires more than proficiency with individual interfaces. Agents must maintain task state over extended interactions, adapt to different software environments, and verify the effects of their actions. We believe benchmarks for generalist computer use should measure both the breadth of environments an agent can operate in and the correctness of the state it leaves behind.


\begin{ack}
This paper is partially supported by a grant from the Research Grants Council of the Hong Kong Special Administrative Region (Project No. T41-517/25-N). We also acknowledge CompAI for their support of this project, conducted as part of a milestone.
\end{ack}

\clearpage
\bibliography{main}

\clearpage
\appendix
\section{Details of Human Experiments}
\label{app:human_experiment}
\begin{figure}[b]
    \centering
    \begin{subfigure}[t]{0.49\linewidth}
        \centering
        \includegraphics[width=\linewidth]{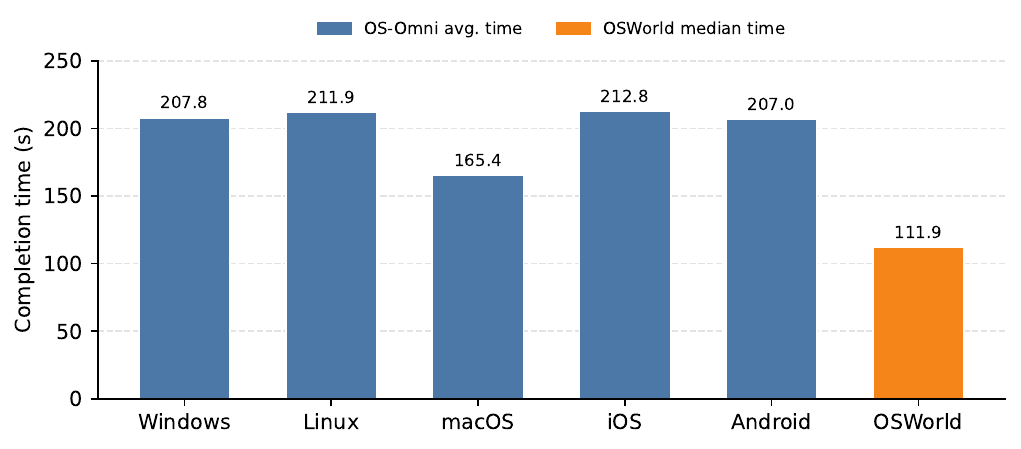}
        \caption{Human completion time on \dataset by platform. Blue bars report the average completion time for each \dataset platform, and the orange bar shows the OSWorld reported median completion time.}
        \label{fig:human_time_comparison}
    \end{subfigure}
    \hfill
    \begin{subfigure}[t]{0.49\linewidth}
        \centering
        \includegraphics[width=\linewidth]{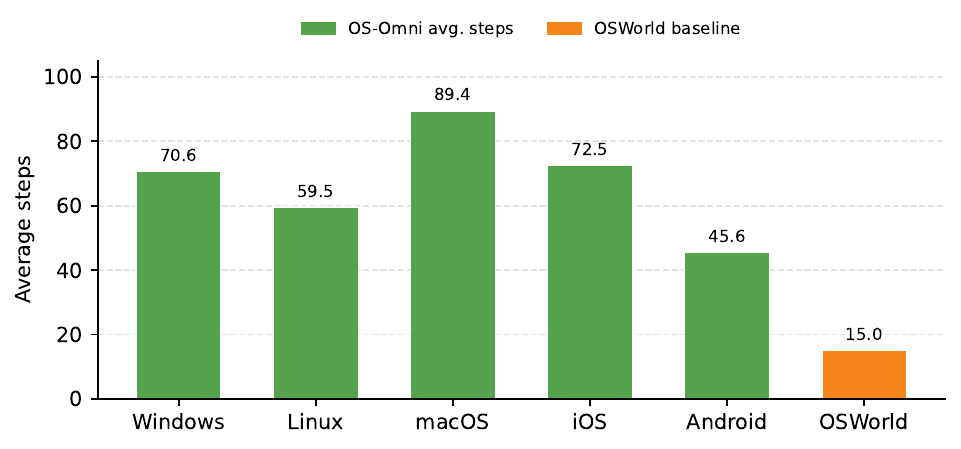}
        \caption{Human interaction steps on \dataset by platform. Bars report the average number of interaction steps needed by human annotators on each platform.}
        \label{fig:human_steps_comparison}
    \end{subfigure}
\end{figure}

As shown in Figures~\ref{fig:human_time_comparison} and~\ref{fig:human_steps_comparison}, human users require substantial effort across all \dataset platforms. Averaged over Windows, Linux, macOS, iOS, and Android, a task takes 67.54 interaction steps and 200.97 seconds to complete. This is 1.80$\times$ the OSWorld reported median completion time of 111.94 seconds~\citep{xie2024osworld}. Even the fastest \dataset platform, macOS, takes 165.41 seconds on average, or 1.48$\times$ the OSWorld reference; iOS, Linux, Windows, and Android cluster between 207.00 and 212.79 seconds, corresponding to 1.85--1.90$\times$ the OSWorld time.

The per-platform measurements also show that task burden is not determined by action count alone. Android has the fewest human steps (45.63) but still requires 207.00 seconds on average, while macOS has the most steps (89.43) but the shortest average time among \dataset platforms. This gap indicates that human effort depends on platform-specific feedback latency, application startup and navigation cost, file or state inspection, and the need to verify persistent artifacts, rather than simply the number of GUI actions. OSWorld further reports 72.36\% human accuracy, compared with 88\% on WebArena~\citep{zhou2023webarena}, showing that executable desktop tasks are already harder than pure web tasks. \dataset extends this difficulty to a broader cross-platform setting, where users must handle heterogeneous operating systems, application ecosystems, and state representations.

\section{Details of \textsc{OS-Omni} Environment}
\label{app:osomni_env}

\subsection{Environment Infrastructure}
\label{app:osomni_env_infra}

\textsc{OS-Omni} supports both containerized and non-containerized deployment. The containerized path packages the host-side runner, dependencies, local services, and orchestration logic into a reproducible software environment, which is convenient for headless servers and shared evaluation machines. The non-containerized path runs the same components directly on the host system, which is useful for local development, debugging, and platform-specific runtime control. In both cases, deployment is separated from the environment seen by the agent.

\textsc{OS-Omni} provides runtime support for desktop, mobile, and web endpoints through controlled execution environments. The agent-facing environments are instantiated through platform-specific runtimes, with the host-side components deployed either inside or outside containers. Desktop endpoints are instantiated through managed virtual machines, mobile endpoints are instantiated through emulators or simulators, and web tasks are instantiated through controlled browser environments on the supported operating systems. Ubuntu and Windows are controlled through the desktop runtime interface, macOS through the Apple-platform runtime interface, Android through ADB, iOS through an XCUITest-based automation backend, and web tasks through browser-specific controllers and local benchmark services. This design preserves each platform's native UI stack, application state, and input behavior while keeping the host-side execution stack reproducible.

Across these backends, the runtime layer exposes the same episode contract. It restores or prepares the guest, applies the task initializer, captures observations, executes actions, retrieves artifacts, and invokes the final-state evaluator. In v1, cross-platform tasks run on a single screen: the Ubuntu or Windows guest shows a companion Android emulator in a scrcpy window, and the evaluator inspects both devices. The earlier cross-platform workflows, which ran two platform episodes in sequence and transferred the required artifact through the host, have been retired from the task set. Every guest is restored or reset from a clean baseline before each task so that initial state is reproducible, and multiple guests can run in parallel on a single host.

\subsection{Observation Space}
\label{app:osomni_obs}

\textsc{OS-Omni} supports three observation modes through the execution specification: screenshot-only, accessibility-tree-only, and combined screenshot-plus-tree observation. The primary protocol used in the main results is screenshot-only, because screenshots are naturally available across desktop, mobile, web, and cross-platform workflows without assuming that every application exposes clean structured UI metadata. Accessibility-tree-only and combined-observation modes are supported for alternative protocols, debugging, and ablation studies. Some runtime backends can also expose terminal output for command-line workflows, but terminal output is an auxiliary channel rather than a main benchmark observation mode. Trajectory artifacts (per-step screenshots, model responses, controller status, and the final evaluator output) are recorded for post-hoc inspection but are not surfaced to the agent at decision time. \textsc{OS-Omni} supports observation refactoring and extension if needed, such as restricting the observation to a focused application or adding a scoped DOM extract.

\subsubsection{Screenshot}
\label{app:osomni_screenshot}

To align with the perception of a human user, we capture a screenshot of the entire device screen at every step. Including the mouse cursor also proves helpful in certain cases where mouse information is crucial. For desktop resolution, we default to $1920 \times 1080$, which offers a 16:9 aspect ratio and matches the most commonly used desktop resolution in worldwide statistics. On mobile, the screenshot is captured at the device's native resolution. \textsc{OS-Omni} also supports modifying the resolution of virtual machines to avoid potential memorization of absolute pixel values and to assist studies on topics like generalization across different resolutions. Screenshots are persisted under per-episode run directories so that a trajectory can be replayed visually and the final state can be audited independently of the evaluator's structured checks.

\subsubsection{Accessibility Tree}
\label{app:osomni_a11y}

An accessibility tree (or a11y tree) refers to an intricate structure generated by the OS or browser accessibility APIs that renders a representative model of the on-screen content, providing a means of interaction for assistive technologies. Each node within the accessibility tree hosts important information about a UI element. This could range from the nature of the object (a button, checkbox, list item, or paragraph of text), its current state (checked or unchecked, focused or hidden), and even its spatial orientation on the screen.

Different operating systems employ varied accessibility APIs to construct and manipulate the accessibility tree. These include UI Automation (UIA) and Microsoft Active Accessibility (MSAA) for Windows, the NSAccessibility / AX protocol for macOS, the Assistive Technology Service Provider Interface (ATSPI) for the GNOME desktop used on Ubuntu, the device view-hierarchy dump for Android, and XCUITest element trees for iOS. \textsc{OS-Omni} normalizes these per-platform sources into a shared per-node schema that carries an identifier, a type, visible text, a selector, the enabled / focused / active flags, and the on-screen bounding rectangle, so that the same agent interface can address an element by either node identifier or pixel coordinate. Because the information returned by these accessibility APIs differs in coverage and noise level across platforms, we treat structured metadata as a diagnostic and ablation channel rather than as a privileged oracle.

\subsubsection{Terminal Output}
\label{app:osomni_terminal}

\textsc{OS-Omni} also supports terminal output as an optional observation channel for command-line workflows. When enabled, the runtime returns the text shown in the active terminal session or the output exposed by the platform command interface, depending on the backend. This channel is useful for tasks involving shell commands, script execution, logs, and text-based tools, where the relevant state may be textual and not fully recoverable from screenshots alone.

Terminal output is not part of the main screenshot-only protocol. We reserve it for compatible environments, debugging, and ablation studies so that the primary reported setting remains aligned with the visual information available across desktop, mobile, web, and cross-platform workflows.

\subsection{Action Space}
\label{app:osomni_action}

We implement two kinds of action space: a free-form PyAutoGUI action space, used as the main interface on desktop platforms, and a structured \emph{computer-style} action schema that is exposed by every platform runner and shared across desktop, mobile, and web tasks. The PyAutoGUI action space saves tokens for describing the action space in prompting and matches the OSWorld baselines; the structured schema is finite and parameterized, which makes it more suitable for tool-calling APIs, constrained decoding, and reinforcement-learning-style training. Both interfaces share the same episode lifecycle and the same final-state evaluator, so they are experimental choices rather than separate benchmarks. \texttt{DONE} and \texttt{FAIL} are terminal actions shared by both interfaces. \texttt{WAIT} allows the interface to settle before the next observation without terminating the episode and counts toward the step budget. Episodes may also terminate upon budget exhaustion or an unrecoverable runtime condition.

\subsubsection{PyAutoGUI}
\label{app:osomni_pyautogui}

\texttt{pyautogui} is an open-source, cross-platform Python module utilized for programmatically controlling the mouse and keyboard. It controls cursor movement, single and multi-button clicks, drags, scroll wheel events, and keyboard input. We adopt PyAutoGUI as our core desktop control component, and we use it as the main agent-facing action space on desktop platforms because it tokenizes compactly, generalizes across Ubuntu/Linux, Windows, and macOS, and matches the style of code used in prior baselines.

At each step, the agent is asked to first give a short reflection on the current screenshot and previous actions, and then to return one fenced Python block that contains the next action(s). Specifically, it may also return one of three special blocks: \texttt{WAIT} when it needs the application to settle, \texttt{FAIL} when it determines the task cannot be performed, or \texttt{DONE} when it determines that the task is complete. The on-guest sandbox only allows \texttt{import pyautogui}, \texttt{import time}, direct \texttt{pyautogui} calls, simple loops, and \texttt{print}; modules such as \texttt{os}, \texttt{subprocess}, or \texttt{tkinter} are not importable, and \texttt{pyautogui.screenshot()} and \texttt{pyautogui.locateCenterOnScreen} are explicitly disallowed. This isolates ``the agent moves the mouse and keyboard'' from ``the agent edits files or runs shell commands,'' so that screenshot-only success cannot be accidentally laundered through an unrelated automation
API.

A representative episode therefore alternates between short natural-language reflections and minimal PyAutoGUI snippets:

\small
\begin{Verbatim}[fontsize=\small]
# step k reflection
The Calc workbook is open on Sheet1 and the cursor is in A1.
I need to enter the SUM formula, so I will click B10 first.

```python
import pyautogui, time
pyautogui.click(x=412, y=638)
time.sleep(0.3)
pyautogui.typewrite("=SUM(B2:B9)", interval=0.02)
pyautogui.press("enter")
```
\end{Verbatim}
\normalsize

\subsubsection{Structured Action Schema}
\label{app:osomni_action_schema}

To facilitate tool-calling APIs and potential reinforcement-learning applications, \textsc{OS-Omni} also exposes a structured variant in which actions are wrapped into a finite class with parameterized enumeration. This structured approach supports constrained decoding by providing a distinct set of actions to be learned and optimized. The schema is shared across desktop, mobile, and web lanes so that the same agent interface covers heterogeneous endpoint types while preserving backend-specific execution. In every case, the \texttt{target} field of a pointing action accepts either an absolute pixel coordinate \texttt{[x,~y]}, a normalized coordinate \texttt{\{"x":~0.5,~"y":~0.5\}}, or, when the accessibility tree is exposed to the agent, a node identifier from that tree. Tables~\ref{tab:osomni_desktop_actions}~and~\ref{tab:osomni_mobile_actions} summarize the main desktop and mobile action schemas.

\begin{table}[ht]
\centering
\caption{Action types and parameters defined in the desktop action schema (Ubuntu/Linux, Windows, and macOS). Each desktop guest exposes the same schema to the agent.}
\label{tab:osomni_desktop_actions}
\small
\setlength{\tabcolsep}{5pt}
\renewcommand{\arraystretch}{1.12}
\begin{tabular}{p{2.4cm}p{4.0cm}p{6.6cm}}
\toprule
Action Type & Parameters & Note \\
\midrule
\texttt{TAP}          & \texttt{target}                                              & Click at the specified element or position. \\
\texttt{DOUBLE\_TAP}  & \texttt{target}                                              & Double-click at the specified element or position. \\
\texttt{RIGHT\_CLICK} & \texttt{target}                                              & Right-click at the specified element or position. \\
\texttt{DRAG}         & \texttt{from\_target}, \texttt{to\_target}, \texttt{duration\_ms} & Drag from one position to another with the left button held. \\
\texttt{SCROLL}       & \texttt{direction}, \texttt{distance}, \texttt{target}       & Scroll up, down, left, or right; optional anchor target. \\
\texttt{TYPE\_TEXT}   & \texttt{text}, \texttt{target}, \texttt{clear\_first}        & Type text, optionally after focusing the target field. \\
\texttt{KEYBOARD}     & \texttt{action}, \texttt{count}                              & Press a single key (\texttt{enter}, \texttt{escape}, \texttt{backspace}, \dots) one or more times. \\
\texttt{HOTKEY}       & \texttt{keys}                                                & Press a keyboard shortcut, e.g. \texttt{[ctrl, c]}. \\
\texttt{WAIT}         & \texttt{ms}                                                  & Wait for the UI to settle. \\
\texttt{DONE}         & \texttt{message}                                             & Decide the task is done. \\
\texttt{FAIL}         & \texttt{message}                                             & Decide the task cannot be performed. \\
\bottomrule
\end{tabular}
\end{table}

\begin{table}[ht]
\centering
\caption{Action types and parameters defined in the mobile action schema. A and I indicate availability on Android and iOS, respectively.}
\label{tab:osomni_mobile_actions}
\small
\setlength{\tabcolsep}{5pt}
\renewcommand{\arraystretch}{1.12}
\begin{tabular}{p{2.4cm}p{1.6cm}p{4.4cm}p{4.6cm}}
\toprule
Action Type & Platform & Parameters & Note \\
\midrule
\texttt{TAP}          & A, I & \texttt{target}                                  & Tap at the specified element or position. \\
\texttt{DOUBLE\_TAP}  & I    & \texttt{target}                                  & Double tap at the specified element or position. \\
\texttt{LONG\_PRESS}  & A, I & \texttt{target}, \texttt{duration\_ms}           & Press and hold the target. \\
\texttt{SWIPE}        & A    & \texttt{from\_target}, \texttt{to\_target}, \texttt{duration\_ms} & Swipe from one position to another. \\
\texttt{DRAG}         & I    & \texttt{from}, \texttt{to}, \texttt{duration\_ms} & Drag from one element/position to another, e.g. to reorder. \\
\texttt{SCROLL}       & A, I & \texttt{direction}, \texttt{distance}, \texttt{target}, \texttt{duration\_ms} & Scroll the screen or specified container. \\
\texttt{TYPE\_TEXT}   & A, I & \texttt{text}, \texttt{target}, \texttt{clear\_first} & Type text, optionally after tapping a target field. \\
\texttt{KEYBOARD}     & A, I & \texttt{action}, \texttt{count}                  & Send a keyboard or system key (\texttt{enter}, \texttt{back}, \texttt{home}, \texttt{escape}, \dots). \\
\texttt{CLIPBOARD}    & I    & \texttt{action}, \texttt{text}                   & Set, get, or paste clipboard content. \\
\texttt{ALERT}        & I    & \texttt{action}                                  & Accept or dismiss a system alert / popup. \\
\texttt{SYSTEM}       & I    & \texttt{action}                                  & System action (\texttt{home}, \texttt{hideKeyboard}). \\
\texttt{LAUNCH\_APP}  & A, I & \texttt{name}                                    & Launch an app by package or bundle identifier. \\
\texttt{WAIT}         & A, I & \texttt{ms}                                      & Wait for the UI to settle. \\
\texttt{DONE}         & A, I & \texttt{message}                                 & Decide the task is done. \\
\texttt{FAIL}         & A, I & \texttt{message}                                 & Decide the task cannot be performed. \\
\bottomrule
\end{tabular}
\end{table}

The desktop schema retains the full mouse, keyboard, and shortcut surface that PyAutoGUI exposes, but presents it as a discrete, parameterized action space with explicit terminal symbols, making the desktop lane directly comparable to the mobile lanes during benchmark reporting. The mobile schemas extend the same conceptual interface with the additional intents that touch UIs require, including \texttt{LONG\_PRESS}, \texttt{SWIPE}, \texttt{ALERT}, \texttt{SYSTEM}, and \texttt{LAUNCH\_APP}.
\texttt{DONE} and \texttt{FAIL} terminate the episode, whereas \texttt{WAIT} is non-terminal. Episodes may also terminate upon budget exhaustion or an unrecoverable runtime condition. After termination, the evaluator checks final-state correctness and any applicable execution constraints, rather than similarity to a reference trajectory.

\section{Details of the \textsc{OS-Omni} Benchmark}

\subsection{Platform Selection}
\label{app:os_selection}
\textsc{OS-Omni} evaluates agents across Windows, Ubuntu, macOS, iOS, Android, and Web environments. The selection is intended to cover desktop, mobile, and browser-based computer use rather than optimize for market share alone. In StatCounter's March 2026 worldwide statistics, Windows accounts for 60.8\% of desktop traffic, OS X/macOS for 14.76\%, Linux for 3.16\%, and ChromeOS for 1.62\%.\footnote{\url{https://gs.statcounter.com/os-market-share/desktop/worldwide}, March 2026 data.} We include Windows and macOS to cover major proprietary desktop ecosystems, and Ubuntu/Linux to provide an open, reproducible desktop stack with broad office, communication, graphics, networking, and developer workflows. The Web group is included as a first-class benchmark environment because many everyday computer tasks are performed through browsers and service-backed interfaces. The current release therefore centers on controlled desktop, mobile, and web runtimes with comparable final-state evaluation.

For mobile environments, Android and iOS account for 67.46\% and 32.27\% of worldwide mobile operating-system traffic, respectively.\footnote{\url{https://gs.statcounter.com/os-market-share/mobile/worldwide}, March 2026 data.} They introduce interaction modes absent from desktop-only benchmarks, including touch gestures, software keyboards, permission dialogs, app switching, and device-specific state. Web tasks are instantiated through widely used browsers on supported operating systems, including Chrome, Edge, Firefox, and Safari; in v1, Web tasks run in Chrome or Chromium on the Ubuntu VM. Together, these environment groups require agents to operate under a unified screenshot-only protocol across pointer-and-keyboard desktops, touch-first mobile devices, and browser-based workflows. We additionally include cross-platform workflows to evaluate cases in which artifacts or information must be prepared in one environment and consumed in another.

\subsection{Software Selection}
\label{app:software_selection}
We select applications to span common end-user workflows while preserving deterministic initialization and final-state evaluation. The suite is intentionally centered on everyday applications and task patterns, so that the benchmark covers the kinds of software needs that arise in ordinary desktop and mobile use. It emphasizes software with persistent, user-visible state, including documents, spreadsheets, presentations, messages, calendars, notes, budgets, shopping carts, media libraries, references, diagrams, images, 3D scenes, packet captures, files, and editor workspaces. Applications whose success depends primarily on remote availability or hidden server-side state are used only when their state can be seeded and evaluated through controlled artifacts.

On desktop platforms, \textsc{OS-Omni} uses overlapping application families rather than an identical software stack. This supports cross-platform comparison while respecting native availability and user conventions. Representative families include office suites, communication tools, finance software, graphics and 3D editors, diagramming tools, packet-analysis tools, reference managers, media applications, and text or code editors. For shopping workflows, we use OpenShop, a controlled benchmark application, instead of real e-commerce services. This preserves the core interaction structure of shopping tasks, including search, cart editing, option selection, and checkout confirmation. At the same time, it avoids personal data collection, payment handling, real purchases, account security risks, and benchmark drift from external commercial systems.

On mobile platforms, the selected applications exercise touch-first interaction and app-level persistence through finance, notes, mail, calendar, reminders, contacts, browsing, file management, and system-utility tasks. Web tasks use browser-native interaction with benchmark-controlled services when external state would otherwise make evaluation unstable. Across all platforms, inclusion requires three properties: the environment can initialize meaningful state, the evaluator can inspect the final state, and the interface supports realistic visually grounded interaction, state tracking, and multi-step planning. Service-backed and platform-limited workflows are covered by the \texttt{Open}-series applications described next.

\subsection{The Open-Series Applications}
\label{app:open_series}
Section~\ref{sec:custom_software} introduces the 18 controlled applications that we develop for \textsc{OS-Omni}, collectively called the \texttt{Open} series. They are used by 180 of the 300 tasks in v1 (60.0\%). Their real-world counterparts either depend on user accounts, payments, and live third-party data, or are not available on every operating system. The \texttt{Open} applications reproduce the interaction structure of these services while keeping every piece of task-relevant state under benchmark control. Table~\ref{tab:open_series_apps} lists the applications and the number of v1 tasks that use each of them on each platform.

\paragraph{Design principles.}
All \texttt{Open} applications follow the same principles, so that tasks built on them can be initialized and evaluated deterministically.
\begin{itemize}[leftmargin=*,itemsep=1pt,topsep=2pt]
    \item \textbf{Controlled data.} Accounts, products, flights and hotels, posts, notes, transactions, and documents come from a fixed seed shipped with the application. Task logic needs no external service, and record identifiers are stable across runs.
    \item \textbf{Deterministic time.} Applications whose behavior depends on dates run on a fixed business date (for example, \texttt{OpenBank} treats 2026-06-01 as the current day), so relative dates in instructions and date-dependent states, such as whether a transfer is completed or scheduled, are reproducible.
    \item \textbf{Persistent, inspectable state.} Each application writes its business state to a local store: browser storage for the web builds, an embedded database or JSON store for the mobile builds, and a bundled local service for the desktop builds. Evaluators read the final records directly instead of inferring them from the final screen.
    \item \textbf{Business event records.} Applications also append durable business events, such as a submitted checkout, a posted review, or a created transfer. Evaluators use these records to attribute actions to the current episode and to detect duplicated actions. An order between actions is required only when the task instruction states it; browsing actions such as searching, opening a page, or switching tabs are never required or counted unless the instruction asks for them.
    \item \textbf{One data model across builds.} The nine cross-platform applications share their data model and seed across their builds, so the same kind of workflow can be posed on different endpoints and compared.
\end{itemize}

\paragraph{Deployment and initialization.}
On Windows and macOS, the applications are installed in the virtual-machine images as desktop builds. On Android and iOS, they are installed on the emulator or simulator as mobile builds, and their stores are read from the device. The nine browser-based applications, together with web builds of several cross-platform applications, are served inside the Ubuntu VM and opened in Chrome; their state lives in Chrome's local storage, which the harness flushes to disk before evaluation and the evaluator decodes with a LevelDB reader. For every task, the initializer resets the application store to the seed, adds the task-specific records (for example, previously placed orders, existing reviews, saved notes, or seeded messages), launches the application, and verifies the starting state before the agent receives control.

\begin{table}[t]
\centering
\scriptsize
\setlength{\tabcolsep}{3.2pt}
\caption{\textbf{\texttt{Open}-series applications in \textsc{OS-Omni} v1.} Counts are the number of v1 tasks that use each application on each platform. Cross-application tasks are counted under their platform; three of them combine \texttt{OpenShop} with Thunderbird Calendar (one on Windows, two on Linux). In total, 180 of the 300 tasks use at least one \texttt{Open} application.}
\label{tab:open_series_apps}
\begin{tabular}{l p{5.1cm} rrrrrr r}
\toprule
\textbf{Application} & \textbf{Domain and main operations} & \textbf{Win} & \textbf{Linux} & \textbf{macOS} & \textbf{iOS} & \textbf{Android} & \textbf{Web} & \textbf{Total} \\
\midrule
\multicolumn{9}{l}{\textit{Controlled cross-platform applications}} \\
\texttt{OpenShop} & E-commerce: search, product options, cart, checkout, order notes, reviews & 5 & 2 & 4 & 4 & 6 & 2 & 23 \\
\texttt{OpenNote} & Social note sharing: posting, liking, bookmarking, commenting, following, profile & 4 & -- & 4 & 4 & 6 & 3 & 21 \\
\texttt{OpenTwitter} & Social networking: posts, replies, reposts, likes, saves, direct messages, profile & 4 & -- & 4 & 4 & 6 & 2 & 20 \\
\texttt{OpenBank} & Online banking: recipients, transfers, bill payments, disputes, alerts, statements & 4 & -- & 3 & 3 & 6 & 3 & 19 \\
\texttt{OpenDriver} & Ride-hailing with rider and driver interfaces: rides, documents, support tickets, payouts & 4 & -- & 4 & 4 & 4 & 3 & 19 \\
\texttt{OpenTrip} & Flight and hotel booking: search, checkout, vouchers, reviews, rescheduling & 4 & -- & 4 & 4 & 4 & 2 & 18 \\
\texttt{OpenPhotoEdit} & Image editing: adjustments, filters, calibrated exports & -- & -- & 4 & 4 & 4 & 3 & 15 \\
\texttt{OpenFood} & Food delivery: ordering, coupons, merchant and courier roles, customer support & 4 & -- & -- & -- & 4 & 3 & 11 \\
\texttt{OpenNavigation} & Route planning: destinations, waypoints, travel modes, incident reports & 4 & -- & -- & -- & 3 & 3 & 10 \\
\midrule
\multicolumn{9}{l}{\textit{Web-based applications for software coverage gaps (served in the Ubuntu VM)}} \\
\texttt{OpenDoc} & Document editing, comments, and export & -- & -- & -- & -- & -- & 3 & 3 \\
\texttt{OpenSheet} & Spreadsheets: formulas, filters, and export & -- & -- & -- & -- & -- & 3 & 3 \\
\texttt{OpenSlide} & Presentations: layouts, shapes, notes, comments, slide show & -- & -- & -- & -- & -- & 2 & 2 \\
\texttt{OpenMail} & Email, attachments, and contact management & -- & -- & -- & -- & -- & 3 & 3 \\
\texttt{OpenFinance} & Personal finance, budgeting, and portfolio management & -- & -- & -- & -- & -- & 3 & 3 \\
\texttt{OpenCode} & Code editing, debugging, and version control & -- & -- & -- & -- & -- & 3 & 3 \\
\texttt{OpenBlender} & 3D scene editing and animation & -- & -- & -- & -- & -- & 3 & 3 \\
\texttt{OpenWireshark} & Network packet analysis & -- & -- & -- & -- & -- & 2 & 2 \\
\texttt{OpenZotero} & Reference management & -- & -- & -- & -- & -- & 2 & 2 \\
\midrule
\textbf{Total} & & 33 & 2 & 27 & 27 & 43 & 48 & \textbf{180} \\
\bottomrule
\end{tabular}
\end{table}

\subsection{Task Example Sources}
\label{sub:task_exam_srcs}
We construct task examples from sources that reflect ordinary computer use while remaining reproducible under execution-based evaluation. Public documentation, application tutorials, community Q\&A, how-to resources, common productivity workflows, controlled benchmark applications, and cross-application scenarios are used to identify realistic user intents, application states, and evaluable final conditions. These sources seed tasks in office editing, email, chat and messaging, calendar management, finance, file handling, media, diagrams, packet analysis, reference management, and developer-style editing.

Not all realistic workflows can be used directly. Tasks are included only when the benchmark can initialize a meaningful state and evaluate the final result without relying on a hidden reference trajectory. For workflows involving external services, privacy-sensitive data, or unstable online state, we either use controlled local artifacts or a benchmark application with the same interaction structure. OpenShop is one such example: it lets us evaluate search, cart editing, option selection, and checkout confirmation without involving real purchases, payment information, or user accounts.

We also include tasks that cover common interaction conditions in real applications, such as dense visual layouts, small controls, modal dialogs, delayed saves, software keyboards, app switching, long forms, and information transfer across applications or platforms. The goal is not to imitate a single demonstration, but to define a user intent, an initial state, and a final-state check. Each task is therefore represented as a YAML specification containing the instruction, initializer, limits, metadata, and evaluator.

Task authoring follows the same validation discipline as the main benchmark pipeline. Annotators first design tasks from hands-on use cases, calibrate the difficulty so that completion requires meaningful planning rather than a single obvious click, and then execute the oracle version to confirm that the initializer, runtime, and evaluator support the intended goal. Failure-state checks are used to catch underspecified instructions, accidental shortcuts, and evaluators that accept incomplete results.

\subsection{Task Examples Collection}
\label{app:examples_collection}
\textsc{OS-Omni} v1 contains 300 task specifications: 48 tasks for each of Windows, Ubuntu, macOS, iOS, Android, and Web, and 12 cross-device workflows. Each task is paired with an oracle specification used for validation, debugging, and sanity checks. The oracle versions are not treated as separate benchmark tasks; instead, they provide a controlled reference for checking task feasibility, initializer correctness, runtime stability, and evaluator behavior. Table~\ref{tab:task_collection_statistics} summarizes the task distribution and oracle coverage.

\begin{table*}[ht]
\centering
\caption{Task distribution and oracle coverage in \textsc{OS-Omni} v1.}
\label{tab:task_collection_statistics}
\small
\setlength{\tabcolsep}{5pt}
\renewcommand{\arraystretch}{1.12}
\begin{tabularx}{\textwidth}{>{\raggedright\arraybackslash}p{0.16\textwidth}>{\raggedright\arraybackslash}p{0.16\textwidth}>{\raggedright\arraybackslash}p{0.18\textwidth}>{\raggedright\arraybackslash}X}
\toprule
Platform group & Tasks & Oracle coverage & Representative task families \\
\midrule
Windows & 48 & 48/48 & Office editing, communication, shopping, diagrams, packet analysis, reference management, and development workflows \\
Ubuntu & 48 & 48/48  & Office editing, email and chat, finance, graphics, 3D editing, packet analysis, references, media, and file workflows \\
macOS & 48 & 48/48 & Text and code editing, office workflows, email and calendar tasks, references, media, graphics, and productivity applications \\
iOS & 48 & 48/48 & Notes, browsing, mail, calendar, reminders, contacts, finance, shopping, and mobile productivity workflows \\
Android & 48 & 48/48 & Mail, notes, chat, calendar, finance, shopping, file management, media management, contacts, and system utilities \\
Web & 48 & 48/48 & Browser-based workflows in Chrome or Chromium on the Ubuntu VM \\
Cross-device & 12 & 12/12 & Phone-to-desktop round trips: Markor on Android, LibreOffice on the Ubuntu or Windows desktop, results returned to the phone \\
\midrule
Total & 300 & 300/300 & v1 task collection \\
\bottomrule
\end{tabularx}
\end{table*}

\subsection{Initial State Setup Details}
\label{app:benchmark_initial_state_setup}
\textsc{OS-Omni} tasks begin from realistic intermediate states rather than blank applications. Each initializer prepares the files, databases, app profiles, open windows or app views, local account fixtures, local services, and distractor artifacts needed by the task. This design mirrors ordinary computer use, where a user typically resumes work from a partially configured environment, an existing inbox, an open document, a populated calendar, or a mobile app with saved state.

Initial state setup follows the same high-level contract across platforms. First, the runtime is restored or prepared from a clean baseline. Linux and Windows use VM restoration, macOS uses Tart clones, Android uses emulator reset with ADB-controlled seeding, and iOS uses prepared Simulator state. Second, task-specific artifacts are placed into the environment, such as documents, spreadsheets, images, notes, packet captures, contact entries, or shopping records. Third, optional setup actions open the target application or navigate to a stable starting view. These setup operations are part of the benchmark specification and are completed before the first agent observation, regardless of the observation mode used in the run.

\subsection{Evaluation Configuration Details}
\label{app:benchmark_evaluation_configuration}
\textsc{OS-Omni} evaluates task completion from the final application state rather than from the action trajectory. After the agent stops or reaches the step limit, the runner dynamically loads the evaluator named in the task file, builds a platform-specific evaluation context, and records the returned \texttt{PASS}, \texttt{FAIL}, \texttt{UNDETERMINED}, or \texttt{ERROR} result together with diagnostic metadata. This design is important for open-ended GUI tasks: there are often many valid ways to complete the same instruction, but the completed environment should contain the requested user-visible result.

\paragraph{Evaluator interface.}
Each evaluator is a Python class loaded from an import path such as \texttt{evaluators.linux.OpenShop:OpenShopApiEvaluator}. All evaluators implement the shared \texttt{BaseEvaluator} interface and return an \texttt{EvalResult} containing a status, an optional message, and an open metadata dictionary. The context always includes the task parameters, the runtime controller, and the final observation. Desktop runners additionally pass the run directory and trajectory path when available, so evaluators can save retrieved artifacts or inspect episode logs. Windows also supports supplemental file and PowerShell checks that run after the primary evaluator passes.

\paragraph{Execution limits.}
The v1 task specifications use a long-horizon interaction budget: each single-platform task sets \texttt{limits.max\_steps: 300} and \texttt{limits.step\_timeout\_seconds: 180}, and each cross-device workflow sets 300--650 steps with a 120-second step timeout. The step timeout is a per-step execution budget rather than a whole-episode wall-clock limit; it bounds the model/action turn and prevents a hung action or tool call from blocking the run indefinitely. The evaluator is invoked after the agent emits \texttt{DONE} or \texttt{FAIL}, after the episode reaches its step budget, or after another terminal runtime condition. If \texttt{step\_timeout\_seconds} is omitted, the shared task loader falls back to 120 seconds.

\paragraph{Evaluation pipeline.}
Most evaluators follow a three-stage pipeline. First, a getter retrieves task-relevant evidence from the final environment: a saved file, application database, profile folder, local service response, UI tree, simulator container, or device storage path. Second, the evaluator normalizes this evidence into structured records, such as workbook cells, document paragraphs, diagram labels, email messages, calendar events, shopping cart items, budget transactions, or browser bookmarks. Third, task-specific predicates compare the structured record against the specification in the YAML file. Failures therefore produce concrete mismatch messages and metadata, such as missing rows, unexpected files, unmatched events, observed cart contents, copied artifacts, command output, or parsed database rows.

\begin{table}[ht]
\centering
\caption{Representative final-state signals used by \textsc{OS-Omni} evaluators.}
\label{tab:evaluation_signal_types}
\small
\setlength{\tabcolsep}{5pt}
\renewcommand{\arraystretch}{1.15}
\begin{tabular}{p{3.1cm}p{5.0cm}p{4.1cm}}
\toprule
Signal & Access path & Examples \\
\midrule
Office/text artifacts & Guest file retrieval; host-side parsers & XLSX, DOCX, PPTX, plain text \\
\midrule
App stores and profiles & SQLite, JSON, XML, plist, profile files & mail, calendar, finance, Zotero \\
\midrule
Controlled services & Local API or localStorage & OpenShop session, cart, order \\
\midrule
Creative/analysis files & Project parsing and app scripts & GIMP, Blender, draw.io, Wireshark \\
\midrule
Mobile state & ADB, device storage, simulator containers, UI state & notes, contacts, Safari, reminders \\
\midrule
Cross-application state & Child evaluators plus transfer artifacts & attachments and follow-ups \\
\bottomrule
\end{tabular}
\end{table}

\subsubsection{Ubuntu/Linux}
\paragraph{LibreOffice.}
Calc, Writer, and Impress tasks are evaluated from the saved workbook, document, or presentation rather than from the clicks used to edit it. The evaluator first resolves the result file: if the path is already available on the host it is read directly, otherwise it is fetched from the guest with \texttt{controller.get\_file}. Spreadsheet checks use workbook-level parsers to compare cells, formulas, sheets, sheet order, and rule-based properties such as filtered ranges or derived tables. Writer and Impress checks parse document and presentation artifacts to verify text, tables, page breaks, slide contents, and transitions. The metadata records the resolved actual and expected paths and the specific mismatch when comparison fails.

\paragraph{Thunderbird and Delta Chat.}
Thunderbird evaluation uses both profile-level evidence and final UI evidence. Some tasks read files from the Thunderbird profile, such as preference files, filter definitions, or SQLite-backed state; others run post-processing commands, cache command output, or inspect the final accessibility tree when a compose window should remain open rather than send a message. The checker supports list, CSV, JSON, SQLite, preference, filter, and accessibility-tree rules. Delta Chat tasks inspect persisted conversation state through local database queries and task-specific message predicates, including sent messages, replies, and group or coordination state.

\paragraph{Actual Budget and OpenShop.}
Actual Budget tasks are evaluated by reading the application's local budget store and SQLite database. The checker searches for accounts, starting-balance transactions, categories, budget amounts, ordinary transactions, and transfer pairs while allowing the internal money representation used by the app. OpenShop tasks query the controlled local shopping app through its local API. The evaluator checks session state, cart item counts, product identifiers, quantities, selected specifications, order totals, address labels, payment options, and order notes. Because OpenShop is benchmark-controlled, these checks avoid real purchases, payment data, and external service drift.

\paragraph{Blender, GIMP, draw.io, Marble, Wireshark, Zotero, Notepadqq, VS Code, and Spotube.}
Creative, analytical, reference, development, and media tasks use application-specific evidence. Blender evaluators run guest-side scripts to extract scene properties and compare object placement, render settings, and camera or lighting requirements. GIMP evaluators retrieve exported images and inspect dimensions, mode, and pixel-level differences with image-processing libraries. draw.io evaluators decode the saved diagram XML, including compressed diagram payloads, then compare pages and cell labels. Marble tasks parse bookmark or configuration state. Wireshark tasks verify exported captures or protocol reports. Zotero checks query the local library database for collections, items, tags, creators, and notes. Notepadqq and VS Code tasks compare edited workspace files or settings JSON. Spotube tasks inspect persistent profile, market, queue, and playback-related state.

\subsubsection{Windows}
\paragraph{LibreOffice.}
Calc, Writer, and Impress tasks are evaluated from saved files and structured document properties. Windows tasks often use checked-in PowerShell scripts that force-save when necessary, retrieve the target file, and run Python checks with libraries such as \texttt{openpyxl}, \texttt{python-docx}, \texttt{python-pptx}, XML parsing, or ZIP inspection. These scripts verify spreadsheet formulas and charts, document formatting and structure, page breaks, images, table-of-contents state, slide shapes, notes, animations, and transitions.

\paragraph{Thunderbird, Calendar, and Delta Chat.}
Thunderbird and Calendar tasks check saved messages, folders, filters, contacts, and events by combining profile inspection, SQLite queries, and guest-side scripts. Delta Chat tasks verify sent messages, group creation, and shared synchronization artifacts through local database or runtime checks. The Windows runner can also attach supplemental file checks or inline PowerShell checks to the primary evaluator, which is useful for tasks whose success depends on both an app state and an auxiliary artifact.

\paragraph{Actual Budget and OpenShop.}
Actual Budget tasks use PowerShell-launched Python checks over the application's SQLite state to validate accounts, categories, transactions, and setup details. OpenShop uses the same controlled benchmark application as the Linux lane but queries it from the Windows guest with Python. Its evaluator polls the local API until the state stabilizes, then checks the session, cart, order, and optional UI evidence such as process/window visibility.

\paragraph{Blender, GIMP, draw.io, Wireshark, Zotero, VS Code, Notepad, Spotube, and PowerShell checks.}
These tasks are evaluated through saved project files, exported artifacts, packet captures, library metadata, source files, plain-text files, persistent application settings, or command-level evidence. Several Windows evaluators are thin wrappers over named PowerShell scripts in \texttt{evaluators/windows/checks}. The script output and exit code become structured metadata, and selected guest artifacts can be copied into \texttt{evaluation\_artifacts} under the run directory for later inspection.

\subsubsection{macOS}
\paragraph{LibreOffice.}
Calc, Writer, and Impress tasks check saved spreadsheet, document, and presentation artifacts retrieved from the Tart guest. The evaluators parse the ZIP/XML structure of LibreOffice-compatible documents to verify cell contents, document paragraphs, title formatting, tables, footer or heading requirements, and slide contents. This avoids depending on the visible final screen, which may not show all required document state.

\paragraph{Thunderbird Email, Thunderbird Calendar, and Delta Chat.}
Email and calendar tasks inspect saved messages, folders, filters, contacts, and events using guest-side Python scripts and local profile or SQLite state. Delta Chat tasks run guest-side database queries and return JSON summaries of conversations, recipients, and messages. The evaluator then checks exact recipients, subjects, message bodies, and required reply or coordination state.

\paragraph{Actual Budget, OpenShop, and Spotube.}
Actual Budget tasks scan \texttt{global-store.json}, locate the active \texttt{db.sqlite}, copy sidecar WAL/SHM files when present, and query accounts, categories, budgets, transactions, and transfers. OpenShop tasks check the controlled shopping application's local API or persistent state. Spotube tasks inspect the local SQLite and settings state used by the application to validate profile, market, plugin, queue, and playback-related requirements.

\paragraph{Blender, GIMP, draw.io, Wireshark, Zotero, VSCodium, and TextEdit.}
These tasks are evaluated from saved scene files, images, diagrams, packet evidence, reference library entries, workspace files, and edited text files. Blender and Wireshark evaluators run guest-side scripts that return JSON payloads. GIMP checks compare exported images with expected image properties or fixtures. draw.io decodes diagram XML and compares page/cell semantics. Zotero copies and queries \texttt{zotero.sqlite} and sidecar files to check collections, item titles, creators, tags, and notes. VSCodium and TextEdit evaluators inspect workspace files, settings, and edited text content.

\subsubsection{Android}
\paragraph{K-9 Mail and Delta Chat.}
Mail and chat tasks inspect sent messages, forwarded content, replies, archived state, contacts, and status changes. Evaluators use ADB-backed access to app state, local mail service evidence, or task-specific API records where available. Delta Chat evaluators validate exact outbound messages, forwarded content, search-and-reply behavior, archive state, and created contact state. Cross-application mail tasks additionally verify that information extracted from one app is correctly written or sent in another.

\paragraph{Fossify Calendar, Contacts, and Tasks.org.}
Scheduling and personal-information tasks check persisted event, contact, reminder, due-date, note, and subtask records. Contacts are validated through Android content-provider or database-style ADB queries. Tasks.org evaluators inspect persisted task fields, due dates, notes, parent tasks, and subtasks while accounting for the device timezone.

\paragraph{Markor, LibreOffice, Material Files, and Fossify Gallery.}
Note, document, file, and gallery tasks are evaluated through external storage and document artifacts. Generic filesystem evaluators check expected paths, forbidden paths, file/directory types, and exact directory entries on shared storage. Markor and text-file tasks read final note contents. LibreOffice Viewer tasks inspect modified office documents by pulling files from device storage and parsing the document package. Gallery tasks check renamed, moved, or deleted photo artifacts while preserving distractor files.

\paragraph{Oinkoin and OpenShop.}
Oinkoin tasks pull the app's \texttt{movements.db} through ADB root access, copy it to a temporary host location, and query records for matching title, note, category, sign, amount, and category type. OpenShop Android tasks read the WebView localStorage-backed state through the benchmark store helper. They poll until the state stabilizes and then check login status, cart size, required or forbidden cart items, latest order contents, address/payment choices, and order notes.

\paragraph{Spotube, Calculator, Clock, and Settings.}
Media and utility tasks check persistent app or device state. Spotube evaluators inspect XML/shared-preference or database state for profile, queue, playback, and search outcomes. Calculator and Clock tasks use UI-tree, screenshot-difference, activity, or device-state evidence when the final state is primarily visible rather than stored in a file. Settings tasks use ADB-readable device state, such as Bluetooth status.

\paragraph{Cross application Android tasks.}
Android also includes workflows combining Fossify Calendar with OpenShop, K-9 Mail with Markor or LibreOffice, and Fossify Calendar or Organic Maps with Delta Chat. These evaluators compose the relevant child checks and report child metadata separately, so a failure can be attributed to the source app, target app, or transfer step.

\subsubsection{iOS}
\paragraph{Calendar and Reminders.}
Calendar tasks are backed by the simulator's Calendar database. Evaluators locate the simulator by UDID when a direct database path is not provided, query \texttt{Calendar.sqlitedb}, normalize times with the configured timezone, and check required or forbidden event summaries, start/end times, locations, and notes. Reminders tasks combine visible UI evidence with app state: they check the foreground bundle, the selected list view, expected titles, forbidden titles, and exact-list constraints when the task requires a particular visible reminder list.

\paragraph{Contacts, Delta Chat, and Maily.}
Contacts tasks inspect \texttt{AddressBook.sqlitedb} from the simulator container and validate saved names, phone numbers, and related fields. Delta Chat and Maily tasks evaluate sent messages, forwarded content, deleted or retained messages, recipients, subjects, and message bodies, using local app state and controlled mail/message service evidence rather than the final screenshot alone.

\paragraph{FSNotes, Flow, and Habit.}
FSNotes tasks inspect the note root inside the simulator data container to verify created, moved, renamed, archived, or edited Markdown notes. Flow tasks inspect the app support directory and ObjectBox-backed state for expenses, transfers, income records, and budget updates. Habit tasks inspect the local habit database for created, removed, or modified routines. These evaluators poll briefly and include observed summaries in metadata to reduce false negatives from asynchronous app persistence.

\paragraph{OpenShop and Safari.}
OpenShop iOS tasks use the same controlled shopping-state abstraction as Android, but resolve the local store from the simulator container or explicit database path. They check login state, cart contents, forbidden items, latest order state, and checkout details. Safari tasks query simulator history, bookmarks, folders, tabs, active-tab state, and optional foreground-bundle evidence, so they can distinguish opening the target page from merely typing a URL into a visible field.

\subsubsection{Web}
\paragraph{Browser and controlled-service tasks.}
Web tasks are instantiated through Chrome, Edge, Firefox, and Safari on supported operating systems. Evaluators inspect the browser-visible final state and, when applicable, the state of benchmark-controlled services. The checked evidence includes sessions, form submissions, shopping carts, orders, bookmarks, tabs, history entries, downloads, and local storage or database records. Service-backed workflows keep state under benchmark control so that final-state checks do not depend on live commercial services, personal accounts, or unstable remote data.

\subsubsection{Cross-Application and Cross-Platform Workflows}
Cross-application tasks combine the app-level checks above when a user goal spans multiple applications on the same endpoint, such as Writer with Thunderbird, Thunderbird Calendar with OpenShop, K-9 Mail with Markor or LibreOffice, Calendar with OpenShop, or Calendar, Maps, and Delta Chat. The cross-app evaluator does not reward an intermediate handoff by itself; it checks the final state of every required destination artifact or application record. In v1, cross-device tasks run on one screen: the Ubuntu or Windows desktop shows a companion Android phone in a scrcpy window. The agent edits or builds a file in Markor on the phone, moves it to the desktop itself (for example, through the synced clipboard), builds the deliverable in LibreOffice, and sends results computed on the desktop back to the phone as a Markor note, Tasks items, Calendar events, or the exported file. The evaluator checks the phone's source file and returned records (exactly the expected ones and no others), checks that the copies on the phone match the desktop files, and checks the desktop document. Any agent action that invokes or types \texttt{adb} or \texttt{scrcpy} fails the task. These workflows replace the earlier cross-VM workflows (Android-to-Windows and Windows-to-Linux), which ran source and target stages on different endpoints. When a deterministic checker cannot represent all acceptable outcomes, the case is routed to the manual verification protocol described in Section~\ref{app:manual_verification_for_ambiguous_cases}.

\subsection{More Task Examples}
\label{app:more_task_examples}
\begin{figure*}[p]
\centering
\includegraphics[width=\textwidth]{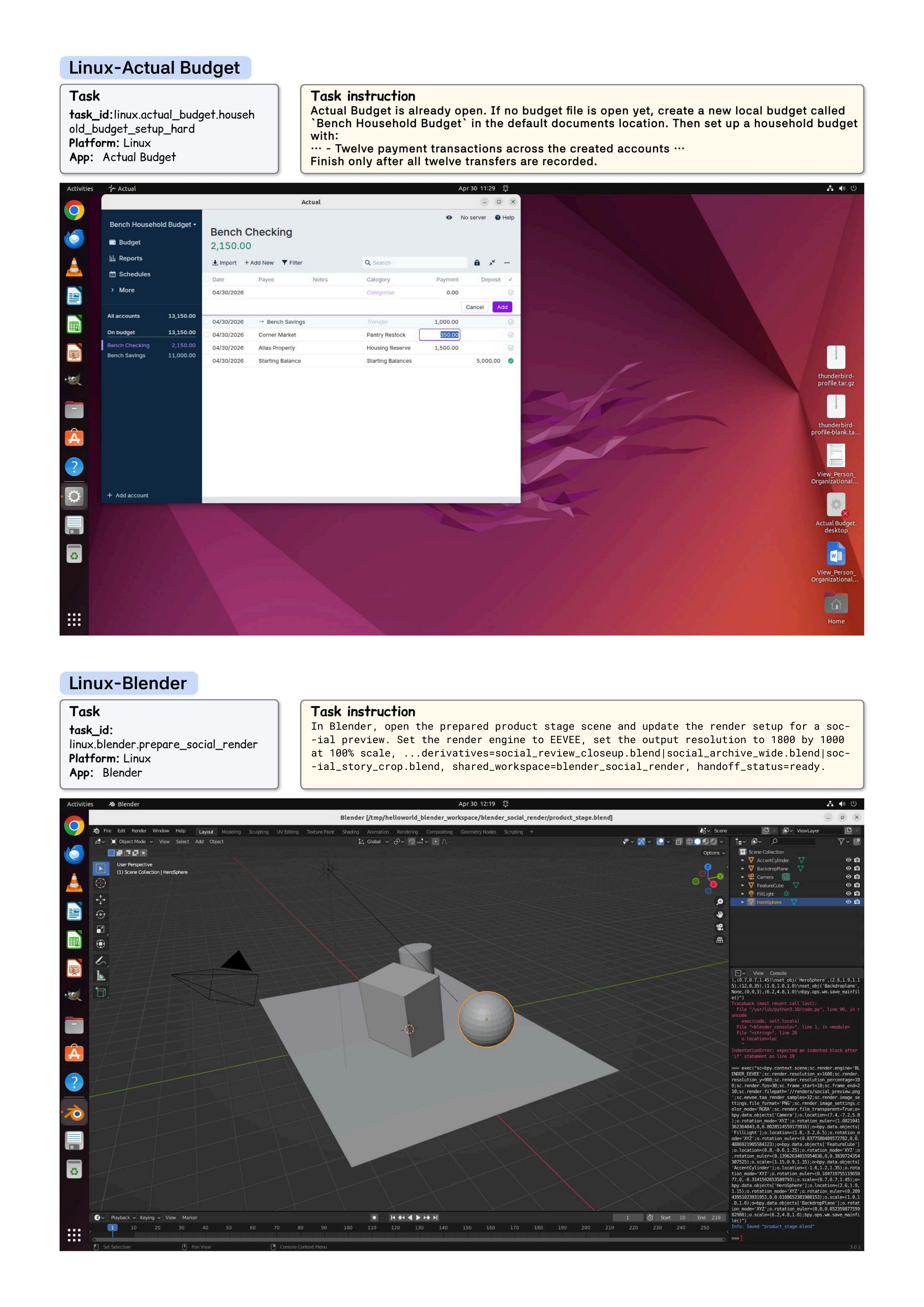}
\caption{Representative Ubuntu/Linux task examples.}
\par\smallskip\centering{\footnotesize\hyperlink{app:case_figure_list}{Back to list of figures}}
\label{fig:linux_task_examples_1}
\end{figure*}

\begin{figure*}[p]
\centering
\includegraphics[width=\textwidth]{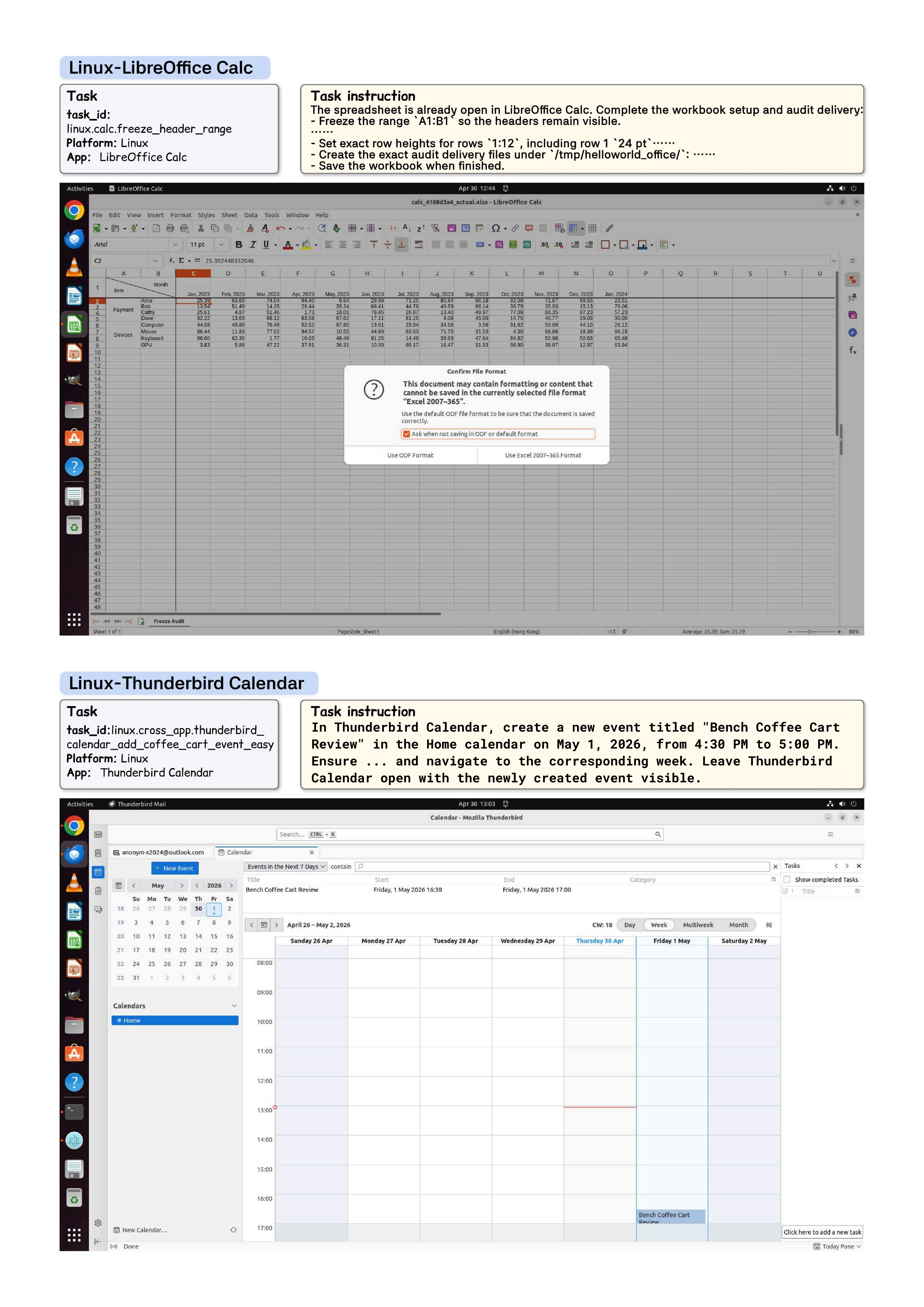}
\caption{Representative Ubuntu/Linux task examples.}
\par\smallskip\centering{\footnotesize\hyperlink{app:case_figure_list}{Back to list of figures}}
\label{fig:linux_task_examples_2}
\end{figure*}

\begin{figure*}[p]
\centering
\includegraphics[width=\textwidth]{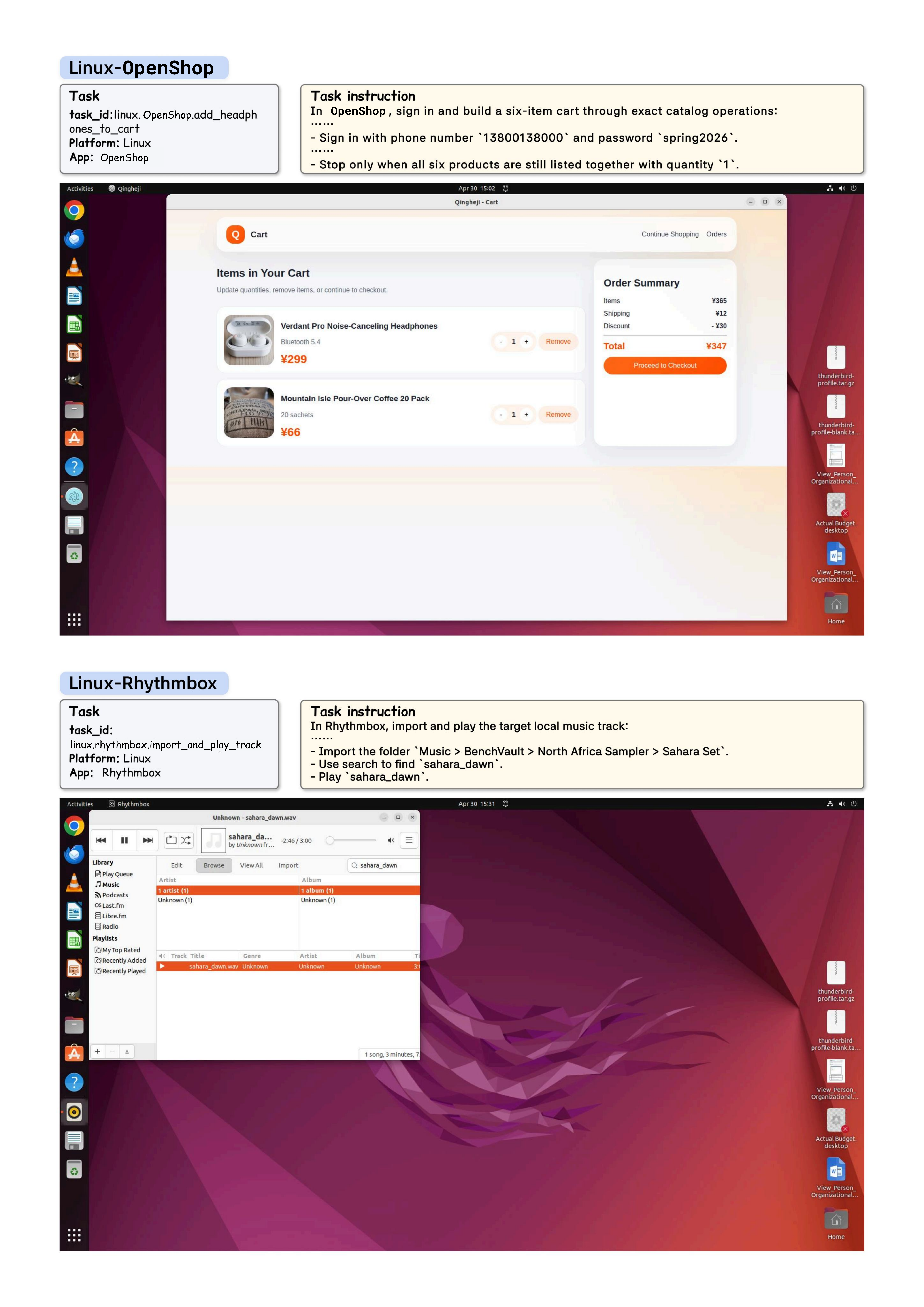}
\caption{Representative Ubuntu/Linux task examples.}
\par\smallskip\centering{\footnotesize\hyperlink{app:case_figure_list}{Back to list of figures}}
\label{fig:linux_task_examples_3}
\end{figure*}

\begin{figure*}[p]
\centering
\includegraphics[width=\textwidth]{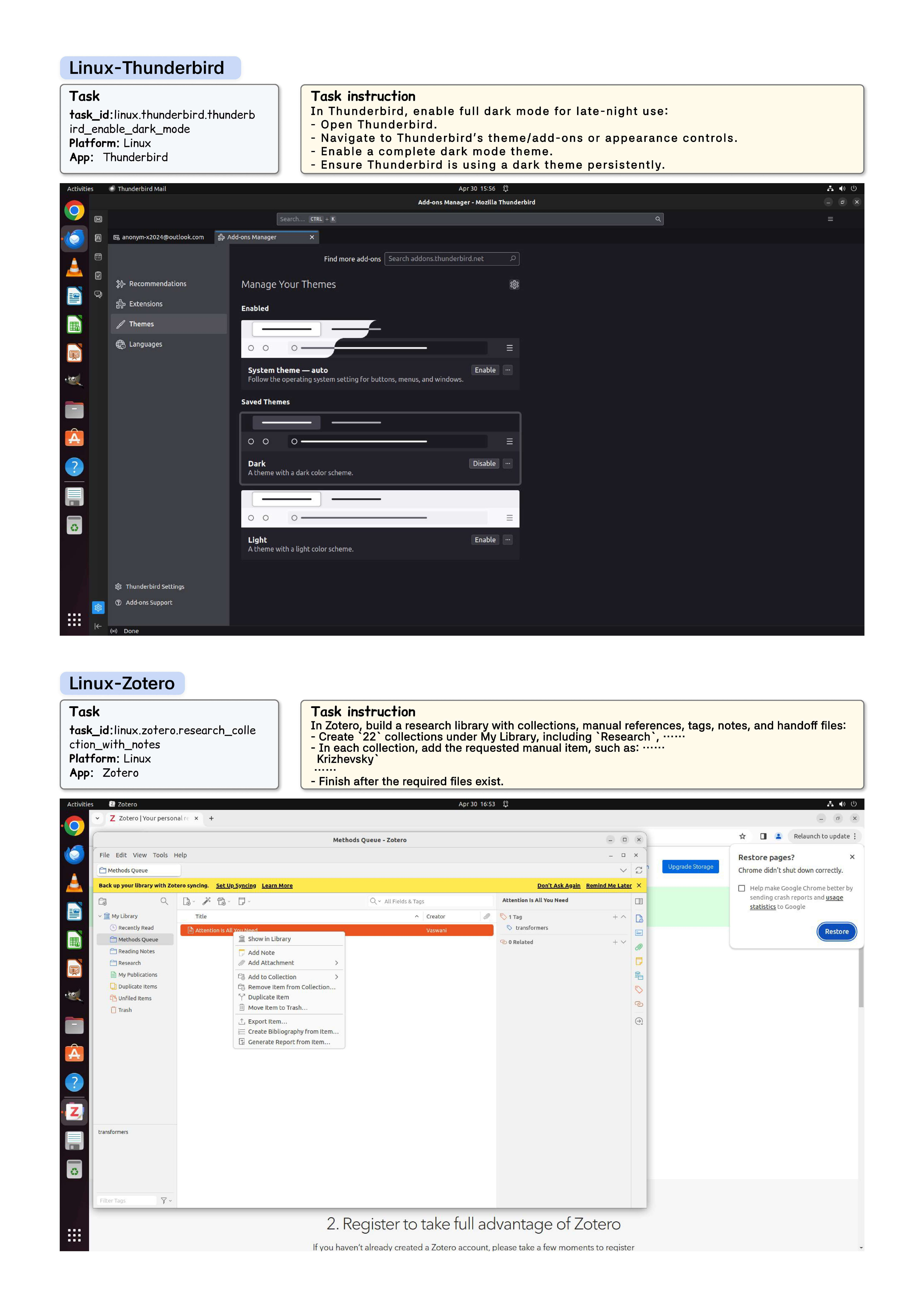}
\caption{Representative Ubuntu/Linux task examples.}
\par\smallskip\centering{\footnotesize\hyperlink{app:case_figure_list}{Back to list of figures}}
\label{fig:linux_task_examples_4}
\end{figure*}

\begin{figure*}[p]
\centering
\includegraphics[width=\textwidth]{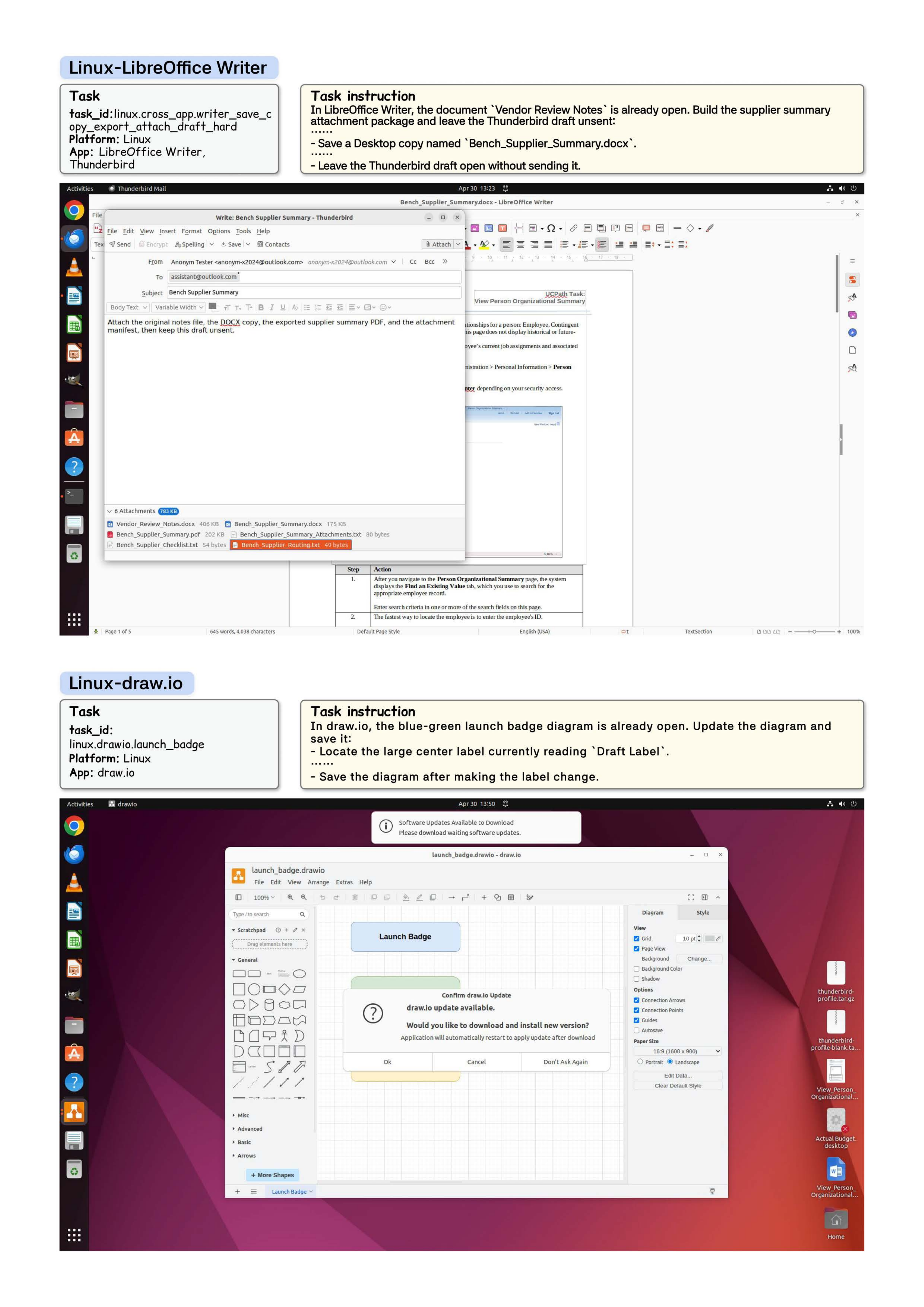}
\caption{Representative Ubuntu/Linux task examples.}
\par\smallskip\centering{\footnotesize\hyperlink{app:case_figure_list}{Back to list of figures}}
\label{fig:linux_task_examples_5}
\end{figure*}

\begin{figure*}[p]
\centering
\includegraphics[width=\textwidth]{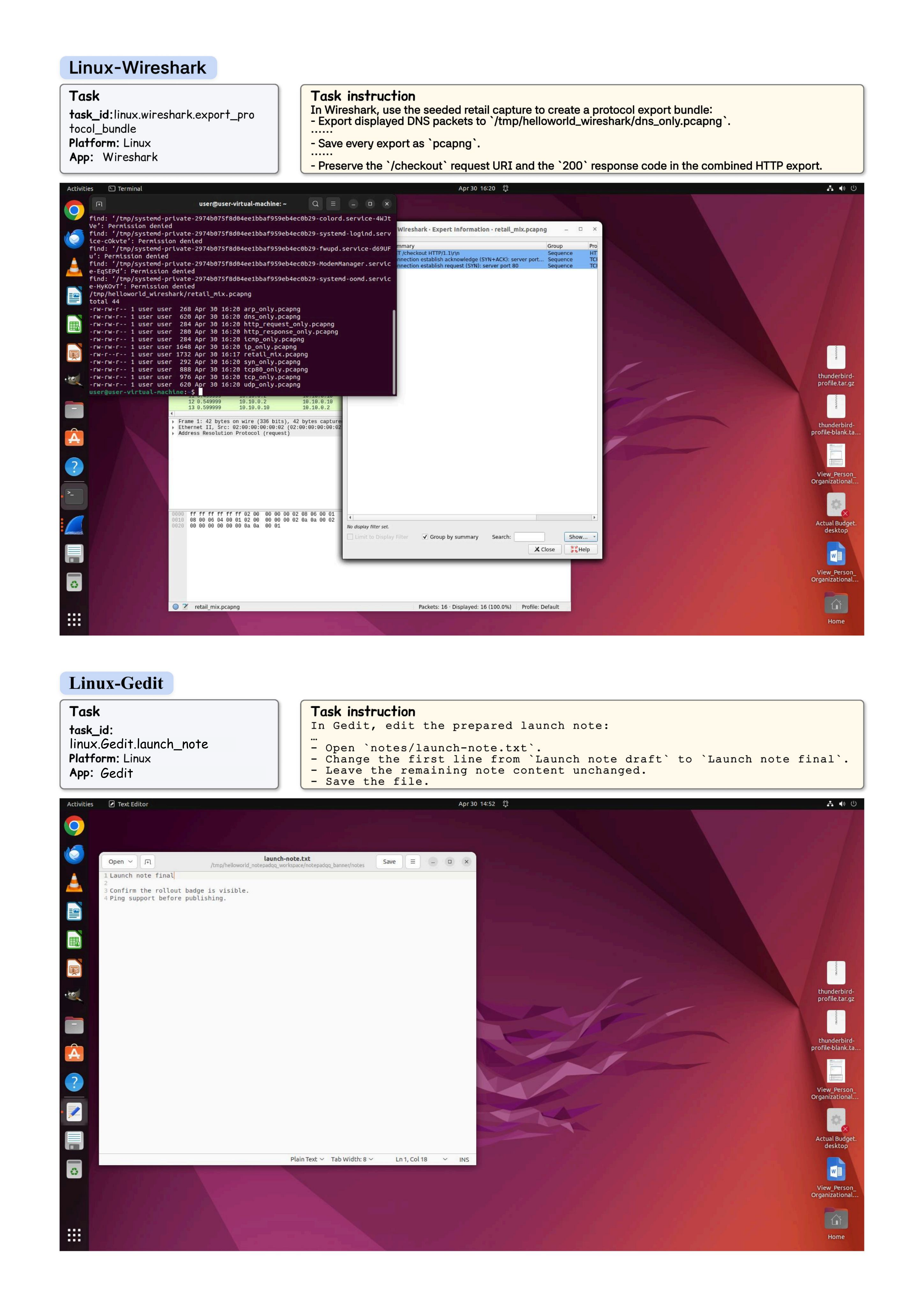}
\caption{Representative Ubuntu/Linux task examples.}
\par\smallskip\centering{\footnotesize\hyperlink{app:case_figure_list}{Back to list of figures}}
\label{fig:linux_task_examples_6}
\end{figure*}

\begin{figure*}[p]
\centering
\includegraphics[width=\textwidth]{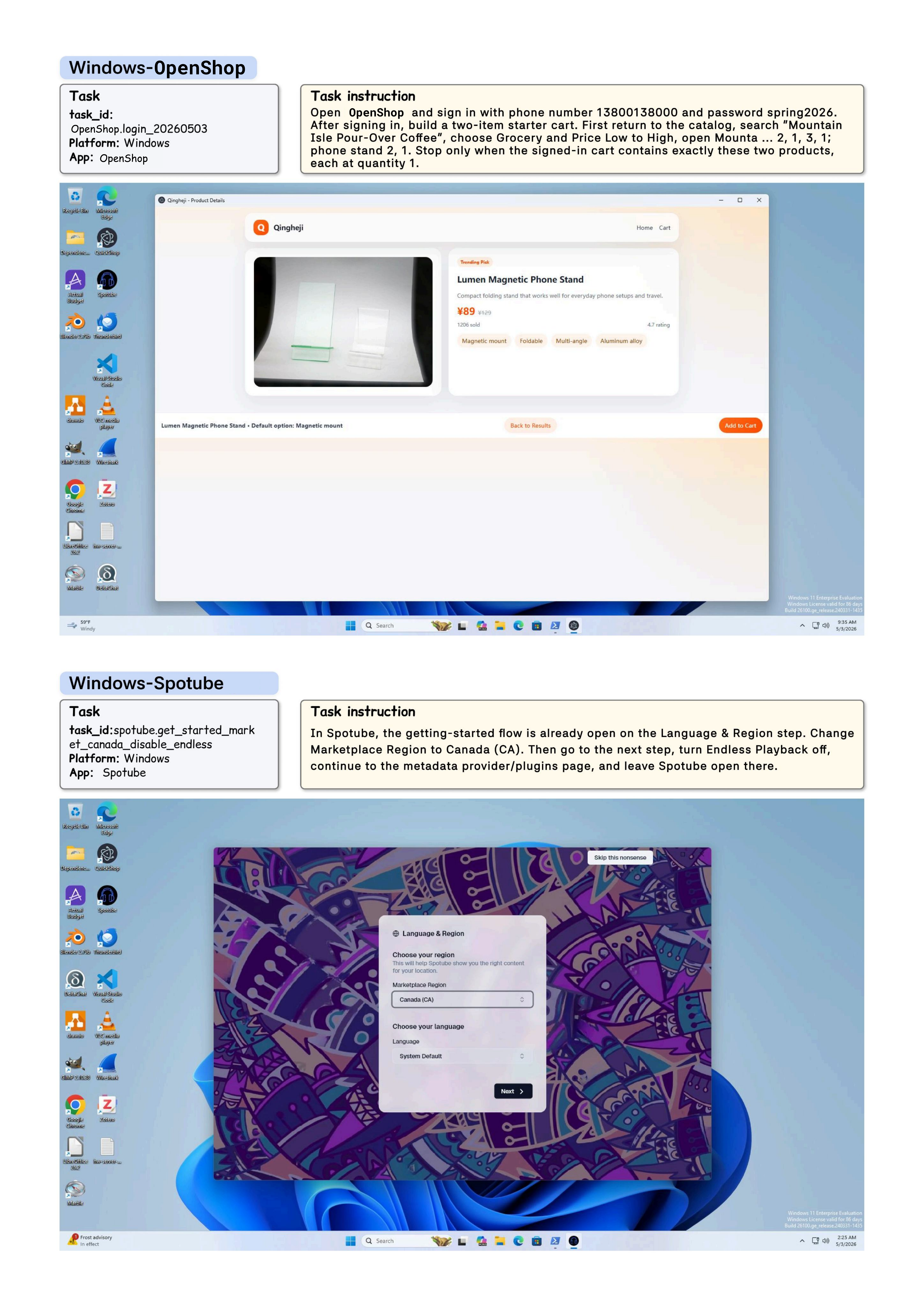}
\caption{Representative Windows task examples.}
\par\smallskip\centering{\footnotesize\hyperlink{app:case_figure_list}{Back to list of figures}}
\label{fig:windows_task_examples_1}
\end{figure*}

\begin{figure*}[p]
\centering
\includegraphics[width=\textwidth]{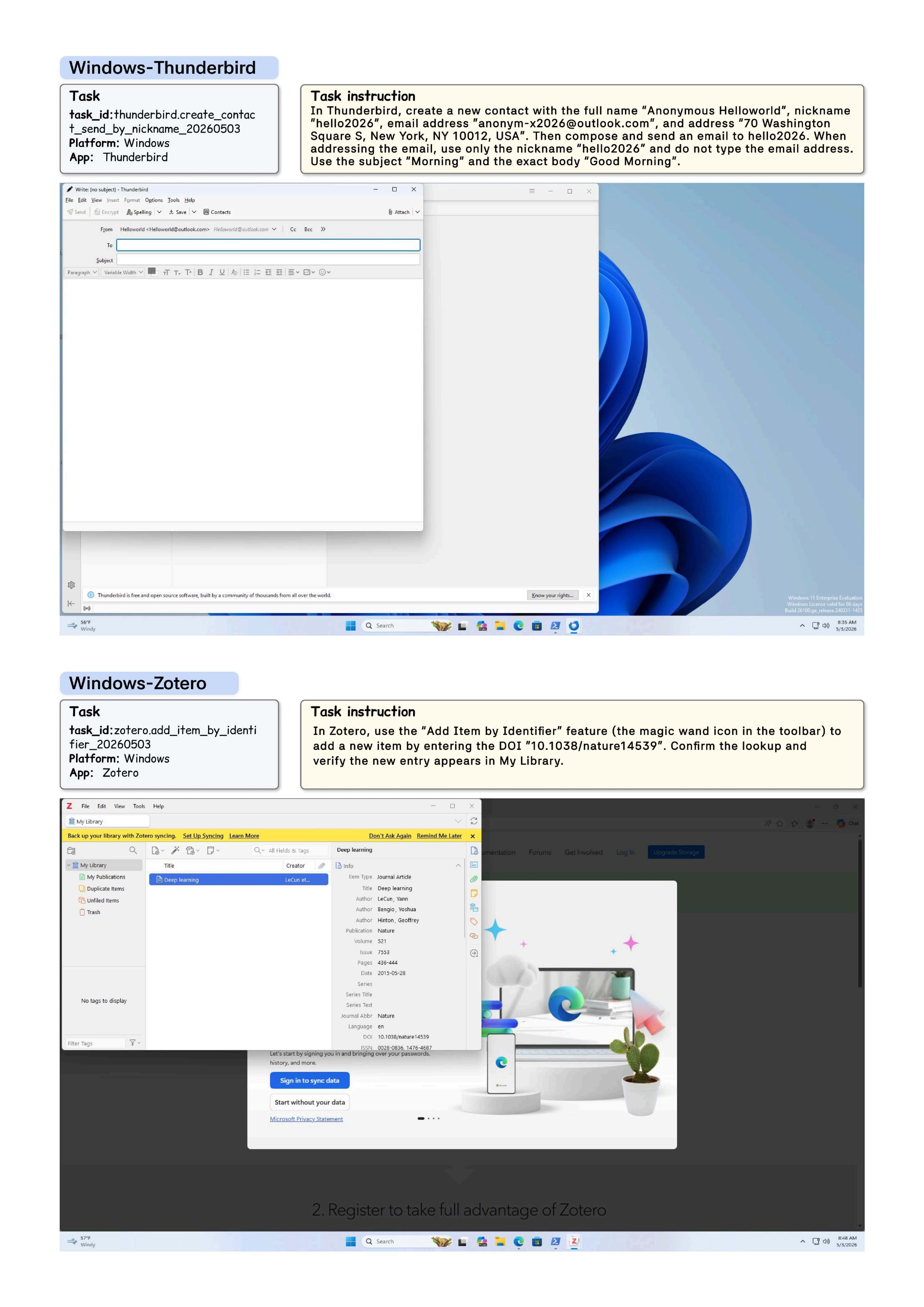}
\caption{Representative Windows task examples.}
\par\smallskip\centering{\footnotesize\hyperlink{app:case_figure_list}{Back to list of figures}}
\label{fig:windows_task_examples_2}
\end{figure*}

\begin{figure*}[p]
\centering
\includegraphics[width=\textwidth]{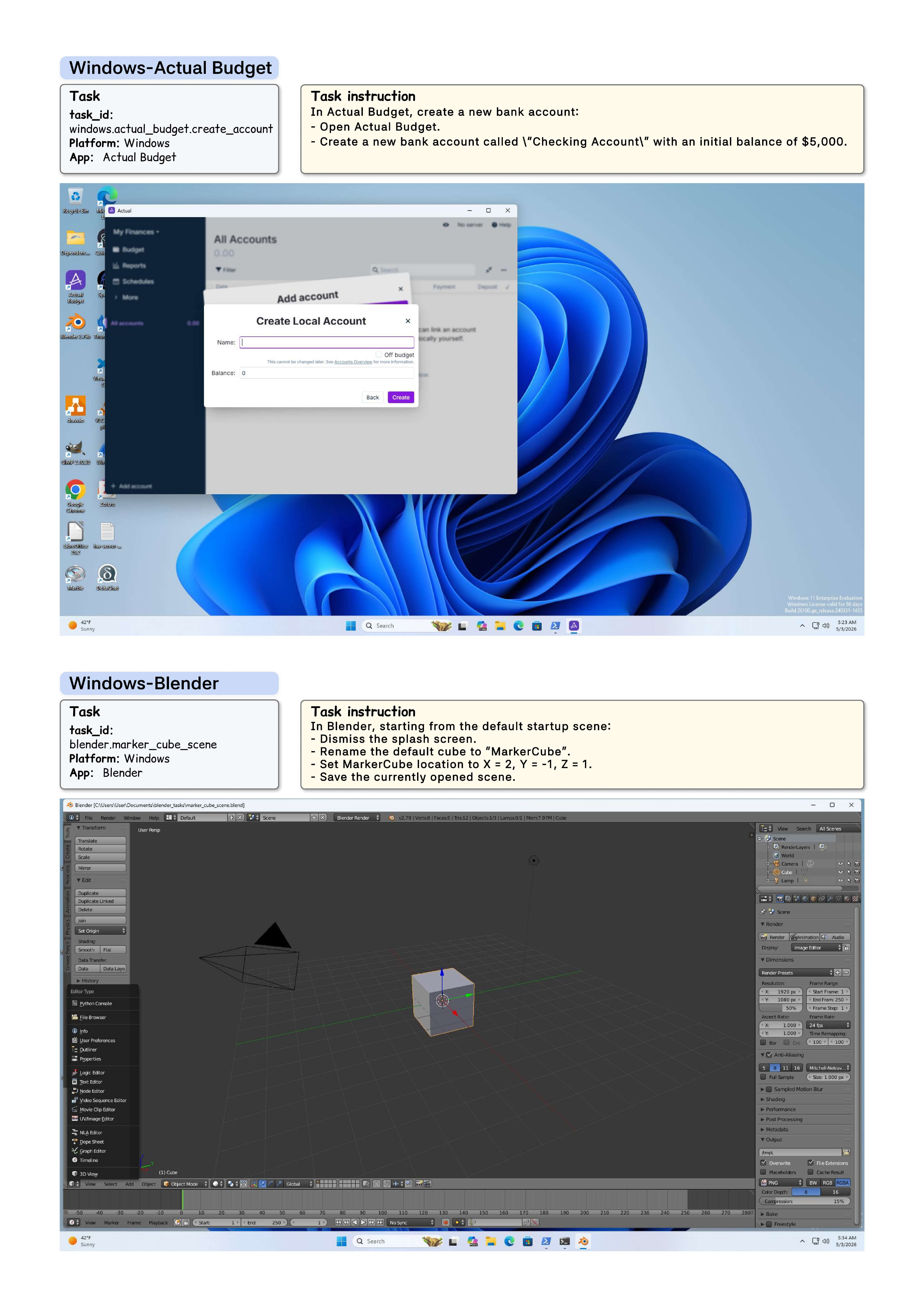}
\caption{Representative Windows task examples.}
\par\smallskip\centering{\footnotesize\hyperlink{app:case_figure_list}{Back to list of figures}}
\label{fig:windows_task_examples_3}
\end{figure*}

\begin{figure*}[p]
\centering
\includegraphics[width=\textwidth]{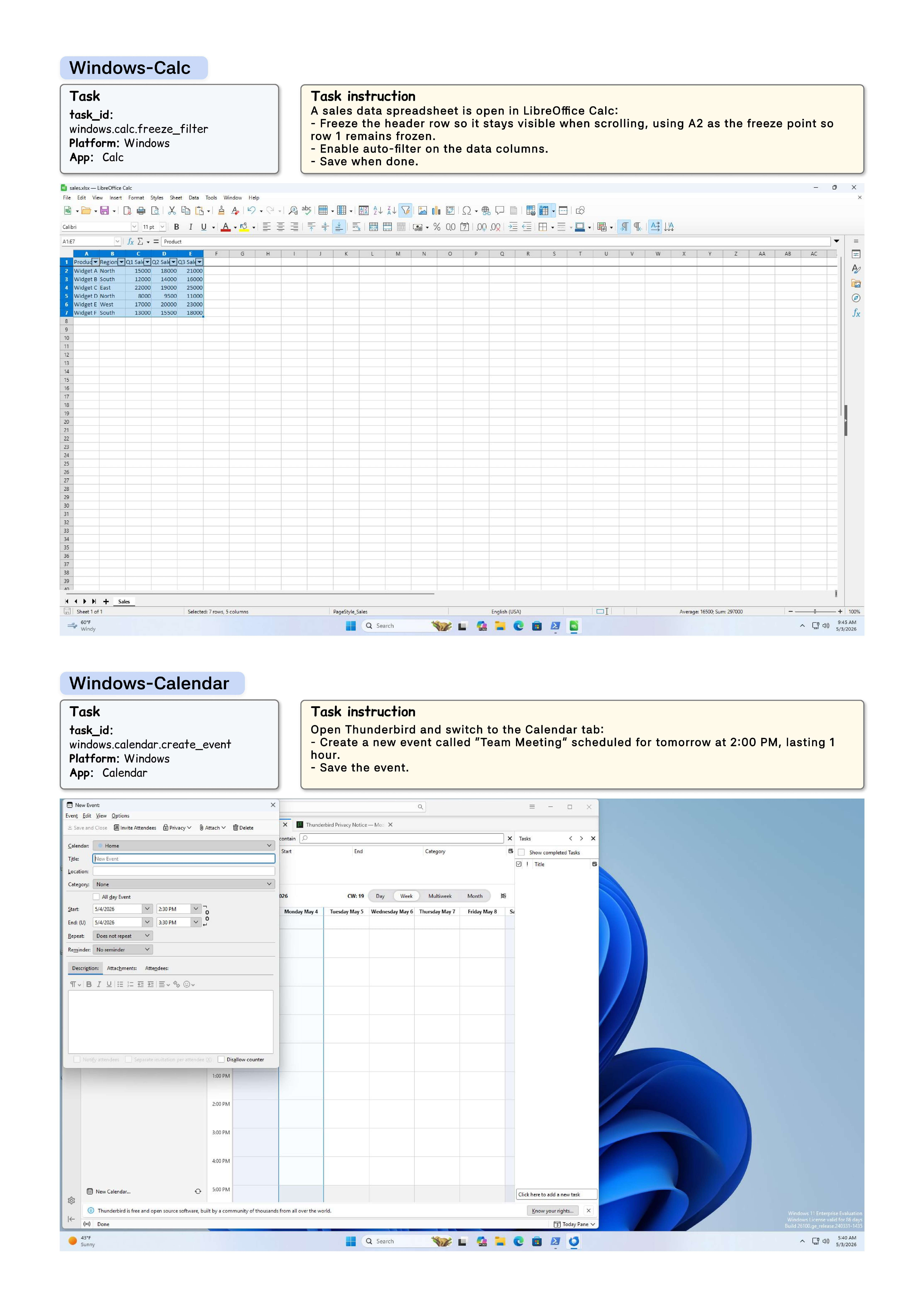}
\caption{Representative Windows task examples.}
\par\smallskip\centering{\footnotesize\hyperlink{app:case_figure_list}{Back to list of figures}}
\label{fig:windows_task_examples_4}
\end{figure*}

\begin{figure*}[p]
\centering
\includegraphics[width=\textwidth]{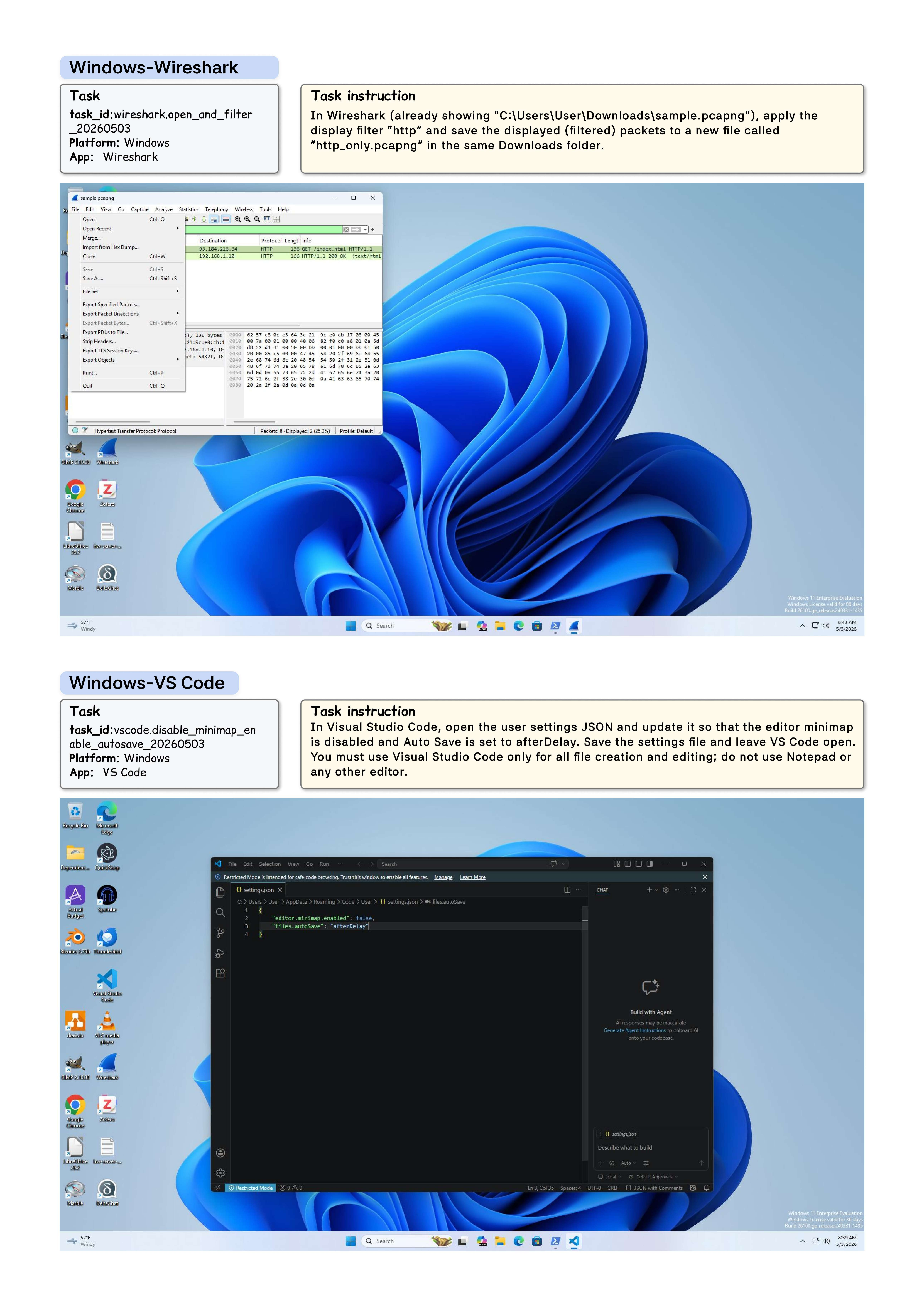}
\caption{Representative Windows task examples.}
\par\smallskip\centering{\footnotesize\hyperlink{app:case_figure_list}{Back to list of figures}}
\label{fig:windows_task_examples_5}
\end{figure*}

\begin{figure*}[p]
\centering
\includegraphics[width=\textwidth]{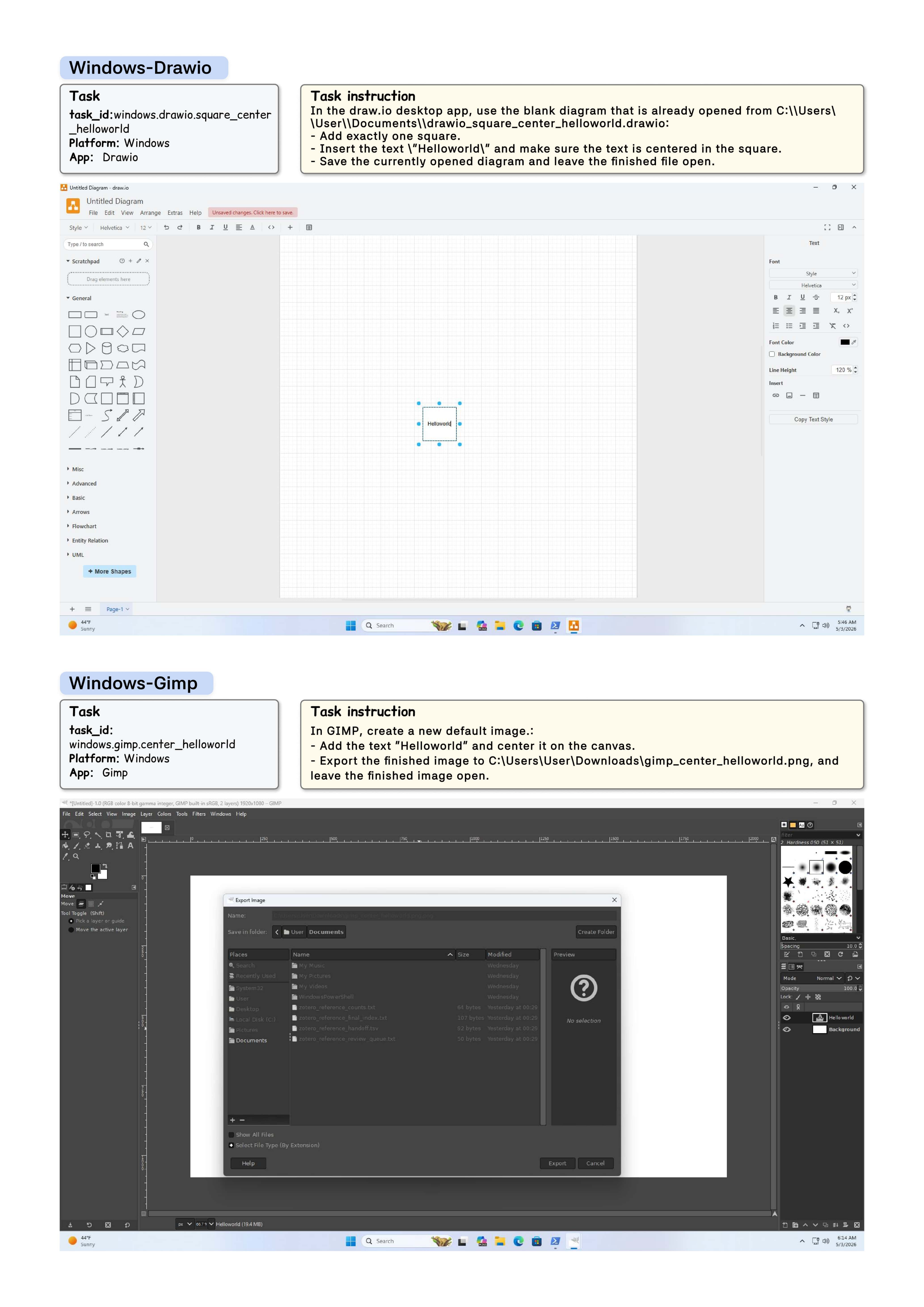}
\caption{Representative Windows task examples.}
\par\smallskip\centering{\footnotesize\hyperlink{app:case_figure_list}{Back to list of figures}}
\label{fig:windows_task_examples_6}
\end{figure*}

\begin{figure*}[p]
\centering
\includegraphics[width=\textwidth]{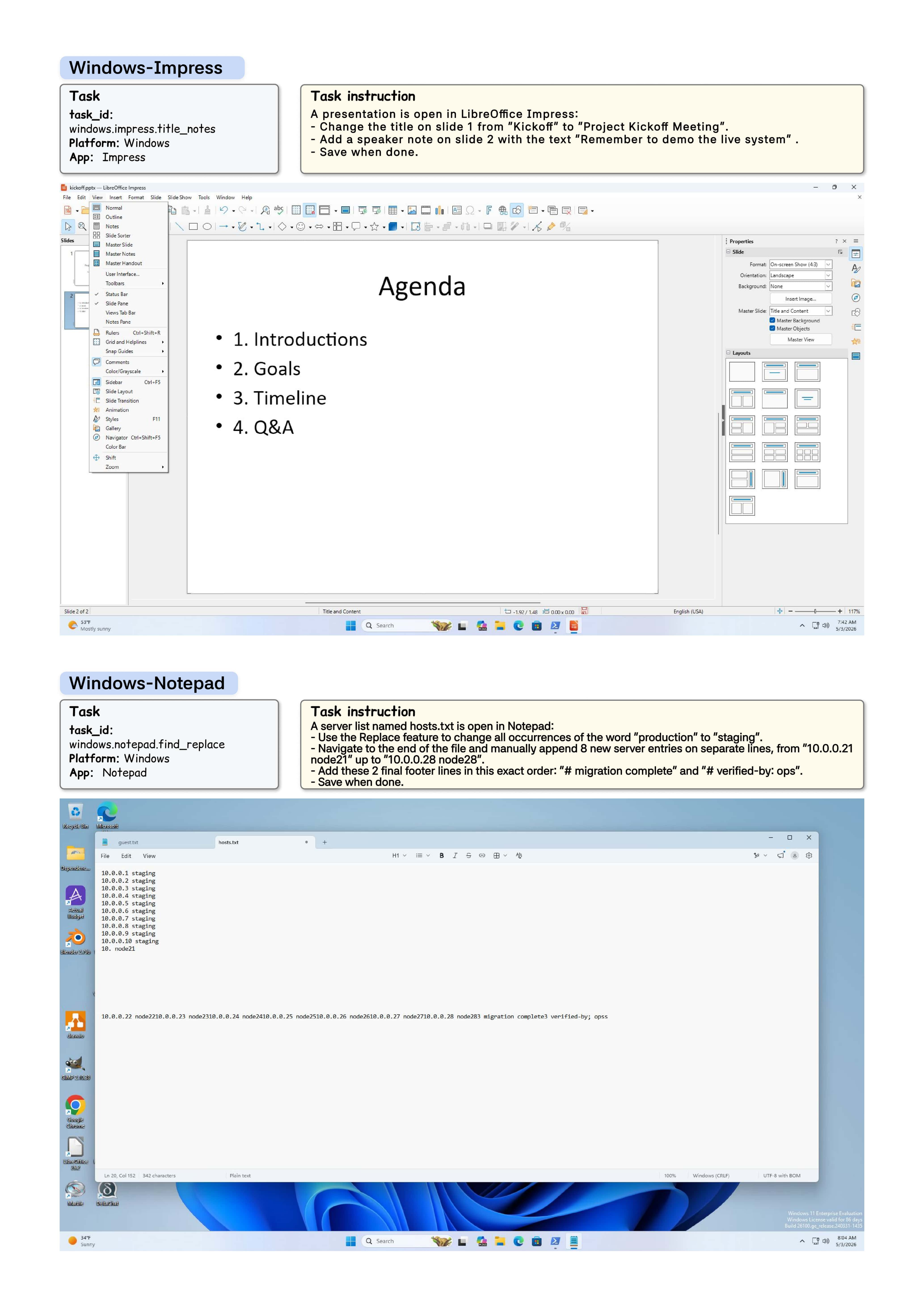}
\caption{Representative Windows task examples.}
\par\smallskip\centering{\footnotesize\hyperlink{app:case_figure_list}{Back to list of figures}}
\label{fig:windows_task_examples_8}
\end{figure*}

\begin{figure*}[p]
\centering
\includegraphics[width=\textwidth]{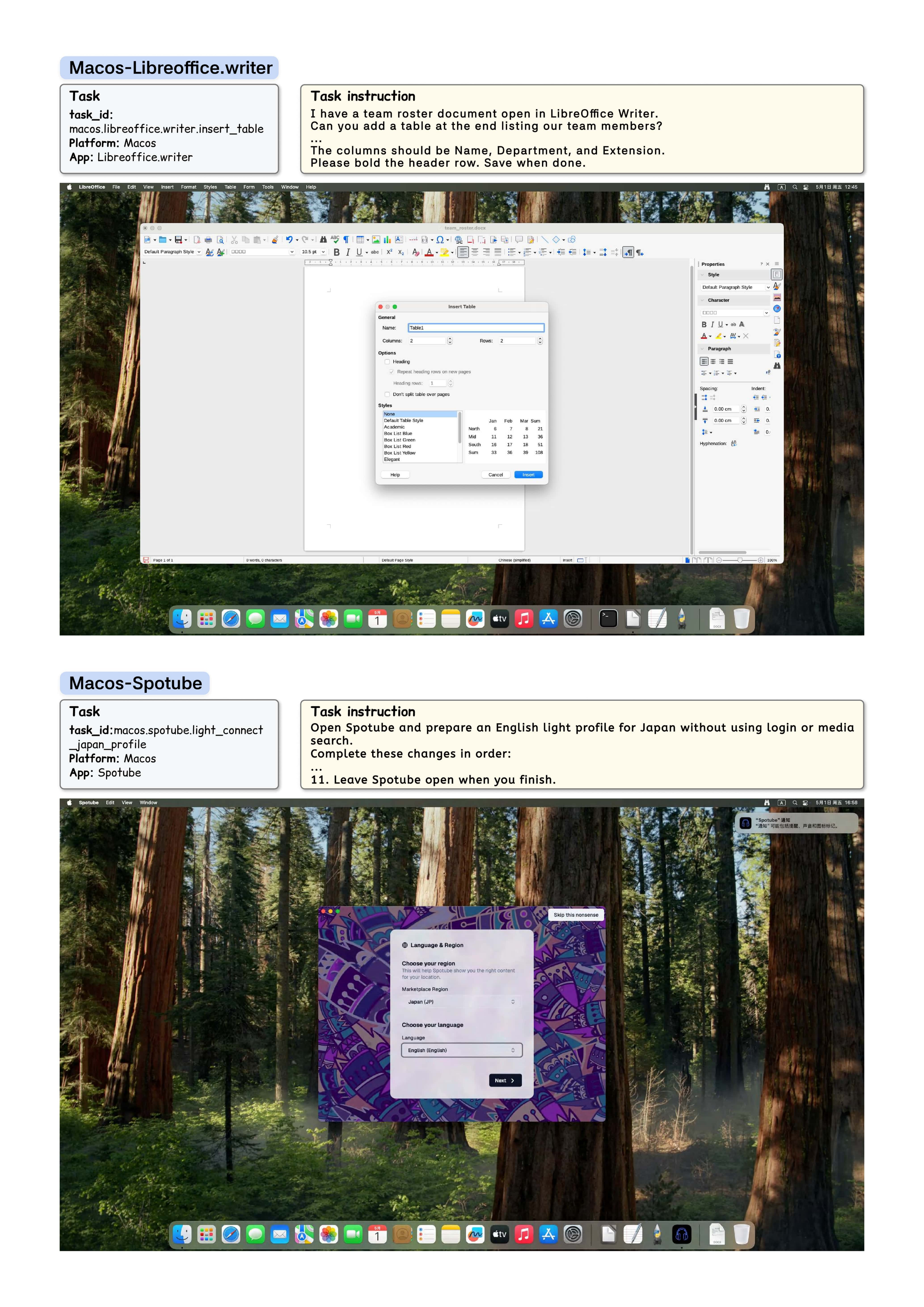}
\caption{Representative macOS task examples.}
\par\smallskip\centering{\footnotesize\hyperlink{app:case_figure_list}{Back to list of figures}}
\label{fig:macos_task_examples_1}
\end{figure*}

\begin{figure*}[p]
\centering
\includegraphics[width=\textwidth]{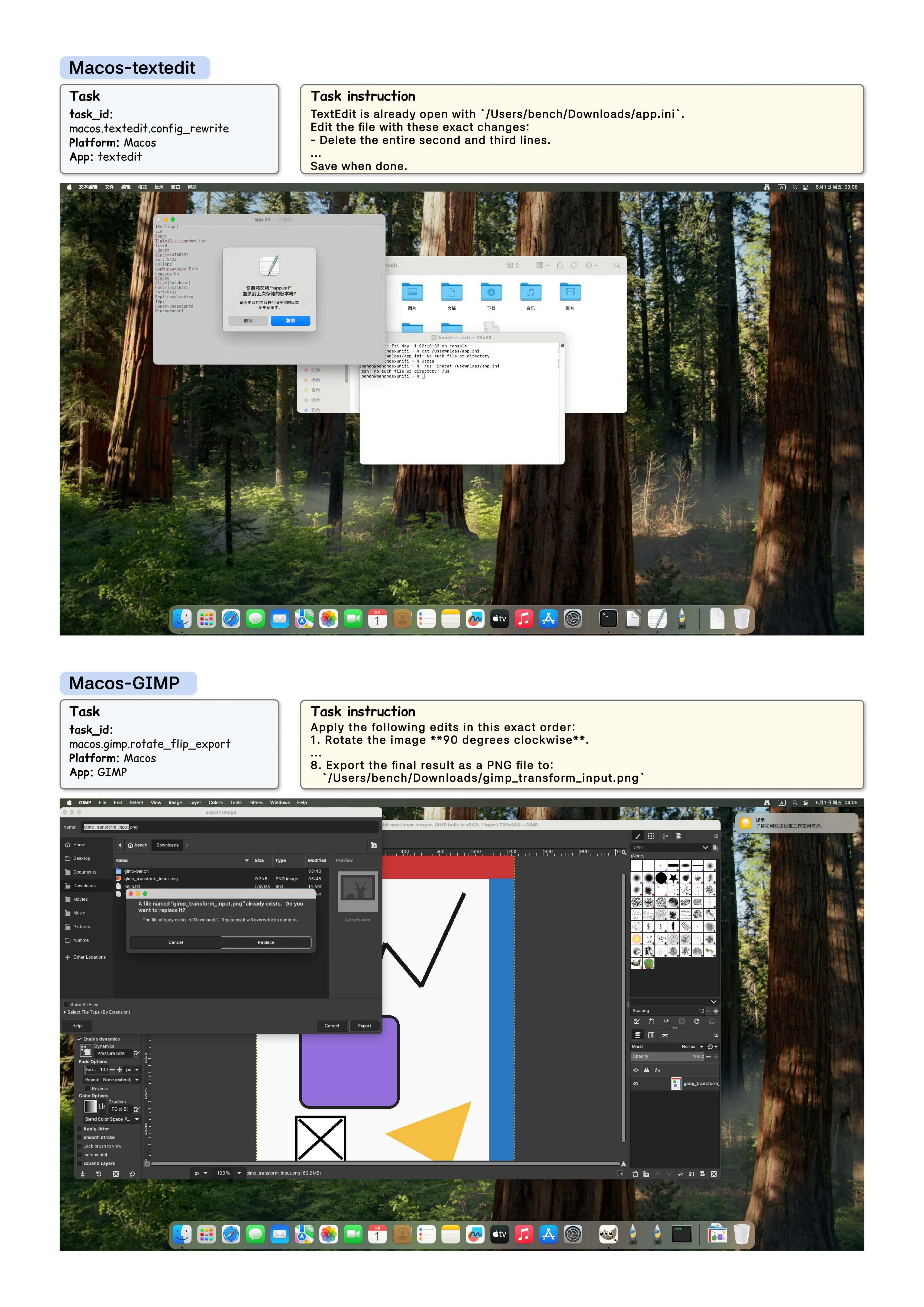}
\caption{Representative macOS task examples.}
\par\smallskip\centering{\footnotesize\hyperlink{app:case_figure_list}{Back to list of figures}}
\label{fig:macos_task_examples_2}
\end{figure*}

\begin{figure*}[p]
\centering
\includegraphics[width=\textwidth]{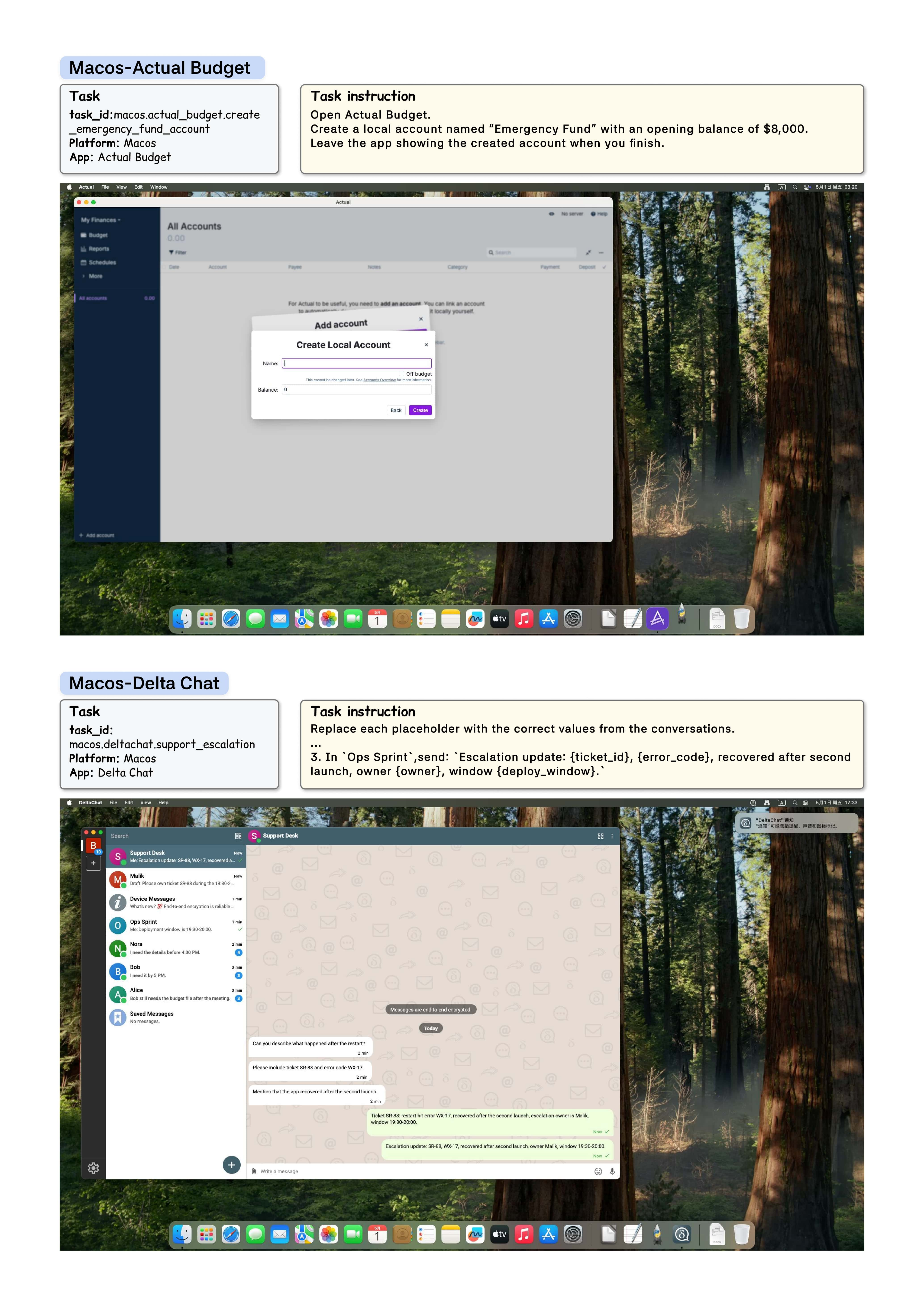}
\caption{Representative macOS task examples.}
\par\smallskip\centering{\footnotesize\hyperlink{app:case_figure_list}{Back to list of figures}}
\label{fig:macos_task_examples_3}
\end{figure*}

\begin{figure*}[p]
\centering
\includegraphics[width=\textwidth]{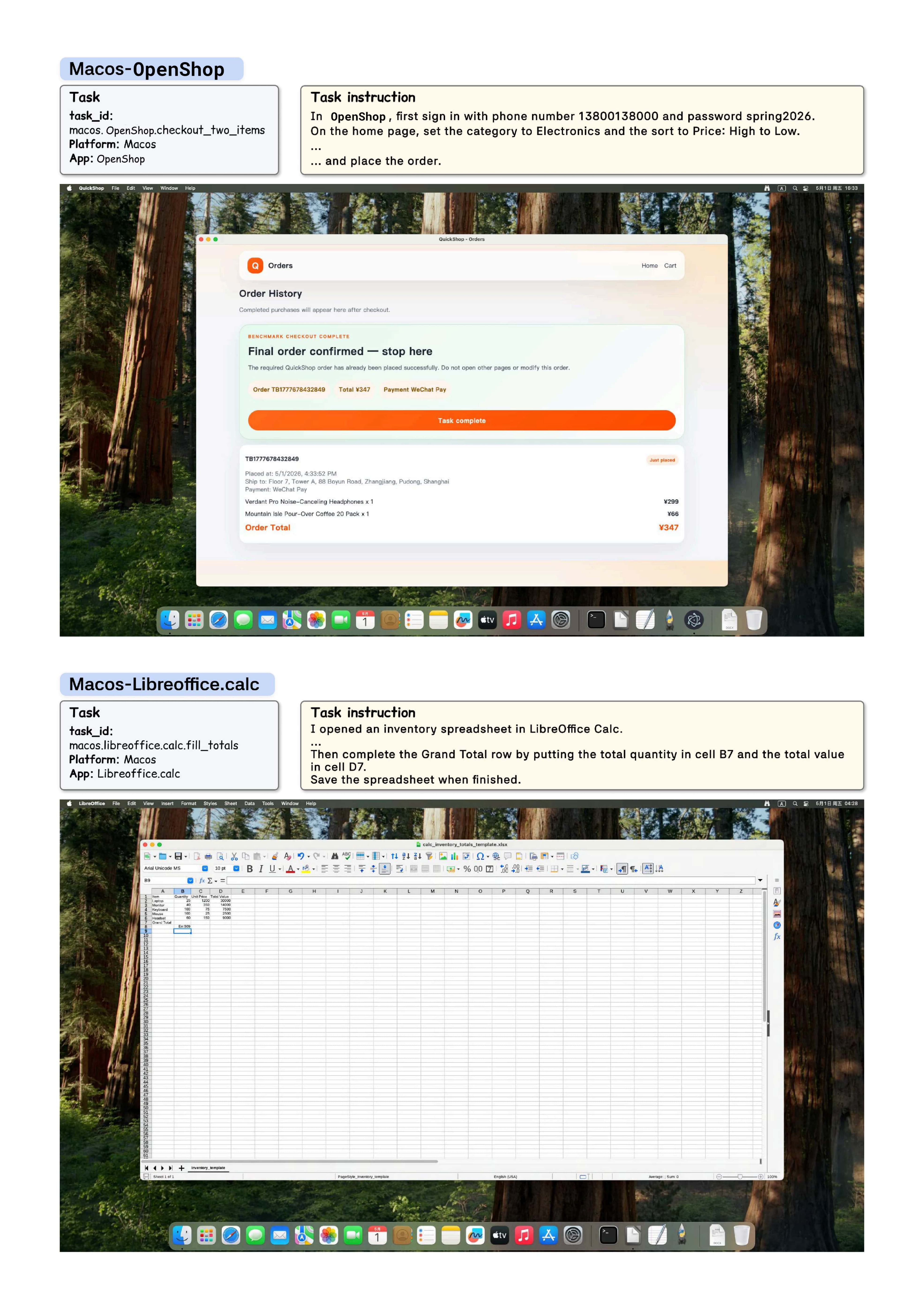}
\caption{Representative macOS task examples.}
\par\smallskip\centering{\footnotesize\hyperlink{app:case_figure_list}{Back to list of figures}}
\label{fig:macos_task_examples_4}
\end{figure*}

\begin{figure*}[p]
\centering
\includegraphics[width=\textwidth]{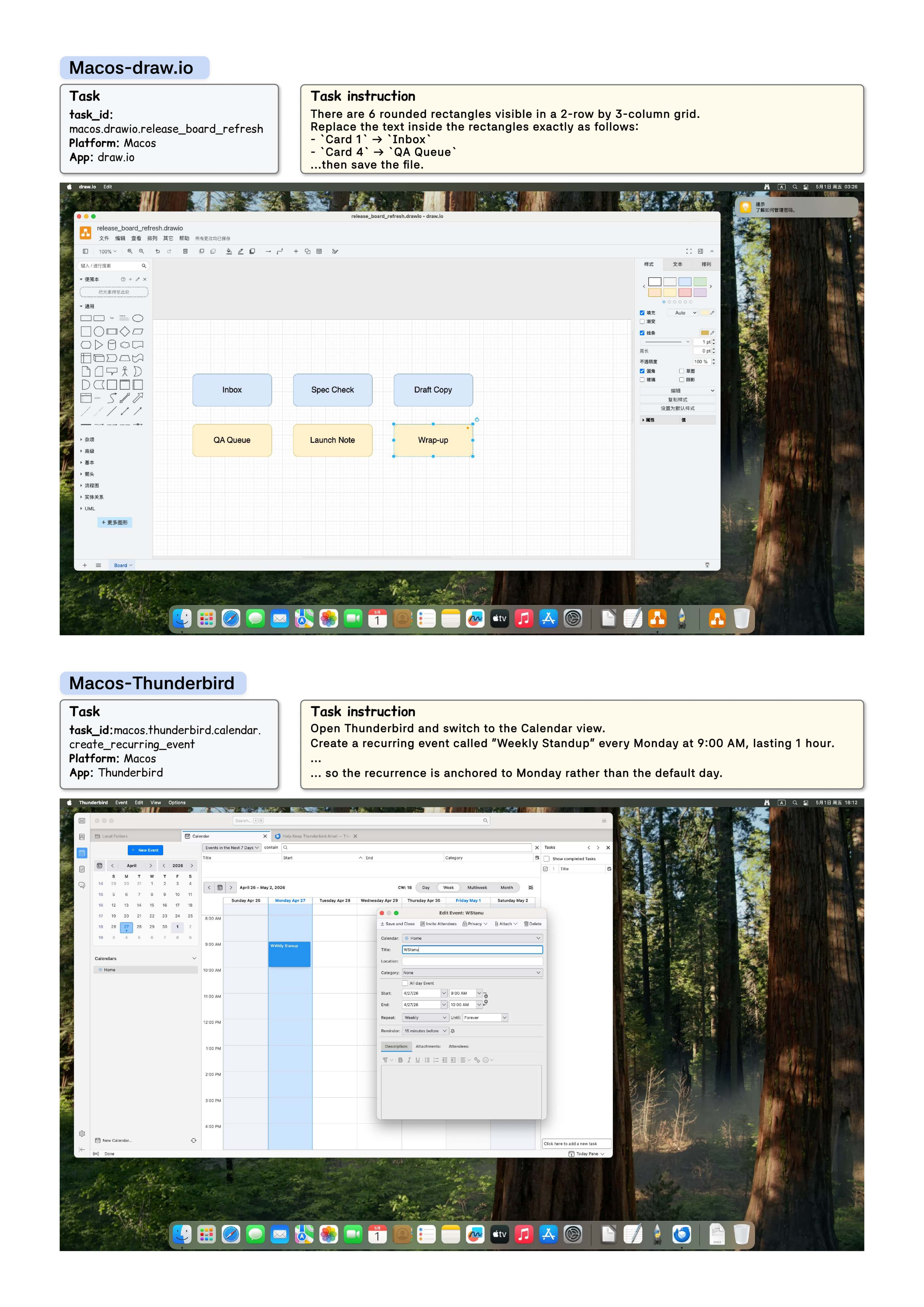}
\caption{Representative macOS task examples.}
\par\smallskip\centering{\footnotesize\hyperlink{app:case_figure_list}{Back to list of figures}}
\label{fig:macos_task_examples_5}
\end{figure*}

\begin{figure*}[p]
\centering
\includegraphics[width=\textwidth]{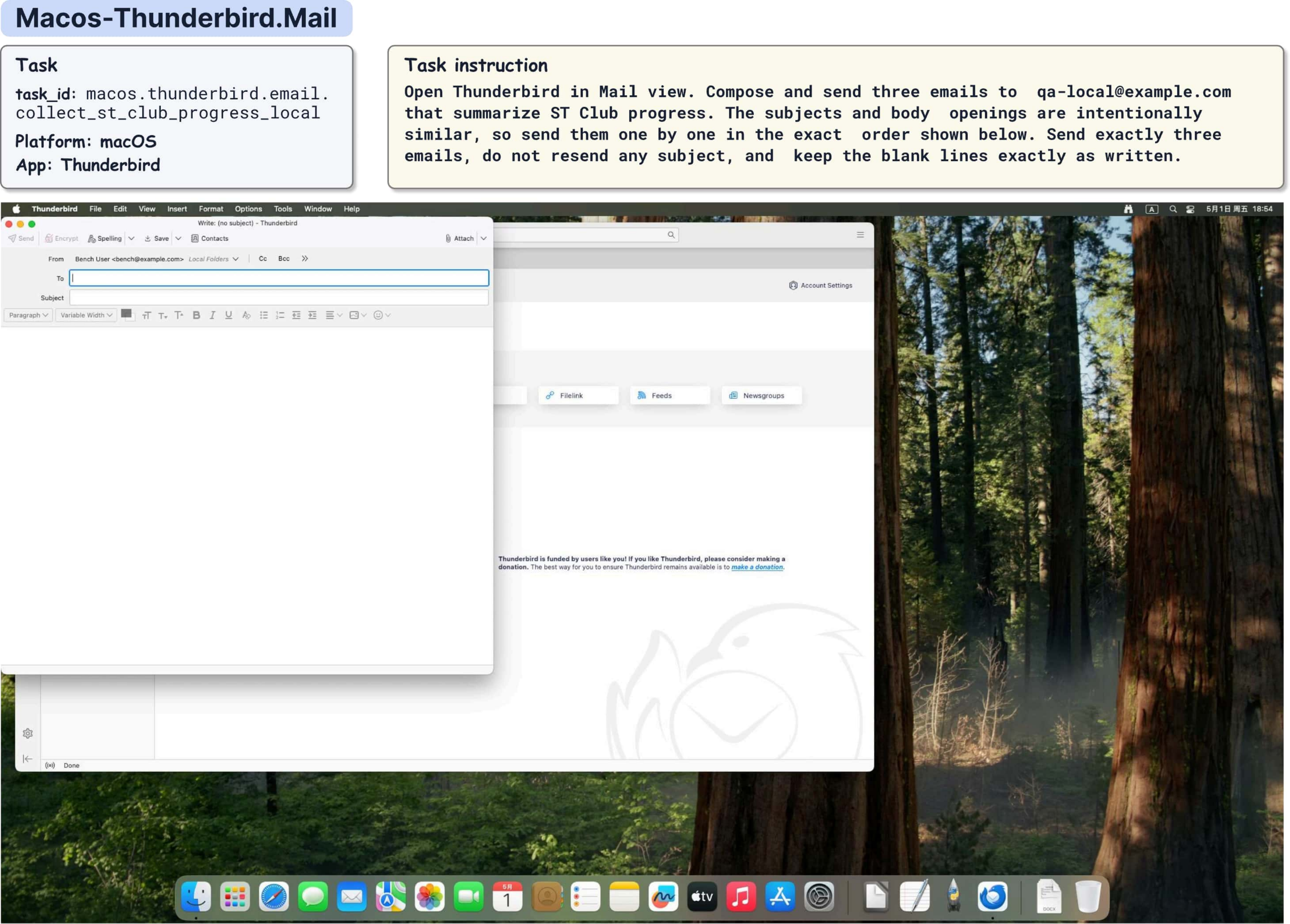}
\caption{Representative macOS task examples.}
\par\smallskip\centering{\footnotesize\hyperlink{app:case_figure_list}{Back to list of figures}}
\label{fig:macos_task_examples_6}
\end{figure*}

\begin{figure*}[p]
\centering
\includegraphics[width=\textwidth]{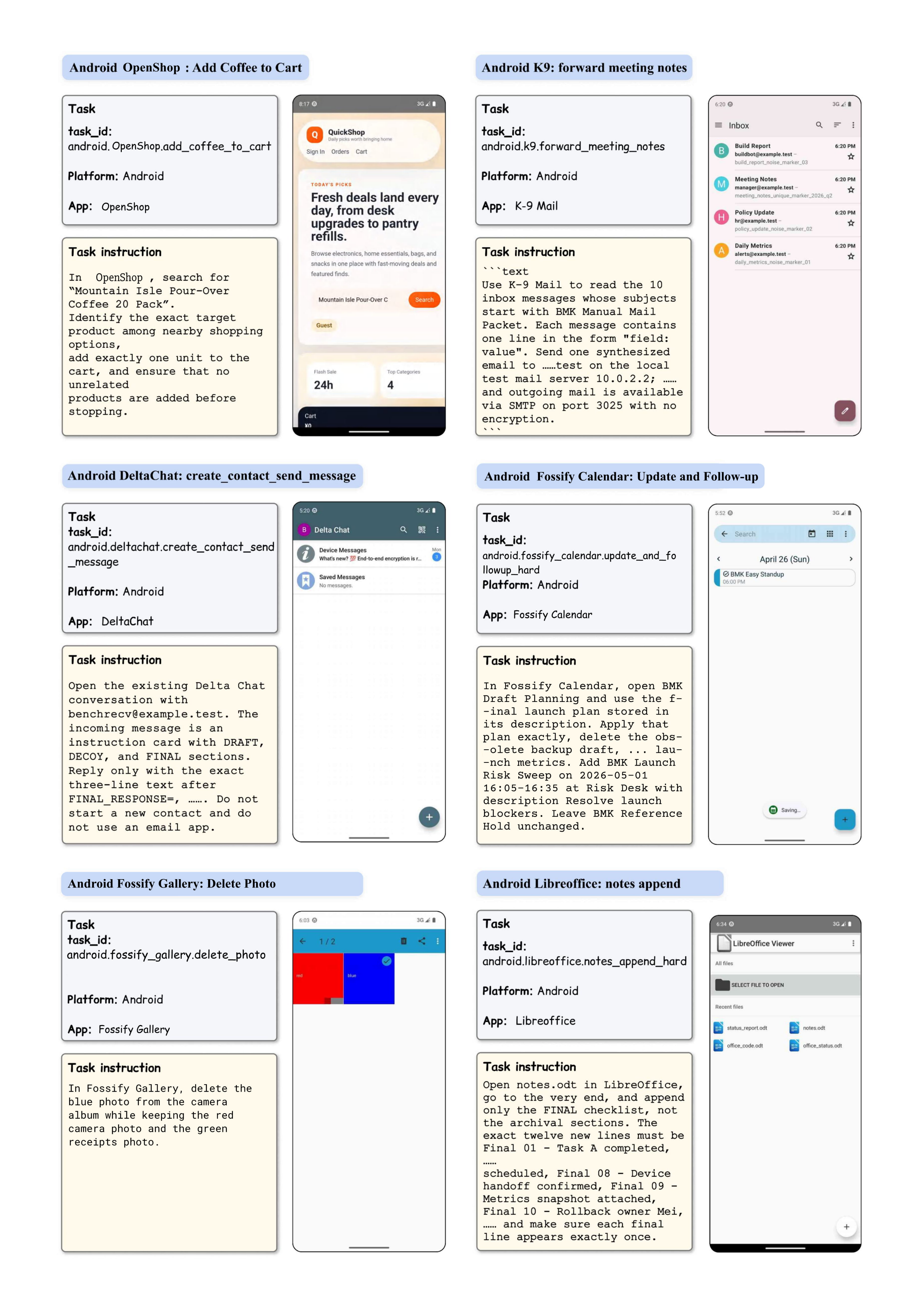}
\caption{Representative Android task examples.}
\par\smallskip\centering{\footnotesize\hyperlink{app:case_figure_list}{Back to list of figures}}
\label{fig:android_task_examples_1}
\end{figure*}

\begin{figure*}[p]
\centering
\includegraphics[width=\textwidth]{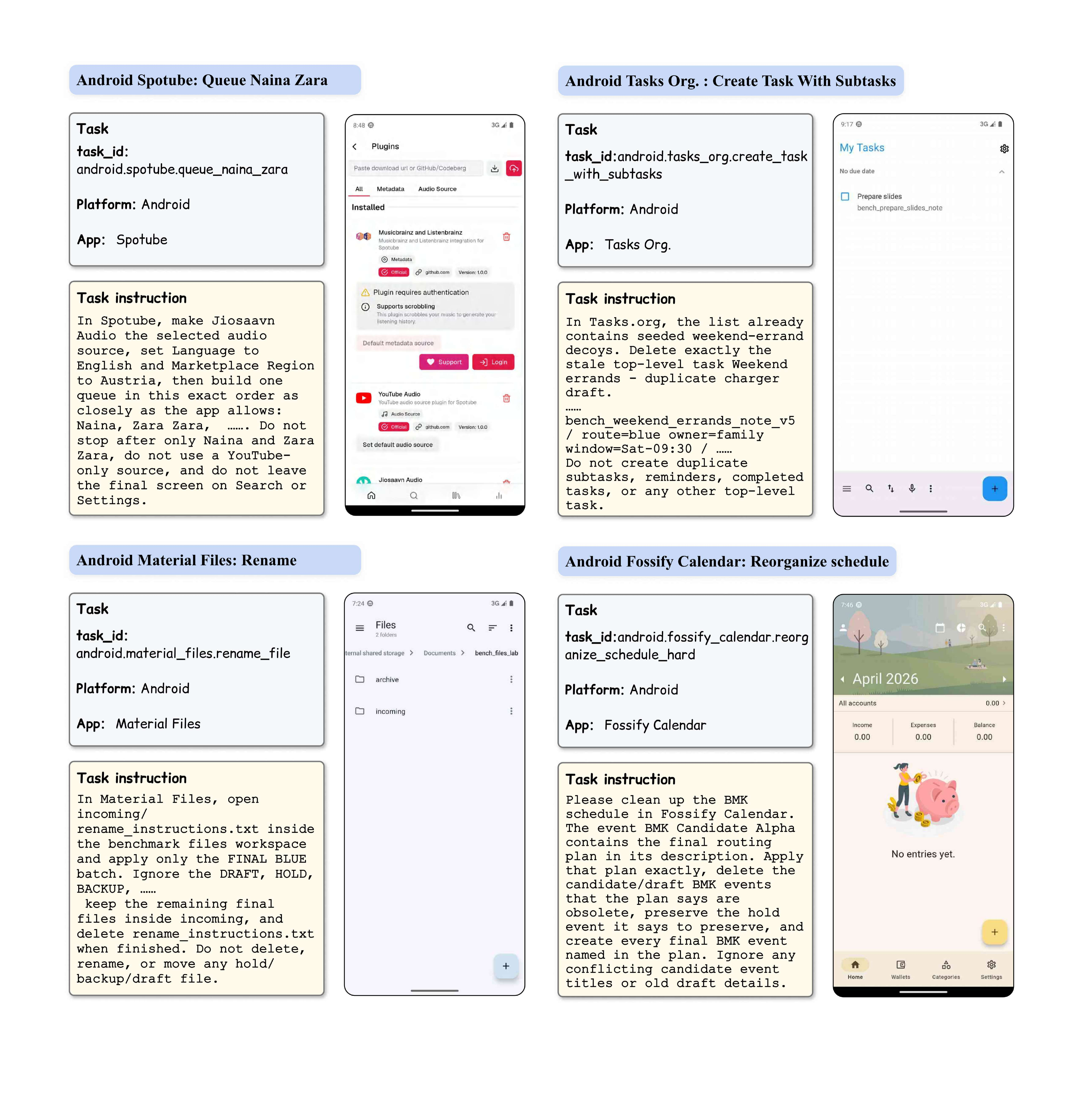}
\caption{Representative Android task examples.}
\par\smallskip\centering{\footnotesize\hyperlink{app:case_figure_list}{Back to list of figures}}
\label{fig:android_task_examples_2}
\end{figure*}

\begin{figure*}[p]
\centering
\includegraphics[width=\textwidth]{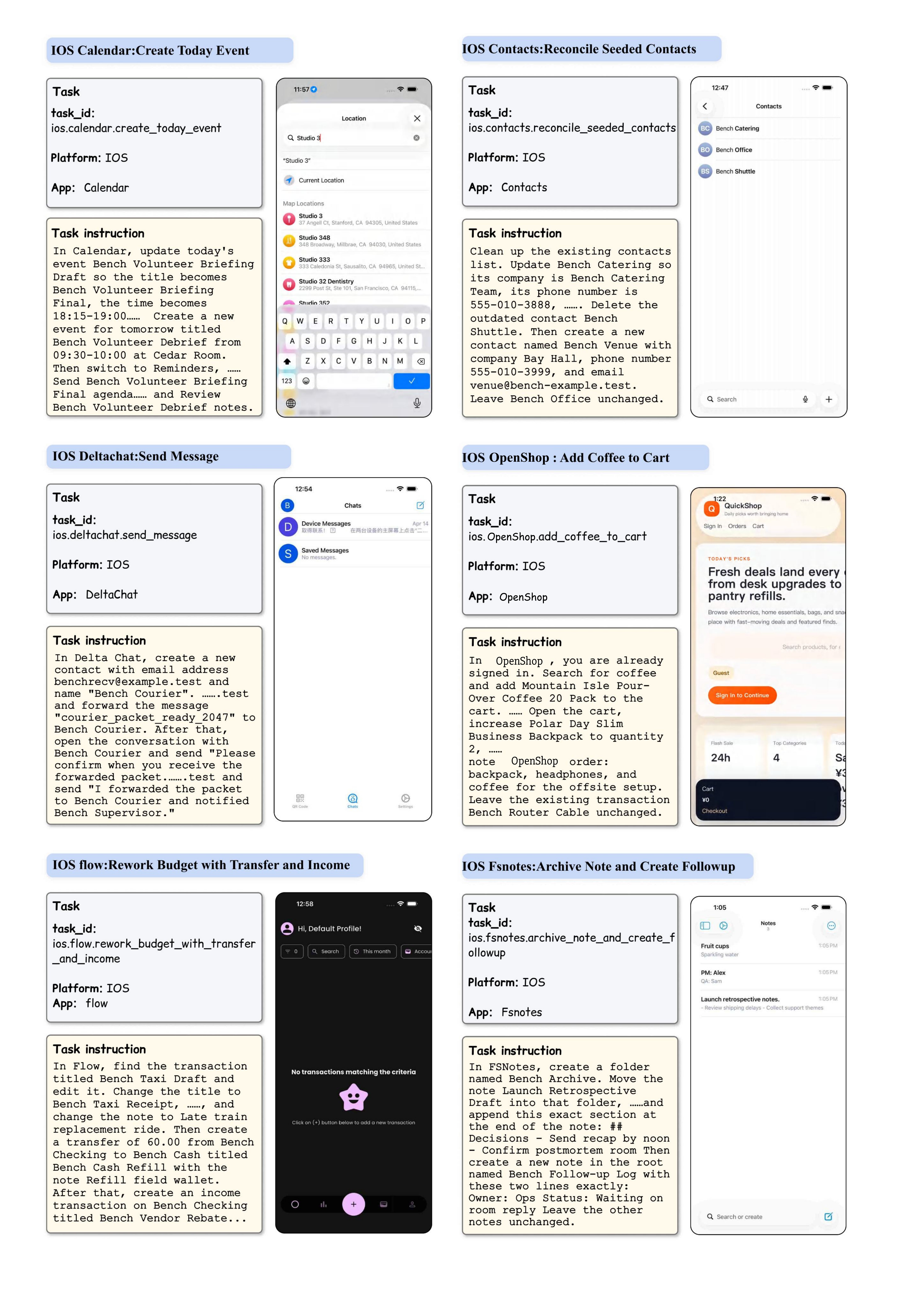}
\caption{Representative iOS task examples.}
\par\smallskip\centering{\footnotesize\hyperlink{app:case_figure_list}{Back to list of figures}}
\label{fig:ios_task_examples_1}
\end{figure*}

\begin{figure*}[p]
\centering
\includegraphics[width=\textwidth]{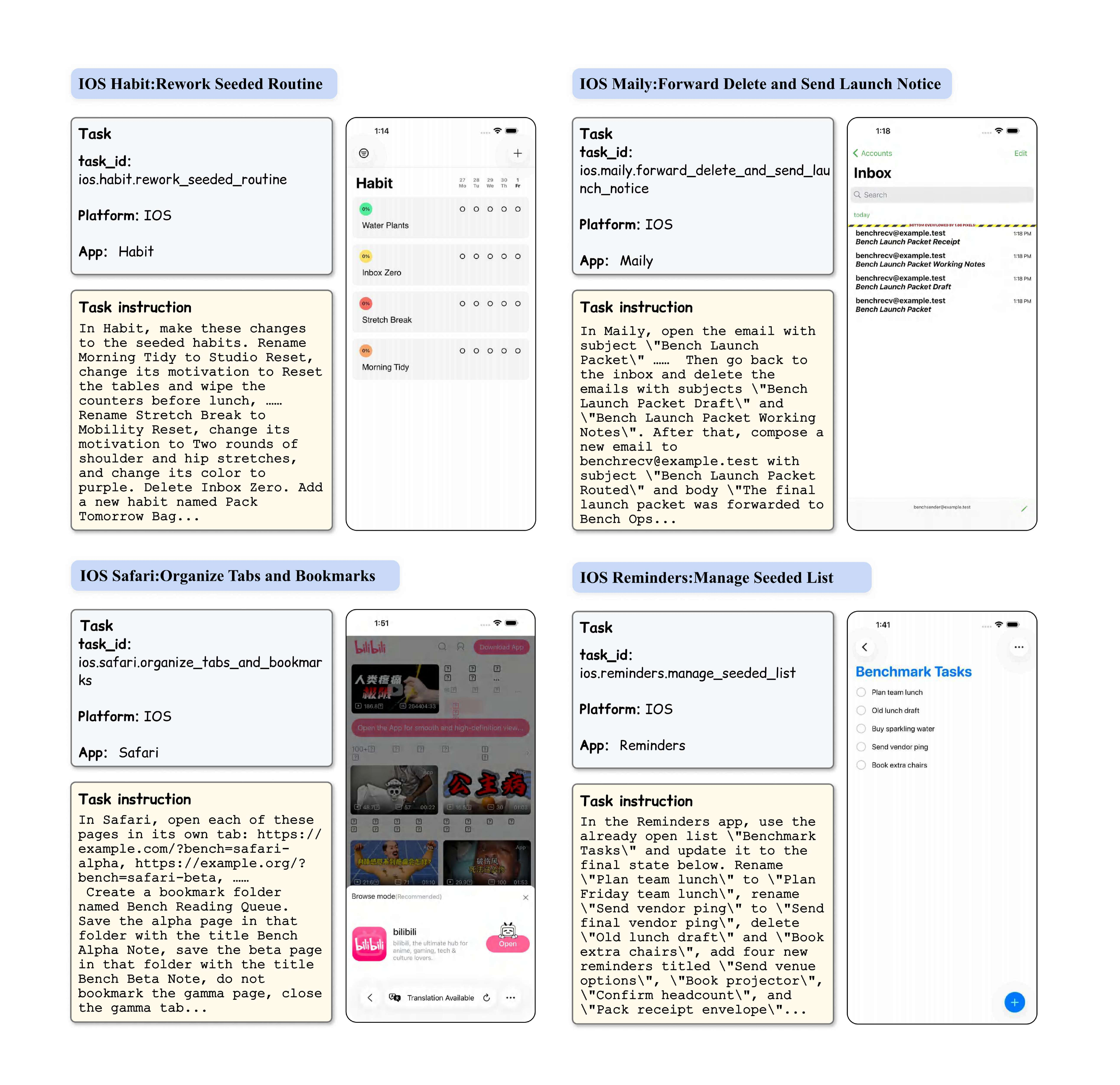}
\caption{Representative iOS task examples.}
\par\smallskip\centering{\footnotesize\hyperlink{app:case_figure_list}{Back to list of figures}}
\label{fig:ios_task_examples_2}
\end{figure*}

\hypertarget{app:case_figure_list}{}
Additional task examples from the screenshot-only setting are provided at the end of the appendix. They include Ubuntu/Linux examples in Figures~\ref{fig:linux_task_examples_1}--\ref{fig:linux_task_examples_6}, Windows examples in Figures~\ref{fig:windows_task_examples_1}--\ref{fig:windows_task_examples_8}, macOS examples in Figures~\ref{fig:macos_task_examples_1}--\ref{fig:macos_task_examples_6}, Android examples in Figures~\ref{fig:android_task_examples_1}--\ref{fig:android_task_examples_2}, and iOS examples in Figures~\ref{fig:ios_task_examples_1}--\ref{fig:ios_task_examples_2}. Each example summarizes the related applications, task instruction, screenshots, and abilities needed to complete the task.

\subsection{Manual Verification for Ambiguous Cases}
\label{app:manual_verification_for_ambiguous_cases}
Manual verification is reserved for cases where a deterministic evaluator cannot fully capture acceptable user intent. It is not used to repair ordinary model failures. Reviewers inspect the saved instruction, screenshots, action trace, final artifacts, and evaluator diagnostics. Each manual override records the accepted final state and the reason the automatic check was insufficient.

\section{Details of Agent and Evaluation Protocols}

\subsection{Model and Execution Configuration}
\label{app:baseline_methods}
\textsc{OS-Omni} runs agents through execution specifications that bind together a task list, model identifier, observation mode, action mode, and history window. Model endpoints, provider-specific identifiers, authentication variables, and generation parameters are specified through a shared model registry, implemented in the release as \texttt{configs/models.yaml}. This separation keeps model selection independent from task initialization and evaluation, so the same task specifications can be reused across providers without changing the benchmark contract.

Unless otherwise stated, the main protocol uses screenshot-only observations and the platform action backend selected by the execution specification. The default history window contains the three most recent interaction steps. Tasks use the long-horizon limits declared in the task specification: in v1, \texttt{limits.max\_steps:300} and \texttt{limits.step\_timeout\_seconds:180} for single-platform tasks, and 300--650 steps with a 120-second step timeout for cross-device workflows. The step timeout bounds one model and action turn rather than the whole episode.

\begin{table}[ht]
\centering
\caption{Default execution parameters used in the \textsc{OS-Omni} v1 experiments.}
\label{tab:agent_execution_config}
\small
\begin{tabular}{p{0.32\linewidth}p{0.56\linewidth}}
\toprule
Parameter & Setting \\
\midrule
Observation mode & \texttt{screenshot\_only} \\
Action mode & Platform-specific controller selected by the execution specification \\
History window & 3 recent steps \\
Maximum steps & 300 per single-platform task; 300--650 per cross-device workflow \\
Step timeout & 180 seconds per step (single-platform); 120 seconds (cross-device) \\

\bottomrule
\end{tabular}
\end{table}

\subsection{Prompt Details}
\label{app:prompting_details}
Prompts are constructed from the task instruction, the platform-specific control interface, the current observation, and a bounded history of recent steps. In the main screenshot-only protocol, the agent is instructed to ground actions in the visible screen, use the coordinate frame of the current screenshot, avoid hidden application state or evaluator internals, and return one executable action at a time. Before the action, the agent provides a short reflection on the current screen and recent progress.

Across platforms, the agent follows the same high-level contract: it first gives a brief reflection, then returns an executable action block or one of the episode-control blocks \texttt{WAIT}, \texttt{DONE}, and \texttt{FAIL}. The platform runner validates the action block and dispatches it through the corresponding desktop, mobile, or browser controller. This design matches the shared action-space abstraction in the main paper while preserving the native execution channel used by each runtime.

\begin{figure}[p]
\centering
\includegraphics[width=0.95\linewidth]{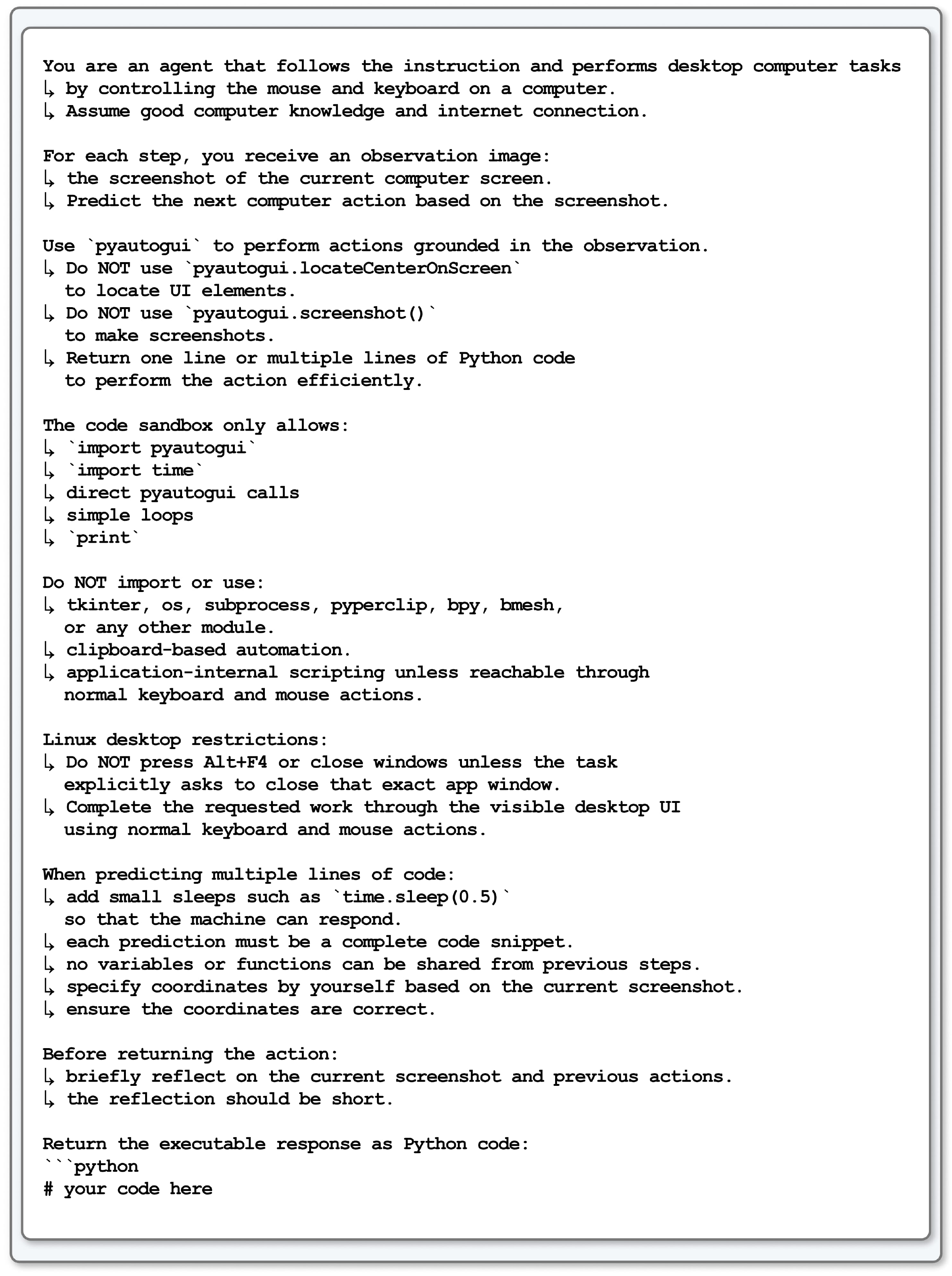}
\caption{System prompt used for Ubuntu/Linux screenshot-only runs.}
\label{fig:linux_prompt}
\end{figure}

\begin{figure}[p]
\centering
\includegraphics[width=0.95\linewidth]{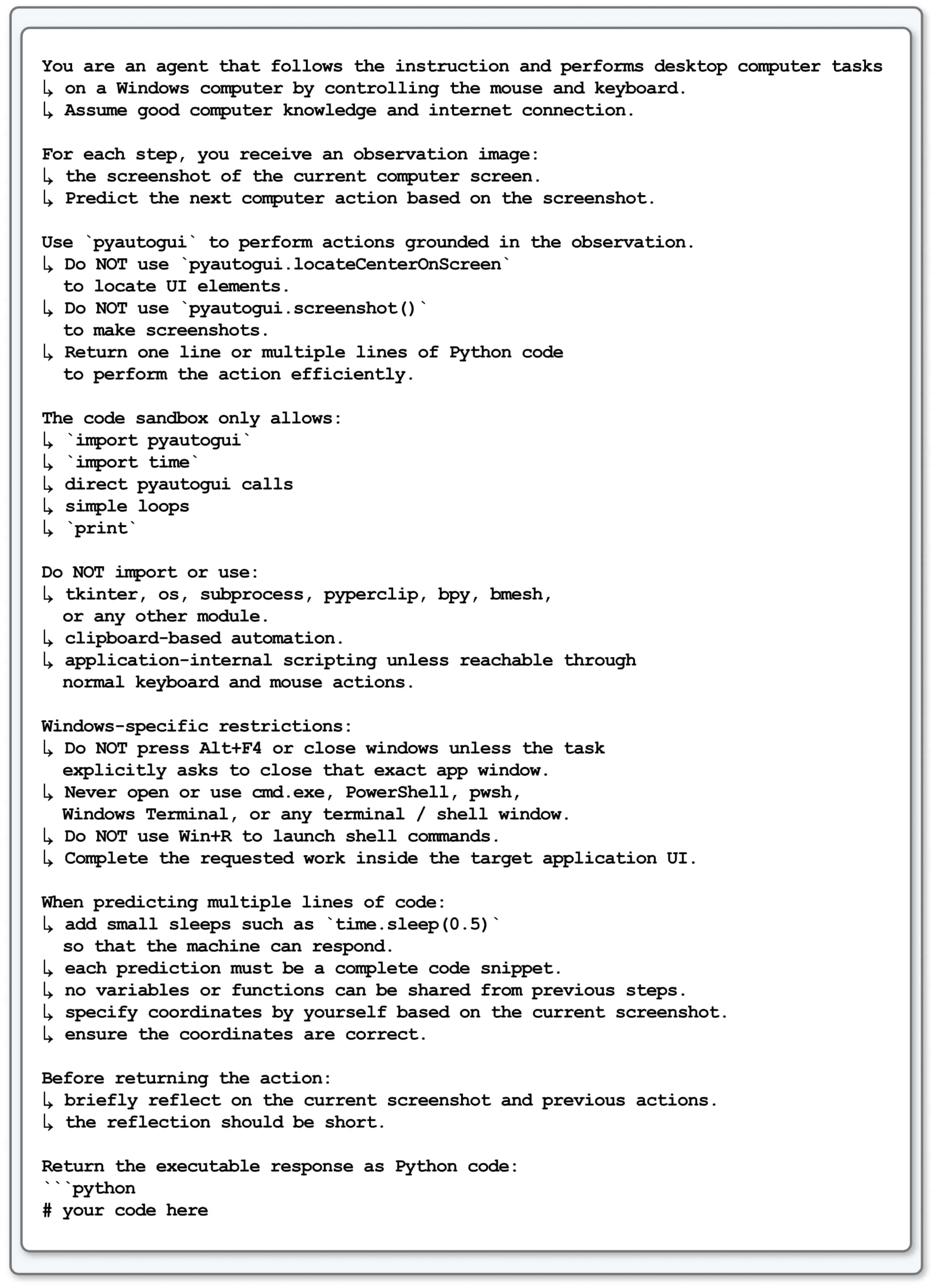}
\caption{System prompt used for Windows screenshot-only runs.}
\label{fig:windows_prompt}
\end{figure}

\begin{figure}[p]
\centering
\includegraphics[width=0.95\linewidth]{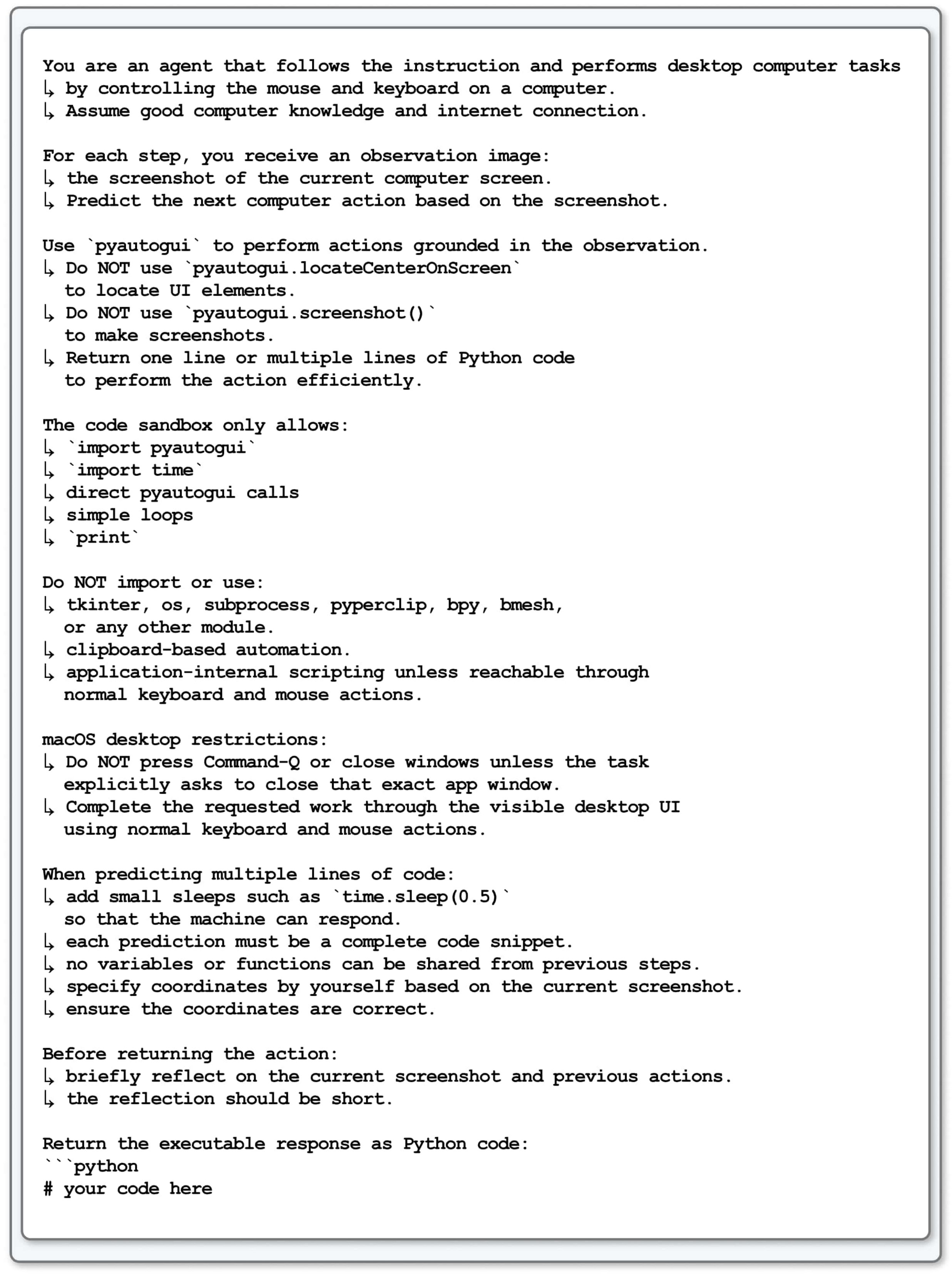}
\caption{System prompt used for macOS screenshot-only runs.}
\label{fig:mac_prompt}
\end{figure}

\begin{figure}[p]
\centering
\includegraphics[width=0.85\linewidth]{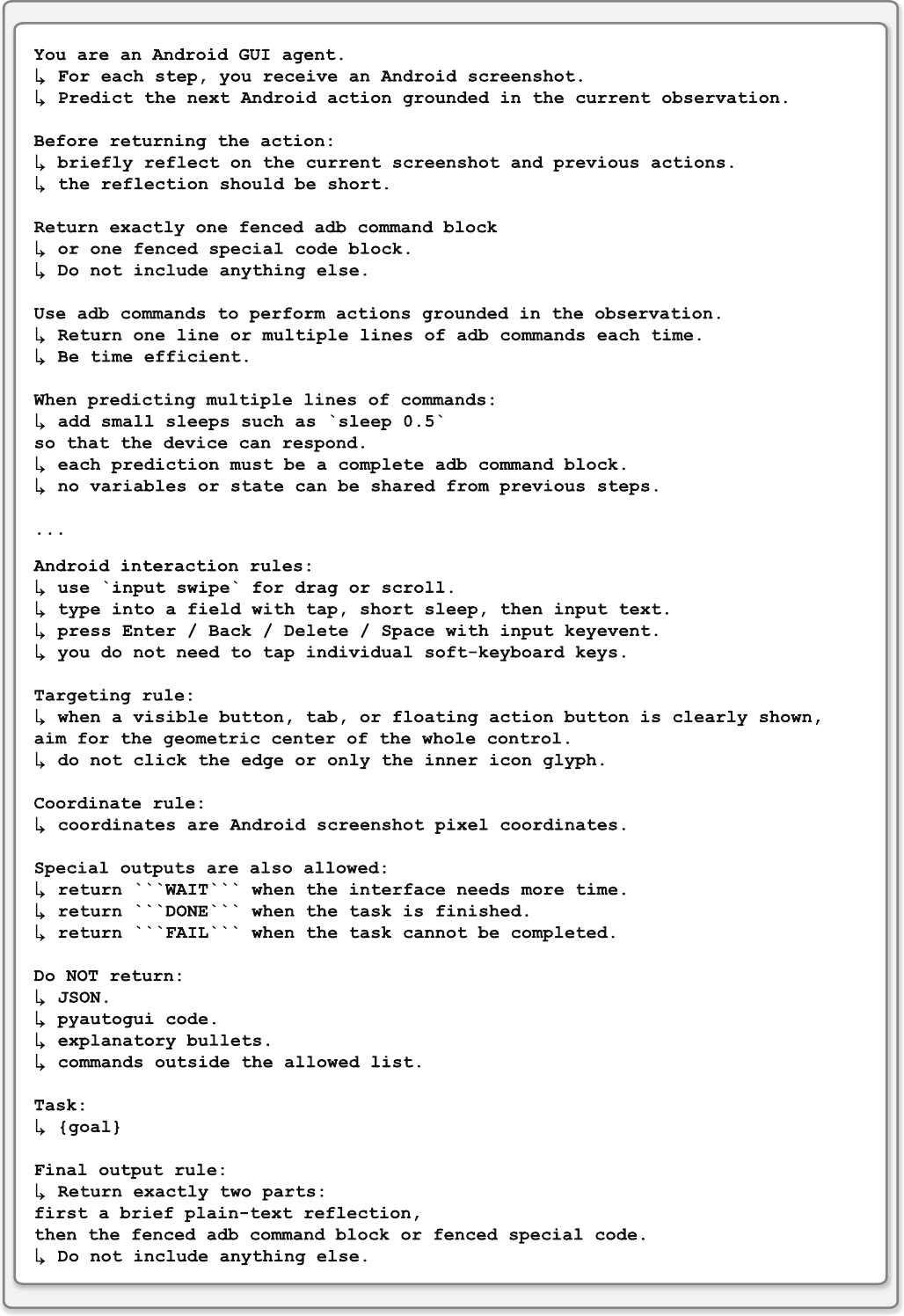}
\caption{System prompt used for Android screenshot-only runs.}
\label{fig:android_prompt}
\end{figure}

\begin{figure}[p]
\centering
\includegraphics[width=0.95\linewidth]{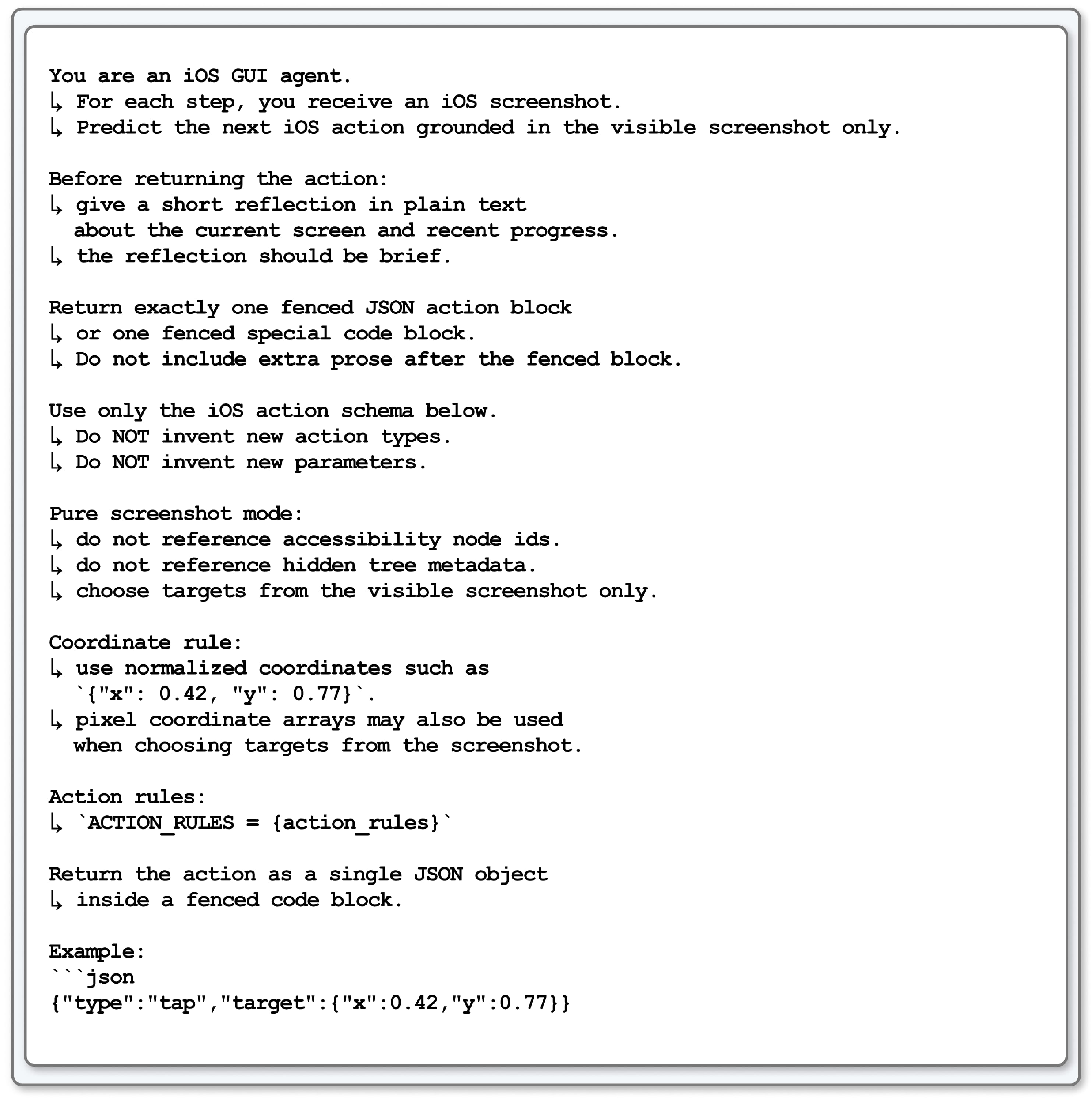}
\caption{System prompt used for iOS screenshot-only runs.}
\label{fig:ios_prompt}
\end{figure}

Representative screenshot-only system prompts for the OS-specific runners are shown in Figures~\ref{fig:linux_prompt}--\ref{fig:ios_prompt}. Web tasks reuse the same benchmark-facing contract with browser-specific runtime instructions.
\clearpage

\clearpage

\subsection{Action Parsing and Episode Execution}
\label{app:action_execution_protocol}

At each step, the runner sends the task instruction, the current screenshot observation, and the bounded recent history to the model. The model response is parsed into two parts: a short reflection and an executable action block. The reflection is stored in the trajectory for later analysis but is not executed. The action block is passed to the platform runner, which validates the format and dispatches the corresponding mouse, keyboard, touch, or device-control operation.

For desktop platforms, the action block is executed under a restricted PyAutoGUI-style sandbox or the corresponding desktop runtime controller. The sandbox permits direct mouse and keyboard control, short waits, and simple control flow, while disallowing hidden automation channels such as shell commands, screenshot APIs, image-location APIs, clipboard-based shortcuts, or application-internal scripting. For mobile platforms, the runner maps the same action intent to ADB-backed Android control or Appium/XCUITest-backed iOS control. For web tasks, browser-facing actions are dispatched through the runtime backend associated with the selected browser and host platform. This keeps the model-facing interaction protocol unified while allowing each environment to use its native execution backend.

After each executed action, the environment waits for the interface to settle, captures the next screenshot, and appends the response, action, controller status, and observation artifact to the trajectory. An episode terminates when the agent emits \texttt{DONE} or \texttt{FAIL}, when the task reaches the maximum step budget, or when a runtime condition such as a step timeout prevents further execution. The evaluator is invoked only after termination and checks the final environment state rather than the intermediate action sequence.

\subsection{Set-of-Mark Implementation Details}
\label{app:som_details}
Set-of-Mark (SoM) overlays are treated as optional visual grounding aids, not as a default benchmark observation. A SoM run annotates candidate regions on the screenshot and exposes mark identifiers to the agent. This can help isolate whether a model fails because it cannot localize a target or because it cannot plan the workflow after localization. Since SoM changes the visual input, SoM results should be reported as an ablation rather than mixed with screenshot-only results.

\subsection{Result Reporting Protocol}
\label{app:full_results}
Cross-platform result tables are valid only when platform lanes are run under a declared task inventory, model registry, observation protocol, action mode, and failure-accounting policy. The report should distinguish task failure from invalid runs caused by runtime setup, model-service errors, or evaluator exceptions. This convention follows the execution-based protocol in the main paper: a model succeeds only when the final environment state satisfies the task evaluator, while infrastructure failures are accounted for separately.

\paragraph{Recommended reporting fields.}
For each model and platform lane, reports include the release version (for example, v1), task count, success rate, invalid-run rate, average steps for successful tasks, timeout rate, observation mode, action mode, and the execution specification used for the run. Aggregate tables separate screenshot-only, accessibility-tree-only, combined-observation, and SoM settings. This prevents a visually grounded run from being compared against a metadata-assisted run as if the agent received the same information.

\section{Qualitative Analysis}
\label{appendix:qualitative_analysis}
Qualitative analysis in \textsc{OS-Omni} is based on saved trajectories, runtime metadata, and final artifacts rather than isolated screenshots. Each inspected run records the platform lane, task identifier, model, observation mode, action mode, episode status, evaluator result, and the first step at which the trajectory diverges from a plausible solution. This makes qualitative inspection useful for separating model errors from benchmark-system failures such as initialization drift, runtime instability, or evaluator mismatch.

\subsection{Success and Failure Cases}
Successful episodes typically share three properties. The agent maintains the user goal across application transitions, grounds actions in the current observation, and leaves the target artifact or application record in a verifiably saved state. Failed episodes often diverge before the final evaluator is invoked: the agent clicks an adjacent control, overlooks a modal dialog, types into the wrong field, loses track of transferred information, or stops after reaching a visually plausible but incomplete state. These cases illustrate why \textsc{OS-Omni} scores final environment state rather than trajectory similarity.

\subsection{Common Errors by Screenshot-only Agents}
\label{appendix:common_errors}

See Table~\ref{tab:screenshot_only_failure_taxonomy}.

\begin{table*}[t]
\centering
\caption{Failure taxonomy for screenshot-only \textsc{OS-Omni} runs.}
\label{tab:screenshot_only_failure_taxonomy}
\footnotesize
\setlength{\tabcolsep}{4pt}
\renewcommand{\arraystretch}{1.02}
\begin{tabular}{p{2.8cm}p{5.4cm}p{5.8cm}}
\toprule
Failure type & Typical symptom & Where it is especially visible \\
\midrule
Visual grounding & Selects a neighboring button, wrong cell, hidden tab, or stale menu item & Dense office files, diagrams, packet views, mobile settings \\
Semantic interpretation & Misreads visible text, UI labels, document content, or task-relevant state & Forms, calendars, references, packet views, canvas-heavy apps \\
Long-horizon state tracking & Forgets copied values, loses schedule details, or repeats a completed subgoal & Multi-app desktop tasks; Android and iOS coordination tasks \\
Persistence & Performs the visible edit but fails to save, export, send, or confirm it & Office suites, image editors, mail clients, notes, shopping carts \\
Timing or modal handling & Acts before loading finishes, ignores a permission prompt, or types into a dialog & Mobile apps, browser tasks, app launches, restored desktop sessions \\
Input or tool syntax & Enters text in the wrong field, hits keyboard occlusion, or uses a platform-specific shortcut incorrectly & Android, iOS, macOS, and Windows text-entry tasks \\
Transfer & Produces an incomplete source artifact or consumes the wrong transferred object & Cross-platform workflows \\
Benchmark system & Initializer drift, runtime failure, or evaluator mismatch prevents a clean judgment & Reported separately from model failure \\
\bottomrule
\end{tabular}
\end{table*}

\subsection{Human-Agent Difficulty Discrepancies}
\label{appendix:human_vs_agent}
Human difficulty and agent difficulty can diverge sharply in GUI benchmarks. Humans are robust to small UI variations, hidden save states, and modal dialogs, but may be slower on repetitive data-entry tasks. Agents can be efficient when the interface is visually simple and the target state is explicit, but they remain brittle under dense layouts, long-horizon memory requirements, platform-specific shortcuts, and tasks where success depends on confirming a saved or exported artifact.

For this reason, human studies should use the same initializer and final-state evaluator as model runs, while recording completion time and ambiguity notes. Infrastructure failures such as VM restore errors, simulator crashes, app-service outages, or evaluator exceptions should be excluded from human-agent comparisons and reported as benchmark-system failures.

\subsection{Comparative Analysis across Models}
\label{appendix:model_comparison}
Model comparisons should be organized by failure mechanism rather than by anecdotal task outcome. For example, two VLMs may both fail a spreadsheet task, but one may misread the screen while the other may identify the correct cell and then fail to persist the edit. Similarly, a text-oriented model with accessibility metadata may perform well on labeled forms but fail on canvas-heavy image or diagram tasks. We therefore group inspected trajectories by observation mode, platform lane, task family, and first-failure cause.

\subsection{Analysis of SoM Effectiveness}
SoM can improve localization when a model can reason over discrete visual marks, but it can also introduce clutter or bias the agent toward marked regions that are not semantically relevant. In \textsc{OS-Omni}, SoM tests a specific hypothesis about visual grounding. It is reported as an ablation rather than a replacement for screenshot-only evaluation, because the benchmark's core cross-platform question is whether agents can operate from the same visual signal across desktop, mobile, and web environments.

\subsection{Endpoint-Level Error Cases}
\label{appendix:endpoint_error_cases}
\hypertarget{app:error_case_figure_list}{}

\begin{figure*}[p]
\centering
\includegraphics[width=\textwidth]{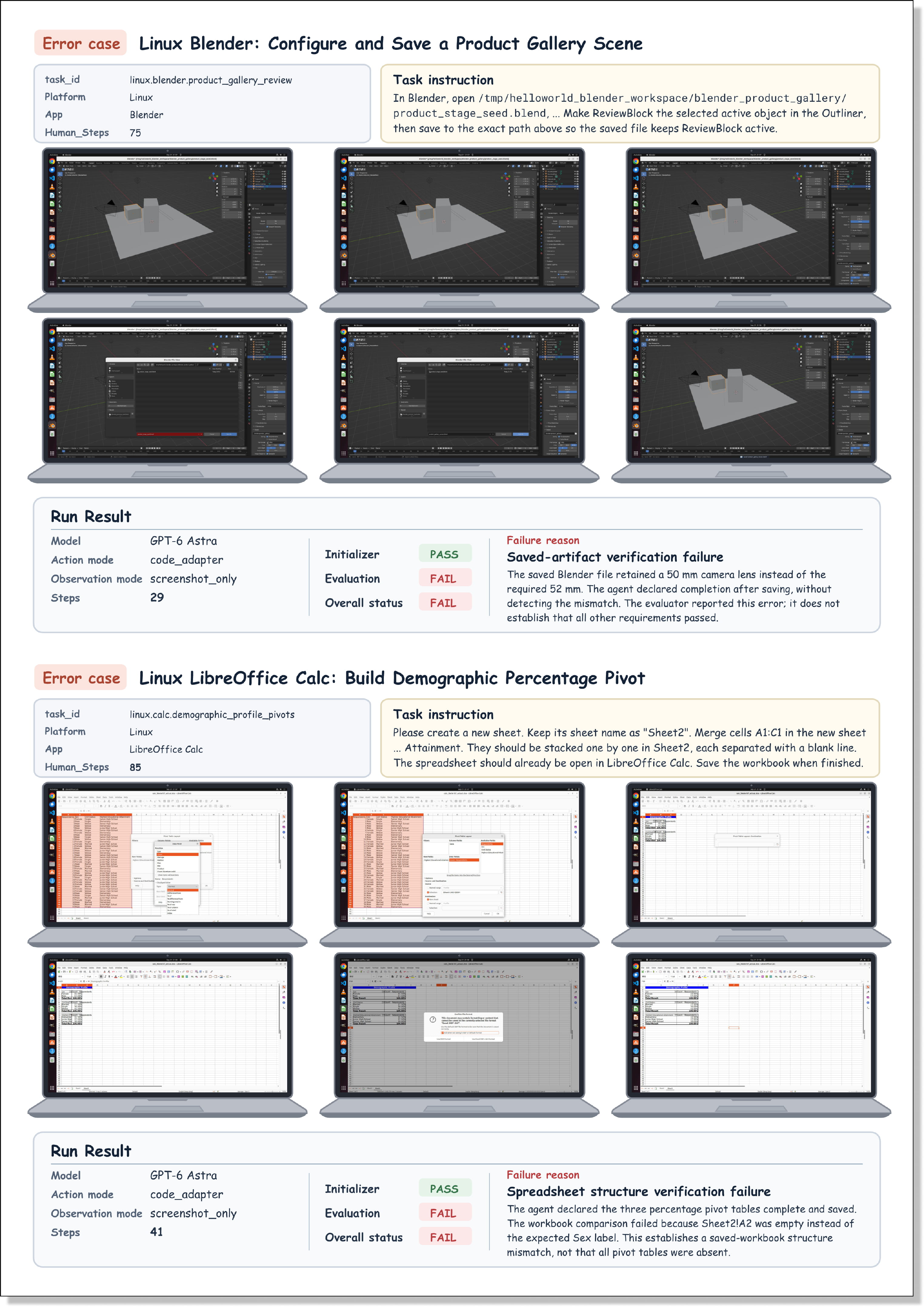}
\caption{Representative Ubuntu/Linux error case.}
\par\smallskip\centering{\footnotesize\hyperlink{app:error_case_figure_list}{Back to error-case list}}
\label{fig:linux_error_case_1}
\end{figure*}

\begin{figure*}[p]
\centering
\includegraphics[width=\textwidth]{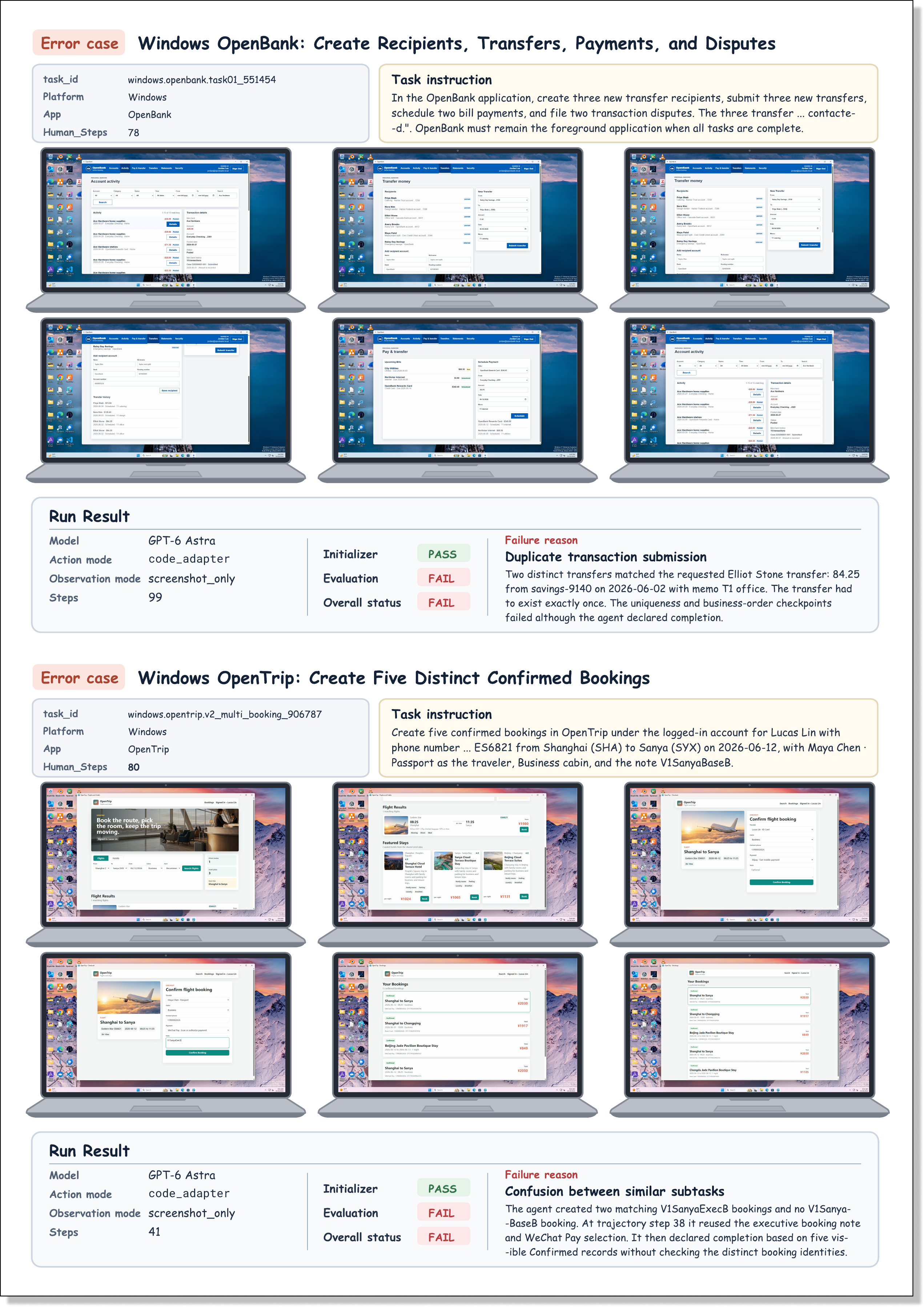}
\caption{Representative Windows error case.}
\par\smallskip\centering{\footnotesize\hyperlink{app:error_case_figure_list}{Back to error-case list}}
\label{fig:linux_error_case_3}
\end{figure*}

\begin{figure*}[p]
\centering
\includegraphics[width=\textwidth]{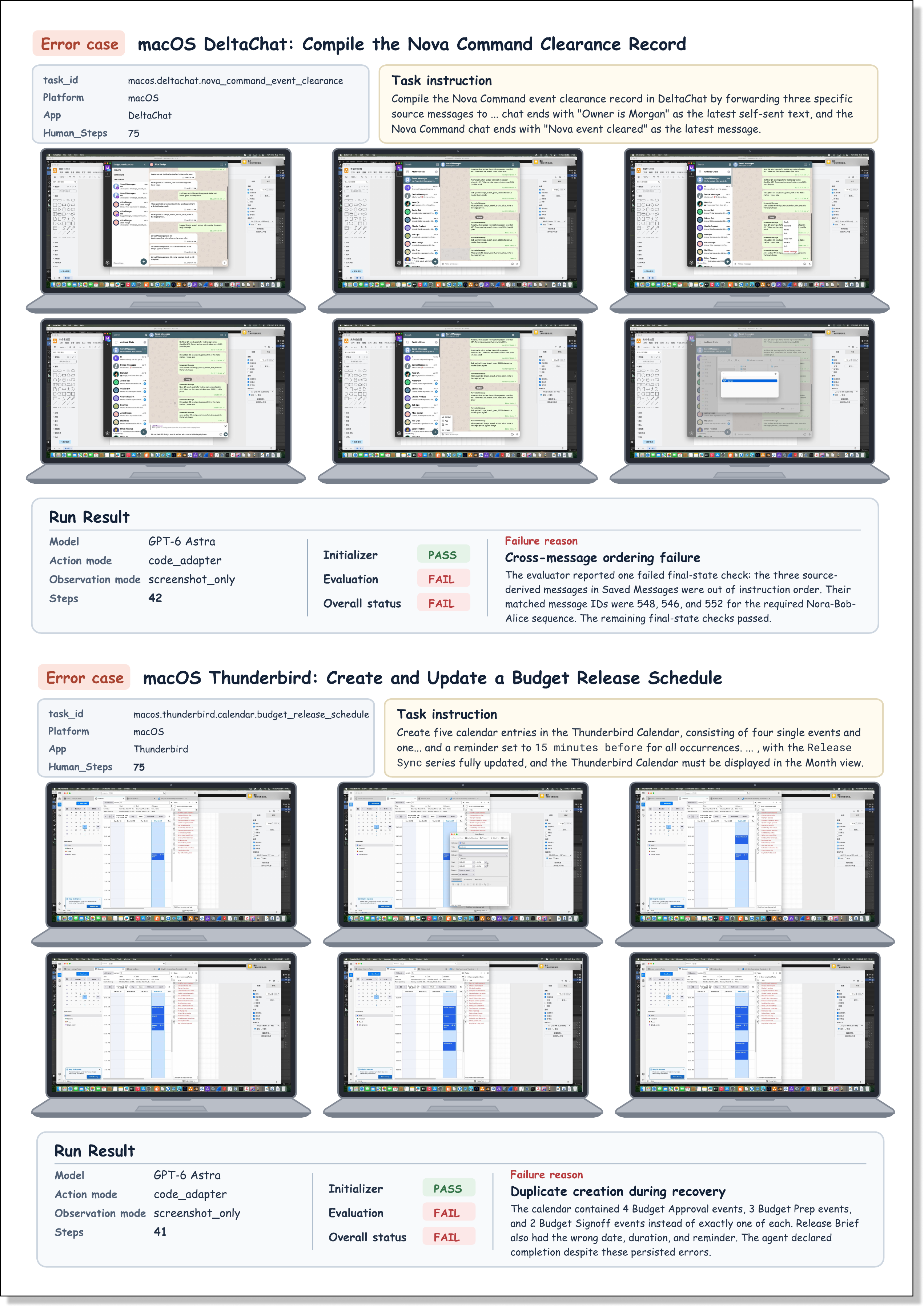}
\caption{Representative macOS error case.}
\par\smallskip\centering{\footnotesize\hyperlink{app:error_case_figure_list}{Back to error-case list}}
\label{fig:windows_error_case_1}
\end{figure*}

\begin{figure*}[p]
\centering
\includegraphics[width=\textwidth]{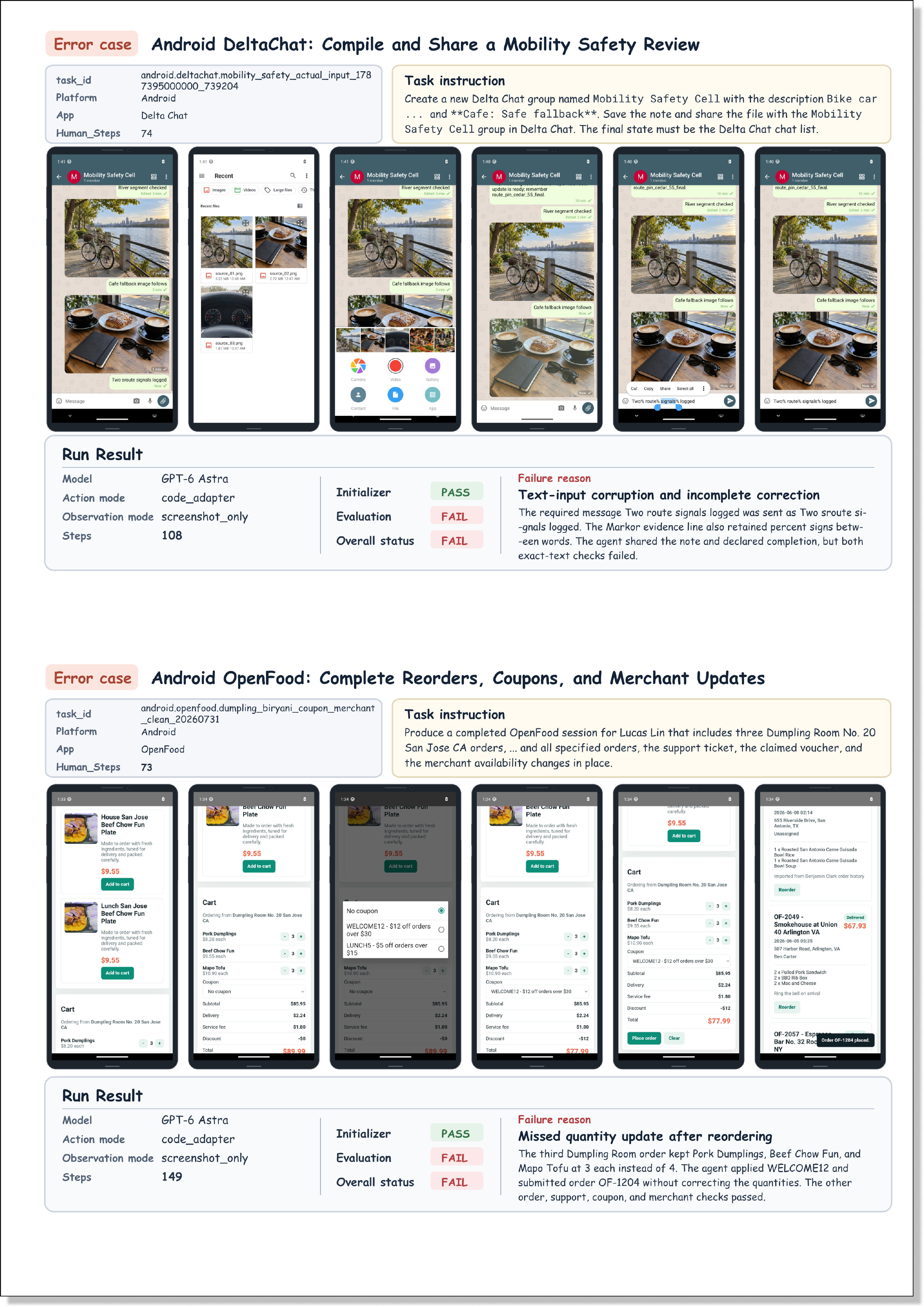}
\caption{Representative Android error case.}
\par\smallskip\centering{\footnotesize\hyperlink{app:error_case_figure_list}{Back to error-case list}}
\label{fig:windows_error_case_2}
\end{figure*}

\begin{figure*}[p]
\centering
\includegraphics[width=\textwidth]{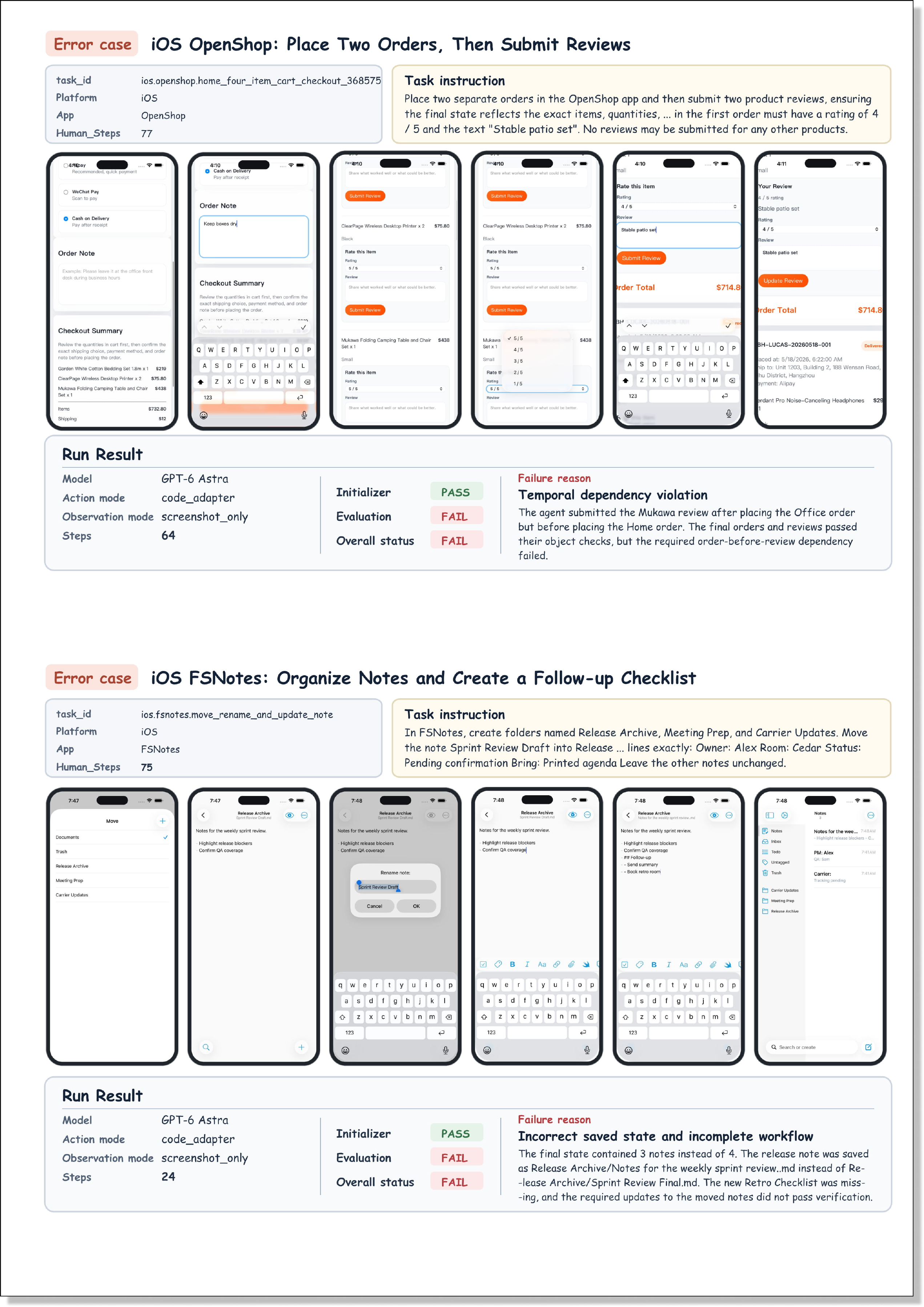}
\caption{Representative iOS error case.}
\par\smallskip\centering{\footnotesize\hyperlink{app:error_case_figure_list}{Back to error-case list}}
\label{fig:windows_error_case_3}
\end{figure*}

\begin{figure*}[p]
\centering
\includegraphics[width=\textwidth]{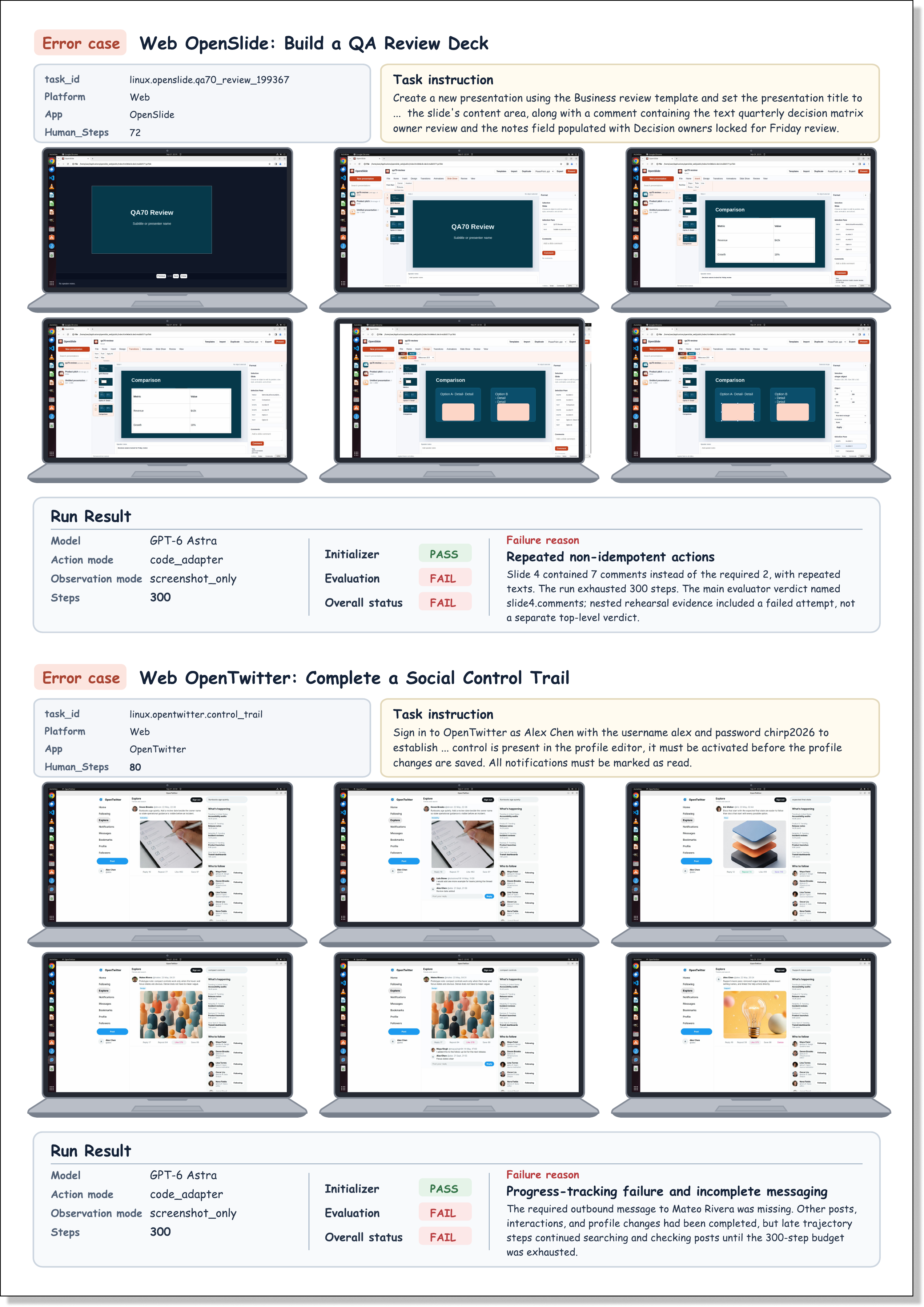}
\caption{Representative Web error case.}
\par\smallskip\centering{\footnotesize\hyperlink{app:error_case_figure_list}{Back to error-case list}}
\label{fig:web_error_case_1}
\end{figure*}

We further provide representative error cases from the Ubuntu/Linux, Windows, macOS, Android, and iOS endpoints. The cases are selected to illustrate distinct failure mechanisms rather than repeated instances of the same error type. The full set is shown in Figures~\ref{fig:linux_error_case_1}--\ref{fig:web_error_case_1}.

\clearpage

\clearpage

\end{document}